%% file: arxiv_version.tex
\documentclass[11pt]{article}

\usepackage{appendix}
\usepackage{amssymb,amsmath,amsfonts,dsfont}
\usepackage{graphicx}
\usepackage{graphicx}
\usepackage{dcolumn}
\usepackage{ifthen}
\usepackage{booktabs}
\usepackage{bm}
\usepackage{units}

\usepackage[utf8]{inputenc} 
\usepackage[T1]{fontenc}    

\usepackage{float}
\usepackage[skip=0pt]{caption}

\usepackage{hyperref}
\hypersetup{
	bookmarks=true,         
	unicode=false,          
	pdftoolbar=true,        
	pdfmenubar=true,        
	pdffitwindow=false,     
	pdfstartview={FitH},    
	pdfauthor={},     
	pdfsubject={Subject},   
	pdfcreator={},   
	pdfproducer={Producer}, 
	pdfnewwindow=true,      
	colorlinks=true,       
	linkcolor=red,          
	citecolor=green,        
	filecolor=magenta,      
	urlcolor=cyan           
}
\usepackage{breakurl}

\usepackage[
    backend=bibtex,
    url=false,isbn=false,doi=false,eprint=false,
    date=year,
    hyperref=auto,
    style=alphabetic,
    natbib=true,
    sorting=nty,
    firstinits=true,
    maxnames=2, maxbibnames=10,
]{biblatex}

\bibliography{bibliography.bib} 

\AtEveryBibitem{%
  \clearname{translator}%
  \clearlist{publisher}%
  \clearfield{pagetotal}%
  \clearname{editor}
  \clearfield{pages}
  \clearfield{series}
  \clearlist{address}
  \clearfield{address}
  \clearname{address}
}

\usepackage{tikz}
\usepackage{pgfplots}
\usepgfplotslibrary{groupplots} 
\usetikzlibrary{pgfplots.groupplots}
\usetikzlibrary{matrix,arrows,decorations.pathmorphing}
\usepgfplotslibrary{fillbetween}
\usetikzlibrary{shadows,arrows}
\usetikzlibrary{positioning}
\pgfplotsset{compat=1.10}

\usepackage{changepage} 

\usetikzlibrary {arrows.meta} 
\usetikzlibrary{fit} 

\usepgfplotslibrary{colorbrewer}
\pgfplotsset{compat=newest}

\usetikzlibrary{calc}
\usetikzlibrary{positioning}

\usepackage{graphics}
\usepackage{comment}
\usepackage{caption} 
\usepackage{changepage}

\usepackage{hyperref}       
\usepackage{url}            
\usepackage{amsfonts}       
\usepackage{nicefrac}       
\usepackage{microtype}      

\usepackage{wrapfig}

\newcommand\Vrecon{V_{\text{rec}}}
\newcommand\Vtrain{V_{\text{train}}}

\newdimen\nodeDist
\usepackage{url}
\usepackage{breakurl}
\usepackage{outlines} 

\usepackage[vlined,ruled,linesnumbered]{algorithm2e}

\usepackage{comment}

\usepackage{colortbl}
\definecolor{tblblue}{RGB}{101,124,191}
\definecolor{tblred}{rgb}{1,0.93,0.93}
\definecolor{DarkBlue}{rgb}{0,0,0.7} 
\definecolor{BrickRed}{RGB}{203,65,84}

\newcommand\vtheta{{\bm \theta}}

\newcommand\mSigma{\bm \Sigma}

\DeclareMathOperator{\argmin}{argmin}

\usepackage{defs}
\usepackage{bbm}
\usepackage{ulem}

\begin{document}

\begin{center}

	{
        \strut
        \bfseries {\LARGE{	
        Resolution-Robust 3D MRI Reconstruction with 2D Diffusion Priors: Diverse-Resolution Training Outperforms Interpolation
	    }}
        \strut
    }
	
	\vspace*{.2in}
	
	{\large{
			\begin{tabular}{cccc}
				Anselm~Krainovic$^*$, Stefan~Ruschke$^\dagger$ and Reinhard~Heckel$^*$
			\end{tabular}
	}}
	
	\vspace*{.05in}
	
	\begin{tabular}{c}
	$^*$School of Computation, Information and Technology, Technical University of Munich \\ 
    $^\dagger$School of Medicine and Health, Technical University of Munich
	\end{tabular}

	\vspace*{.1in}
		
	\today
	
	\vspace*{.1in}
	
\end{center}

\begin{abstract}
Deep learning-based 3D imaging, in particular magnetic resonance imaging (MRI), is challenging because of limited availability of 3D training data. 
Therefore, 2D diffusion models trained on 2D slices are starting to be leveraged for 3D MRI reconstruction.
However, as we show in this paper, existing methods pertain to a fixed voxel size, and performance degrades when the voxel size is varied, as it is often the case in clinical practice. 
In this paper, we propose and study several approaches for resolution-robust 3D MRI reconstruction with 2D diffusion priors. 
As a result of this investigation, we obtain a simple resolution-robust variational 3D reconstruction approach based on diffusion-guided regularization of randomly sampled 2D slices. This method provides competitive reconstruction quality compared to posterior sampling baselines. 
Towards resolving the sensitivity to resolution-shifts, we investigate state-of-the-art model-based approaches including Gaussian splatting, neural representations, and infinite-dimensional diffusion models, as well as a simple data-centric approach of training the diffusion model on several resolutions. 
Our experiments demonstrate that the model-based approaches fail to close the performance gap in 3D MRI. In contrast, the data-centric approach of training the diffusion model on various resolutions effectively provides a resolution-robust method without compromising accuracy.
\end{abstract}

\section{Introduction}

Magnetic resonance imaging (MRI) is a widely used medical imaging technology. 
Since the MRI measurement process is long, it is typically accelerated by only collecting few measurements, and an algorithm is used to reconstruct an image or volume from the undersampled measurements. 

For 2D MRI, deep learning methods trained in a supervised fashion perform best 
on 2D MRI benchmarks~\citep{zbontar_FastMRIOpenDataset_2018}. 
However, deep learning-based 3D MRI reconstruction is much more challenging due to a lack of good 3D MRI training data and computational difficulties. 

Recently, diffusion models pre-trained on two-dimensional slices have been used successfully to perform 3D reconstruction for novel-view synthesis~\citep{wuReconFusion3DReconstruction2024} and for medical imaging~\citep{chungSolving3DInverse2023,leeImproving3DImaging2023} with very promising results. 

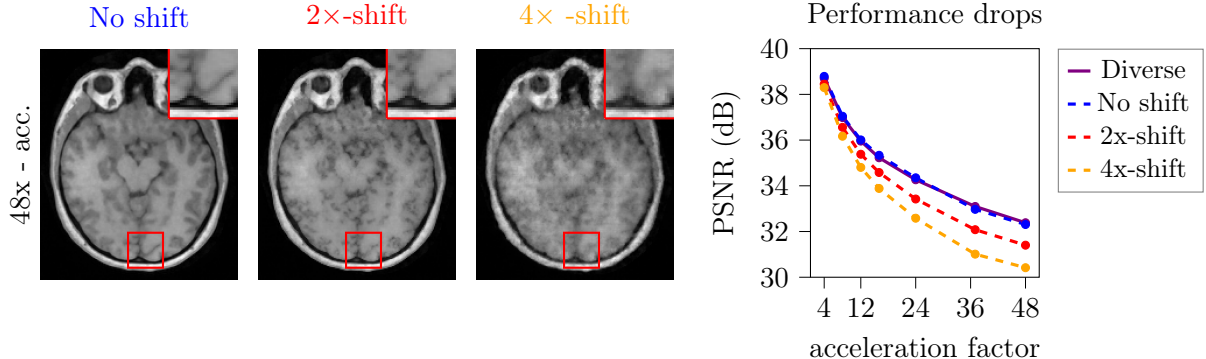
\begin{figure}
    \begin{centering}
        \input{figures_onecolumn/fig1_res_demo_48x_with_plot.tikz}
        \vspace{0.5cm}
    \end{centering}
        \vspace{0.7cm}
        \caption{
            \textbf{3D reconstruction with 2D diffusion priors is sensitive to resolution-shifts.}
            We reconstruct complex 3D volumes from undersampled multicoil 3D brain MRI measurements at a fixed voxel-size $\Vrecon$, and employ variational regularization with 2D diffusion-models pre-trained at voxel sizes with $\Vtrain = \Vrecon$ (no shift), $\Vtrain = 2 \Vrecon$ ($2 \times$-shift) and $\Vtrain = 4 \Vrecon$ ($4 \times$-shift).
            We observe that differences in voxel size lead to significant artifacts in the reconstructed images, both visually (left panel) and quantitatively (right panel).
            As we show in our paper, training on diverse resolutions is an effective solution.
        }
        \label{fig:main}
\end{figure}

However, the existing methods for MRI use diffusion models at a certain image resolution, and thus the resulting methods pertain to a fixed voxel size. 
As we show in this paper, performance degrades significantly when the reconstruction is performed at a different resolution than the model is trained for. This is an issue since in practice, the voxel size and thus resolution of the acquisition, varies significantly as it depends on clinical and subject-specific requirements.

In this paper, we propose and study several approaches for resolution-robust 3D MRI reconstruction with 2D diffusion priors. 
As a result, we present an efficient variational 3D diffusion-based reconstruction method that works well for variable voxel sizes, thus paving the way towards a practical 3D MRI imaging method based on 2D diffusion priors. 
We resolve the performance drop due to resolution shifts by training the diffusion model on slices with diverse resolutions, which effectively closes the performance gap induced by resolution-shifts, without trading off accuracy. 

Perhaps surprisingly, this simple data-centric approach towards resolution shifts performs best compared to a wide array of model-centric approaches. 
Specifically, we propose and compare to approaches based on modern continuous representations, in particular Gaussian splatting~\citep{kerbl3DGaussianSplatting2023b} and implicit neural representations~\citep{mildenhall_NeRFRepresentingScenes_2020}. 
These representations allow to resample the volume during reconstruction at the diffusion prior's resolution, while keeping the diffusion prior fixed. 

We also adapt recent works on infinite-resolution diffusion models~\citep{bond-taylorInftyDiffInfinite2023} and neural operator learning~\citep{kovachkiNeuralOperatorLearning2023}, which propose to interpolate the convolutional kernels of the model, instead, to match the resolution of the reconstructed volume. 

However, as our experiments on accelerated 3D MRI show, both volume and kernel interpolation methods are outperformed by the data-centric solution of diverse training, and fall short to close the drop of reconstruction performance induced by resolution-shifts.

Our experiments are carried out using 3D fully-sampled 3T knee volumes~\citep{eppersonCreationFullySampled2013}, a 3T 3D brain MRI dataset~\citep{souzaCC3592018} and a high field (7T) 3D brain MRI dataset~\citep{alkemadeAmsterdamUltrahighField2020}. Some key results on the 3T brain dataset are depicted in Figure~\ref{fig:main}. 

\section{Problem statement: Resolution-robust 3D MRI reconstruction}
\label{sec:problem_statement}
In this section, we describe the 3D accelerated MRI reconstruction problem we consider in this paper, and describe the issues faced by current diffusion-based reconstruction methods when the voxel size varies.

We consider multicoil-accelerated 3D MRI reconstruction, where the goal is to reconstruct a 3D complex-valued volume $\vx \in \mathbb{C}^{W \times H \times D}$ based on measuring the nuclear magnetic resonance signals emitted by the object, via their induced currents in $C$ receiver coils. The MRI measurements $\vy_i$ are given by
\begin{equation}
    \vy_i = \mM \mFourier{3} \mS_i \vx + \vz_i \in \mathbb{C}^m, \quad i = 1, \dots, C.
    \label{eq:inverse_problem_mri}
\end{equation}
Here, $\mS_i$ encodes the sensitivity map associated with the $i$-th receiver coil, $\mFourier{3}$ is the 3D discrete Fourier transform, and $\vz_i$ is noise.
Moreover, $\mM$ is an undersampling mask, which is the widely adopted 2D Poisson sampling mask or a 2D Gaussian sampling mask in our experiments~\citep{lustigL1Spirit2007}. 
The mask models which parts of the Fourier space are measured, and implements an acceleration of the MRI acquisition by the factor $R = N / m$, where $N = WHD$ is the number of voxels. The acceleration factor ranges from $4$ to $48$ in our experiments.

The voxel size of the reconstructed volume $\vx$ is $w \times h \times d$, 
with values usually ranging from $0.25$mm to $2$mm in practice.
Those values are a setting of the scanner, and are changed across scans, in particular for 3D scans. Most existing reconstruction methods in the literature assume the voxel size to be fixed. However, if a method is trained or tuned for one voxel size, and applied to another, performance drops. In this work, our goal is to develop a method that robustly reconstructs a 3D volume for different voxel sizes.

As an example of how different voxel sizes are in practice; all volumes of the widely used fastMRI knee dataset have the same fixed in-plane resolution ($0.5 \text{mm} \times 0.5 \text{mm})$~\citep{zbontar_FastMRIOpenDataset_2018}, and 
the fastMRI prostate dataset has a similar resolution for the T2 weighted images ($0.56 \text{mm} \times 0.56\text{mm}$), but an approximately $4$ times larger in-plane resolution for the diffusion weighted images ($2.0 \text{mm} \times 2.0 \text{mm}$)~\citep{tibrewalaFastMRIProstatePublicly2023}, indicating the large range of resolutions used in practice for the same anatomy.

\section{Background on diffusion-model based reconstruction for MRI}
\label{sec:background}

Supervised deep learning-based methods are very successful for 2D MRI, but they are not naively applicable to 3D imaging due to a lack of 3D training data. 
For 2D imaging, the best-performing methods are trained in a supervised fashion to reconstruct an image from an undersampled measurement. The data for generating the target images required for supervised training requires the acquisition of fully-sampled data. Acquiring fully-sampled 3D data, in particular high-resolution data, would be very time-consuming and expensive and complicated by the fact that long scan times introduce motion artifacts. 
A promising 3D imaging approach that circumvents these difficulties relies on pre-trained 2D diffusion models trained on slices (or projections) of 3D data. 

\paragraph{Diffusion models.}
Diffusion models are probabilistic models that learn a data distribution $p(\vx)$ of complex 2D MRI images in our setup~\citep{sohl-dicksteinDeepUnsupervisedLearning2015} and exhibit high sample quality~\citep{hoDenoisingDiffusionProbabilistic2020,nicholImprovedDenoisingDiffusion2021}. 
Diffusion models consist of a forward diffusion process that gradually adds Gaussian noise to the input, and a reverse denoising process that is aimed at inverting the diffusion process.
In this paper, we adopt the denoising diffusion probabilistic models (DDPM) formulation by~\citet{hoDenoisingDiffusionProbabilistic2020}, in which the forward process is described by a Gaussian kernel $p(\vx_{t+1} \mid \vx_t) = \mathcal{N}(\vx_{t+1}; \sqrt{1-\beta_t} \vx, \beta_t \mI)$, with increasing noise levels $\beta_t$. 
To learn the reverse diffusion process, \citet{hoDenoisingDiffusionProbabilistic2020} approximate the reverse kernel $p(\vx_{t-1} \mid \vx_t, \vx_0)$ by a kernel $p_{\phi}(\vx_{t-1} \mid \vx_t)$, parameterized by a neural network $\vep_{\vphi}(\vx_t, t)$, and obtain the training objective 
$\mathcal{L}(\phi, \vx) = \EX[t \sim \mathcal{U}(0,T), \vep \sim \mathcal{N}(0, \mI)]{
        \norm[2]{
                \vep_{\vphi} ( \sqrt{1 - \sigma_t^2} \vx + \sigma_t \vep; t ) - \vep
            }^2
    }$,
by optimizing a variant of the evidence lower bound. Here, $\sigma_t^2 = 1 - \prod_{s=1}^t (1 - \beta_s)$. 
This training objective is similarly obtained when interpreting the networks $\vep_{\vphi}(\vx_t, t)$ as learning (an approximation of) the score-functions $\nabla_{\vx_t} \log p(\vx_t)$. 

\paragraph{Posterior-sampling based 2D imaging.} 
Consider an inverse problem, where a measurement $\vy$ is obtained as $\vy = \mA \vx_0 + \ve$ where $\mA$ is the known forward map and $\ve$ is Gaussian noise. 
If we assume that $\vx_0$ is drawn from a prior distribution, then 
solving the inverse problem consists of sampling from the posterior distribution $p(\vx_0 \mid \vy)$. Diffusion models enable posterior sampling as follows. 

Diffusion models provide access to (an approximation of) the score function $\nabla \log p(\vx_t)$ of the iterates $\vx_t$ of the diffusion process. 
The score function of the conditional density $p(\vx_t \mid \vy)$ is $\nabla \log p(\vx_t \mid \vy) = \nabla \log p(\vy \mid \vx_t) + \nabla \log p(\vx_t)$, where the second term is the score of the unconditionally trained diffusion model.
The likelihood term $p(\vy \mid \vx_t)$, however, requires further approximation, since it is tractable for $t = 0$ only.
It holds that $p(\vy \mid \vx_t) = \int p(\vy \mid \vx_0) p(\vx_0 \mid \vx_t) d \vx_0$. While $p(\vy \mid \vx_0)$ is given by the inverse problem, the posterior $p(\vx_0 \mid \vx_t)$ can be highly multimodal.
Existing posterior sampling methods circumvent this by making different assumptions, which in some form lead to tractable likelihood term $p(\vy \mid \vx_t) \approx \mathcal{N}(\vy; g(\vx_0), r_t \mI)$~\citep{mardaniVariationalPerspectiveSolving2023a}. 

As an example, \citet{song2023pseudoinverseguided} approximate the conditional density $p(\vx_0 \mid \vx_t)$ with a Gaussian $\mathcal{N}(\hat{\vx}_t, \zeta_t^2 \mI)$, where $\hat{\vx}_t = \EX{\vx_0 \mid \vx_t}$, and demonstrate good results for accelerated 2D MRI.
Similarly, \citet{chung2023diffusionDPS} propose denoising posterior sampling (DPS) based on the approximation $p(\vy \mid \vx_t) \approx p(\vy \mid \hat{\vx}_t)$, and demonstrated improved results over measurement-subspace projection methods~\citep{chungScorebasedDiffusionModels2022a, songSolvingInverseProblems2022}. 
Finally, \citet{mardaniVariationalPerspectiveSolving2023a} and \citet{ozturklerRegularizationDenoisingDiffusion2023} presented good results on 2D MRI imaging by using a variational approach, which we will further discuss in Section~\ref{sec:resolution_robust_imaging}.

\paragraph{Posterior sampling-based 3D imaging}
Current 3D MRI reconstruction methods with diffusion priors are based on posterior sampling, and were initially intertwined with classical priors.

\citet{chungSolving3DInverse2023} proposed DiffusionMBIR, which performs slice-wise 2D sampling steps, followed by increasing 3D consistency via a classical total-variation (TV) prior. 
Subsequently, \citet{chungDecomposedDiffusionSampler2023} improved upon DiffusionMBIR by employing the more efficient decomposed diffusion sampler (DDS) for 2D sampling, while retaining the TV prior. 
The following two works, however, demonstrated that data-driven regularization outperforms classical priors for 3D consistency. 

Specifically, \citet{leeImproving3DImaging2023} proposed to use two perpendicular diffusion models (TPDM) for regularization, and obtained improved results on 3D CT reconstruction compared to DiffusionMBIR.
But, TPDM requires expensive calculations of the network Jacobian. We use TPDM as a baseline in our experiments, but employ the more efficient DDS sampler~\citep{chungDecomposedDiffusionSampler2023}. 
Recently, \citet{songDiffusionBlendLearning3D2024} proposed training a diffusion model on small stacks of neighboring 2D slices, outperforming TPDM and DiffusionMBIR on 3D CT reconstruction.
However, obtaining high-resolution 3D volumes for training is infeasible for 3D MRI in practice, as we discuss in Section~\ref{sec:background}. 

All previous methods pertain to a specific voxel size.
\citet{arefeenINFusionDiffusionRegularized2024} recently investigated using implicit neural representations (INN) for 3D MRI reconstruction, but do not discuss resolution-shifts.  
Moreover, the approach proposed by \citet{arefeenINFusionDiffusionRegularized2024} is also limited to a fixed voxel size, since only two dimensions are encoded continuously with the INN for computational reasons.

The 3D imaging method proposed in this paper is different to these prior works, in that we take a variational approach to 3D reconstruction, regularize on randomly sampled slices along all directions, and importantly make the method applicable to varying voxel sizes.

\section{Proposed method: Resolution-robust 3D imaging with 2D priors} 
\label{sec:resolution_robust_imaging}

In this section, we discuss our proposed approach for resolution-robust 3D imaging. Our approach is based on variational regularization with 2D diffusion priors. To achieve resolution-robustness, we consider training a diverse 2D prior with a proposed augmentation method as well as several interpolation methods.

\begin{figure}
    \centering
    \includegraphics[width=0.8\textwidth]{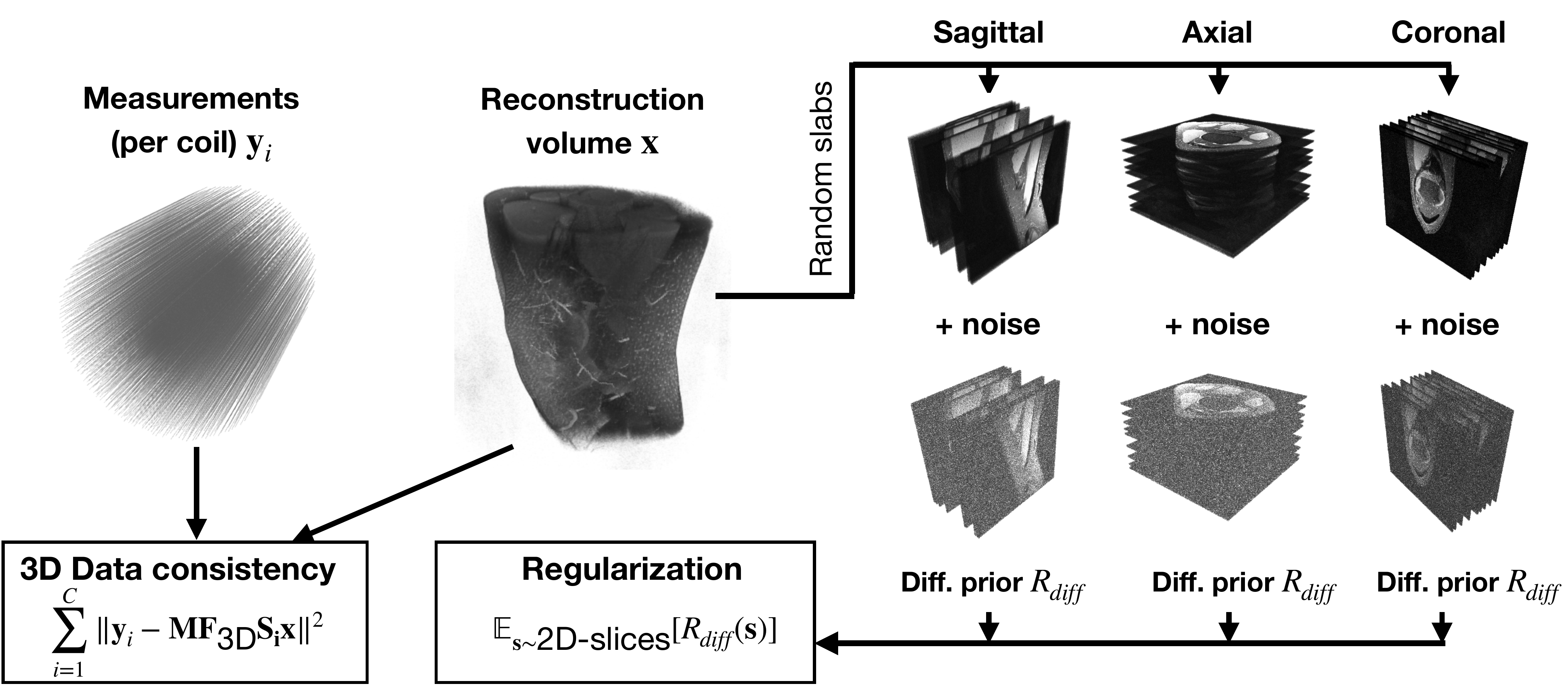}
    \vspace{0.5cm}
    \caption{\textbf{Illustration of the proposed variational 3D MRI reconstruction method}, which in each iteration regularizes randomly sampled slices with a diversely trained 2D diffusion prior.}
    \label{fig:mri_3d_recon_visualization}
\end{figure}

We propose a variational approach for reconstructing the 3D volume. The variational approach enables us to integrate interpolation methods (described in Section~\ref{sec:resrobust}), and performs well compared to posterior sampling baselines. The method is  illustrated in Figure~\ref{fig:mri_3d_recon_visualization}.

\subsection{Variational 3D MRI reconstruction}

We reconstruct the unknown 3D complex volume $\vx \in \mathbb{C}^{W\times H \times D}$  by solving the optimization problem:
\begin{align}
    \begin{split}
        \hat{\vx} (\vy) = \underset{\vx \in \mathbb{C}^N}{\argmin} \sum_{i=1}^C \norm[2]{\vy_i - \mM \mFourier{3} \mS_i \vx}^2
        + \lambda \EX[\mathbf{s} \sim \text{2D-Slices}(\vx)]{\mathcal{R}_{\text{diff}}(\mathbf{s})},
   \end{split}
    \label{eqn:var_method_optimization_problem}
\end{align}
where $\mathcal{R}_{\text{diff}}(\mathbf{s})$ is a regularizer based on the pre-trained 2D complex diffusion model $\vep_{\vphi}(\vx_t, t)$, and $\lambda$ is a hyperparameter. 
We scale the measurements $\vy_i$'s such that the reconstruction has approximate variance $1$, similar to the training data of the diffusion model. %
We solve the optimization problem~\eqref{eqn:var_method_optimization_problem} using an iterative first-order method (Adam), where in each iteration $t$ we approximate the expectation in~\eqref{eqn:var_method_optimization_problem} by applying a randomized slicing scheme. 

This variational approach performs very well in terms of reconstruction quality and speed, as shown by our experimental results  in Section~\ref{sec:experiments}.
Moreover, the approach allows to change the forward operator, due to its variational formulation.
This is of practical importance, since sampling patterns and coil sensitivity maps are specific to the MRI imaging device.

We next describe the data consistency term, regularization objective, and randomized slicing in more detail. In the next section, we discuss our considered approaches towards resolution robustness.

\paragraph{3D data consistency.}
We use the full 3D forward map in the data-consistency term. 
Previous works on 3D diffusion-based MRI split the data-consistency term into many-dimensional subproblems, allowing slices to be processed in small batches~\citep{chungSolving3DInverse2023,leeImproving3DImaging2023}.
However, splitting 3D MRI into 2D subproblems assumes a hybrid-Cartesian 3D sampling pattern $\mM$, in which at least one direction (readout) is fully-sampled Cartesian. 
While many commercial scanners and methods are optimized for these patterns, non-Cartesian patterns, such as Stack-of-Stars, provide improved robustness to motion-corruption during acquisition~\citep{blockRoutineClinicalUse2014}.
In our method, we can circumvent computational limitations, since we do not require computation of the diffusion model's Jacobian.

\paragraph{Regularizer.}
Our regularizer relies on
a re-weighted form of the residual denoiser training loss of a 2D-diffusion model:
\begin{equation}
    \mathcal{R}_{\text{diff}}(\mathbf{s}) =\EX[t \sim \mathcal{U}(0, T'), \vep \sim \mathcal{N}(0, \mI)]{w(t) \norm[2]{\vep_{\vphi}(\mathbf{s}_t; t) - \vep}^2}.
    \label{eqn:reg_objective_R_diff}
\end{equation}
Here, $\mathbf{s}_t = \sqrt{1 - \sigma_t^2} \mathbf{s}+ \sigma_t \vep$ is a noisy version of the slice $\mathbf{s}$, defined according to the diffusion forward process.
This is the standard training loss of a 2D-diffusion model.  \citet{mardaniVariationalPerspectiveSolving2023a} derive this loss (with a slightly different scaling $w(t)$) by variationally fitting a Gaussian to the diffusion-model based posterior $\min_{\mu, \sigma} KL(\mathcal{N}(\mu, \sigma^2 \mI) ~||~  p(\vx \mid \vy))$. 

Directly computing gradients of the regularization term $\mathcal{R}_{\text{diff}}$ for optimizing the loss in equation~\eqref{eqn:var_method_optimization_problem}, turns out to be infeasible, since it involves the Jacobian $\mJ_{\vz} := \frac{\partial \vep_{\vphi}(\vz, t)}{\partial \vz}$ of the diffusion model:
\begin{equation*}
    \nabla_{\mathbf{s}}{\norm[2]{\vep_{\vphi}(\mathbf{s}_t, t) - \vep}^2}
    = 2 \sqrt{1 - \sigma_t^2} \transp{\left( \vep_{\vphi}(\mathbf{s}_t, t) - \vep \right)} \mJ_{\mathbf{s}_t}.
\end{equation*}
The diffusion model network's Jacobian is expensive to compute and can be poorly conditioned for small noise levels, which leads to instabilities during training. Similar to~\citet{pooleDreamFusionTextto3DUsing2022}, we approximate the Jacobian as $\mJ_{\vz} \approx \mI$ and observe that it increases the stability and efficiency of variational reconstruction with only small losses in performance. Omitting the term is further justified by~\citet{pooleDreamFusionTextto3DUsing2022} by deriving it from density distillation losses, and by~\citet{mardaniVariationalPerspectiveSolving2023a} by employing results from stochastic process theory and assuming zero variance in the variational distribution. 

To approximate the expectation in equation~\eqref{eqn:reg_objective_R_diff}, 
the time step $t$ is sampled uniformly from $\{0, 1, \dots, T' \}$, where $1 \leq T' < T$. We find that choosing $T' \approx 0.4 \cdot T$ outperforms sampling from the full set of timesteps (see Figure~\ref{fig:appendix_sweep_diffprior_T} of the appendix). 
Finally, we choose the weighting factor as $w(t) = \frac{\sigma_t}{\alpha_t}$, such that at each randomly sampled timestep $t$, one expects a similar noise energy as the noise energy of the optimal MMSE denoiser (see \citep{mardaniVariationalPerspectiveSolving2023a} for justification).

\paragraph{Randomized slicing.}
A key ingredient of our proposed approach is randomized slicing, which is performed along the anatomical planes (coronal, sagittal, and axial) of the volume $\vx$.
Specifically, at each iteration $t$ of solving the reconstruction problem~\eqref{eqn:var_method_optimization_problem} with a first-order gradient method, we extract $S$ random 2D slices from each slice orientation of the volume $\vx_t$ ($3 S$ slices per iteration). 
We experimented with various sampling schemes, and observed that uniformly sampling relatively small slabs (i.e., multiple slices right next to each other)  provides good performance.
We treat the number of slabs and their size as hyperparameters optimized over a validation dataset.
The number of slices, $S$, needs to be sufficiently large, but increasing this parameter beyond $S=50$ in our setups does not provide a benefit (see Figure~\ref{fig:appendix_slice_budget} in the appendix). 

\subsection{Achieving resolution robustness}
\label{sec:resrobust}

The vast majority of the existing diffusion based reconstruction methods are trained using datasets that contain no or only a small variation in the image/voxel resolution and are sensitive to changes in the resolution/voxel size. 
This can be seen from the experiments in Section~\ref{sec:experiments} and Figure~\ref{fig:main} which shows that performance drops when the voxel-sizes during training is larger or smaller than at reconstruction.

For our variational optimization based method (equation ~\eqref{eqn:var_method_optimization_problem}), the voxel-size $\Vrecon$ of the reconstructed volume $\vx$ is given by the measurement's resolution. 
For evaluating the regularization term $\mathcal{R}_{\text{diff}}(\mathbf{s})$, the slices $\mathbf{s}$ of the reconstructed volume $\vx$ therefore have in-plane pixel-size $\Vrecon$, which is often different to the pixel-size $\Vtrain$ of the training data.

Towards establishing a resolution-robust 3D reconstruction method, we study  the following approaches: The data-centric approach of training on diverse resolutions,  interpolating the MRI volume, and using infinite-resolution diffusion models. 

\subsubsection*{Approach 1: Diverse training}
First, we consider augmenting the dataset with volumes of various resolutions. Since common training datasets for MRI consist of volumes at a fixed resolution per anatomy and weighting (e.g., the fastMRI dataset), 
we simulate various resolutions by performing trilinear downsampling of high-resolution 3D volumes, while keeping the field of view fixed. 

\subsubsection*{Approach 2: Volume interpolation methods.}
Second, we consider using continuous representations including implicit neural networks and Gaussian splatting to mitigate the sensitivity of the variational approach to resolution-shifts. Given a continuous representation of a volume $\vc_{\vtheta} \colon [-1,1]^3 \to \mathbb{C}$, parameterized by $\vtheta \in \reals^p$, we reformulate the variational objective in equation~\eqref{eqn:var_method_optimization_problem} as:
\begin{align}
    \begin{split}
        \underset{\vtheta}{\argmin} \sum_{i=1}^C \norm[2]{\vy_i - \mM \mFourier{3} \mS_i \vc_{\vtheta}(\vMesh{3})}^2 + \lambda \EX[\vMesh{2} \sim \text{2D-Meshes}]{\mathcal{R}_{\text{diff}}(\vc_{\vtheta}(\vMesh{2}))}.
   \end{split}
    \label{eqn:interpolated_optimization_problem}
\end{align}
Here, $\vMesh{3}$ is a three-dimensional mesh, describing $W \cdot H \cdot D$ locations in the cube $[-1,1]^3$ at which the measurements are taken at voxel-size $\Vrecon$, and the discrete 3D Fourier transform is applied.
Moreover, $\vMesh{2}$ is a randomly sampled two-dimensional mesh, encoding a random view with matrix size $X \times Y$, for applying the 2D pre-trained diffusion prior.
We keep the field of view constant, and sample views of pixel size $\Vtrain$ by choosing the matrix size $X \times Y$ of the 2D meshes $\vMesh{2}$ accordingly.

\paragraph{Approach 2.1. Grid-resampling interpolation.} We parameterize the volume with a high-resolution grid $\vtheta \in \mathbb{C}^{W \times H \times D}$ and apply trilinear or nearest neighbor interpolation to obtain the values at the mesh-encoded locations.

\paragraph{Approach 2.2. Implicit neural networks.} 
We consider INNs since they work well for novel-view synthesis and medical imaging tasks. We choose a standard architecture with a Fourier feature input layer followed by linear layers with normalization, similar to~\citet{tancik_FourierFeaturesLet_2020a}.
As discussed in Section~\ref{sec:experiments}, fitting INNs is computationally very expensive. 
\paragraph{Approach 2.3. Gaussian interpolation}
As a fast and powerful continuous representation, we consider Gaussian splatting, originally proposed by \citet{kerbl3DGaussianSplatting2023b} for solving the volumetric rendering problem in novel-view synthesis. 
Gaussian splatting consists of learning a continuous coordinate-based representation $\vc_{\theta}$, which is given as a weighted sum of an-isotropic Gaussians. 
We adapt it to complex MRI data by choosing
    $c_{\vtheta}(\vx) = \sum_{k=1}^G (a_k+ i b_k) \exp \left(-\frac{1}{2} (\vx - \vmu_k)^T \mSigma_k^{-1} (\vx - \vmu_k) \right)$ 
where $a_k$ and $b_k$ model the real and imaginary parts of the Gaussian, and $\vmu_k$ parameterizes the center the Gaussian.
The spread and orientation of the Gaussian are given by decomposing the positive definite covariance matrix as $\mSigma_k = \mR \mS \transp{\mS} \transp{\mR}$, with a diagonal scaling matrix $\mS = \diag(s_1, s_2, s_3)$. 
The optimization problem~\eqref{eqn:interpolated_optimization_problem} requires differentiable rasterization of the Gaussian representation, which entails evaluating the Gaussian representation at given meshes, and calculating associated gradients for optimization.
We adapt the CUDA kernel for 2D-tile-based rasterization by~\citet{kerbl3DGaussianSplatting2023b} to perform efficient 3D complex volume rasterization instead of volumetric rendering. For initializing the Gaussians, we make use of the reconstruction obtained using the pseudoinverse.

\subsubsection*{Approach 3: Infinite-resolution diffusion models.}
Recently, infinite-dimensional diffusion models have been proposed to generate images at arbitrary resolutions~\citep{hagemannMultilevelDiffusionInfinite2023, dupontGenerativeModelsDistributions2022, bond-taylorInftyDiffInfinite2023}. The goal is to learn distributions over functions, which can encode resolution-independent representations of images.
To achieve that, several works suggest learning the distribution over the parameters of (conditional) implicit neural representations of the data~\citep{hagemannMultilevelDiffusionInfinite2023, dupontGenerativeModelsDistributions2022}.
However, currently these methods do not scale well with resolution, and thus the training data's resolution is often limited to $64 \times 64$~\citep{bond-taylorInftyDiffInfinite2023}.

As an alternative to the neural fields approaches, \citet{bond-taylorInftyDiffInfinite2023} propose to use a multiscale U-Net architecture and demonstrated good sample quality on high-resolution image generation. 
To generalize beyond the fixed training resolution, \citet{bond-taylorInftyDiffInfinite2023} propose to employ depthwise convolutional layers as part of the U-Net, and to interpolate the kernels when the input data's resolution changes.

Interpolating depthwise convolutional kernels can be motivated as follows: If we consider the discretization-independent image to be represented by a continuous function, we aim to perform a parameterized convolution in function space
$\int_{[-1,1]^2} k_{\theta} (\vy - \vx) f(\vx) d\vx$.
Here, $f$ is the image function and $k_{\vtheta}$ is the parameterized kernel function. In practice, the kernels values $k_{\vtheta}(\vx_i)$ are learned on grid points, determined by the discretization of the training data. When applied to an image with a different resolution during reconstruction, there is a mismatch between the discretization of the kernel and the image data. 

\paragraph{Approach 3.1. Bilinear interpolation.}
Following \citet{bond-taylorInftyDiffInfinite2023}, we investigate using bilinear interpolation to adapt the learned kernel to match the discretization of the input data.

\paragraph{Approach 3.2. Zero-padding in Fourier space.} We further investigate zero-padding in Fourier space, which is employed in many neural operator learning methods, such as the Fourier neural operator (FNO)~\citep{liFourierNeuralOperator2020, kovachkiNeuralOperatorLearning2023}.

\paragraph{Approach 3.3. Zero-padding in image space.} Finally, we compare to zero-padding in image-space. It is the simplest method, but does not yield a robust method in our experiments. 

\section{Experiments}
\label{sec:experiments}
In this section, we study our method for robust 3D MRI reconstruction with 2D diffusion priors. 

\paragraph{Datasets.}
We use the Stanford~3D MRI datasets for our experiments on 3D knee MRI, which consists of $19$ fully sampled k-space volumes, acquired using $C = 8$ receiver coils in parallel with a 3D fast spin echo protocol (CUBE) with proton density weighting included fat saturation~\citep{eppersonCreationFullySampled2013}. The voxel-size of the knee volumes is $0.6 \text{mm} \times 0.5 \text{mm} \times 0.5 \text{mm}$, and we split the dataset into $13$ volumes for training, $2$ for validation, and $4$ for testing. 

Moreover, we use the 3D brain MRI acquisitions from the Calgary-Campinas dataset~\citep{souzaCC3592018}, which have been gradient-recalled echo (GRE) protocol with T1 weighting and $C=12$ coils. For all volumes $85 \%$ of the k-space has been acquired, with an isotropic voxel-size of $1 \text{mm}$. We use $40$ volumes for training, $7$ for validation $20$ for testing.

Finally, we perform experiments on the ultra-high field 7T AHEAD brain MRI dataset~\citep{alkemadeAmsterdamUltrahighField2020}. The volumes from $77$ participants are acquired using a gradient echo MP2RAGEME sequence with $C=32$ coils, and at an isotropic voxel size of $0.7 \text{mm}$. We consider the volumes acquired with T1 weighting and split the volumes into $63$ for training, $6$ for validation, $8$ for testing.

For the MRI reconstruction problem, we use a 2D Poisson mask for the high-field knee and brain volumes, and a 2D Gaussian sampling mask for the AHEAD dataset.

\paragraph{Calibration and MVUE reference.} We estimate the sensitivity maps $\mS_i$ from the fully sampled k-space measurements $\vy_i$ of the dataset with the ESPIRiT algorithm~\cite{ueckerESPIRiTEigenvalueApproach2014} implemented in the BART toolbox~\cite{ueckerBARTVersion2014}.
We take the minimum-variance unbiased estimate (MVUE) computed based on the entire measurements $\vy_i$ and estimated sensitivity maps $\mS_i$ as reference for training and evaluation. 

\paragraph{Training of 2D diffusion priors.}
We use the U-Net architecture for the diffusion model $\vep_{\vphi}(\vx_t, t)$ with the implementation adopted from~\citet{dhariwalDiffusionModelsBeat2021a}.  
The model's blocks consist of convolutional layers with timestep embedding and skip connections, and includes attention blocks at lower levels of the hierarchy.
For the infinite-resolution diffusion models experiments we follow~\citet{bond-taylorInftyDiffInfinite2023} and employ depthwise convolutional kernels in the first and last residual block of the architecture. 

We train the diffusion models on complex 2D slices taken from the MVUE reference volumes, at all three anatomical planes (coronal, sagittal and axial). We treat imaginary and real parts as two color channels, similar to~\citet{chungDecomposedDiffusionSampler2023}.

For training the diffusion models we follow the DDPM setup of~\citet{hoDenoisingDiffusionProbabilistic2020}, see Appendix~\ref{sec:appendix_details_exp_results}. 

\subsection{3D reconstruction without resolution-shifts}
We first investigate the performance of the variational approach without resolution shifts, and compare it to posterior sampling and classical baselines, and find the variational approach to perform best. 
For variational and posterior sampling methods we use the same 2D diffusion model for a fair comparison.

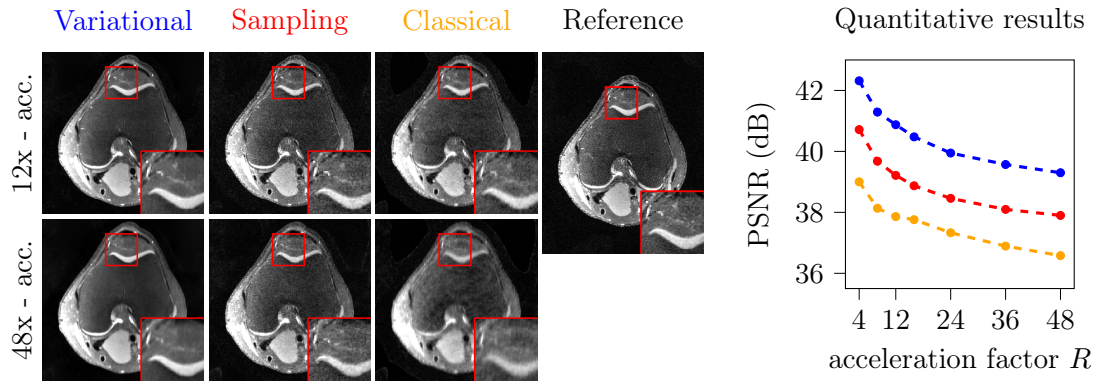
\begin{figure}
    \begin{centering}
        \hspace{0.5cm}
        \input{figures_onecolumn/3d_reconstruction_comparison_2x3.tikz} 
        \vspace{0.5cm}
    \end{centering}
    
        \caption{
            \textbf{Diffusion-based and compressive sensing reconstruction of 3D MRI volumes.} The right panel shows the PSNR of the reconstructions obtained from performing 3D reconstruction from undersampled multicoil 3D MRI measurements.
            It can be seen that the proposed variational method yields superior reconstruction quality, followed by posterior sampling and the compressive sensing baselines. Left panel: Slices from the reconstructed volumes demonstrate that the variational approach provides competitive visual quality and smoother reconstructions.
        }
        \label{fig:recon_methods_std_performance_comparison}
\end{figure}

\paragraph{Compressed sensing baseline.} As a classical compressive sensing baseline, we use $L_1$-Wavelet regularized least-squares~\citep{lustigSparseMRIApplication2007}. 

\paragraph{Posterior sampling baseline.} We compare to TPDM~\citep{leeImproving3DImaging2023} as the 3D posterior sampling baseline, but adapt it by incorporating recent improvements for posterior sampling methods.
TPDM performs slice-wise denoising posterior sampling (DPS), with an unconditional sampling step every $K$ iterations. \citet{leeImproving3DImaging2023} performs DPS using a primary diffusion model, and unconditional sampling using an auxiliary model. The diffusion models are trained on single, but different anatomical views.
We adapt TPDM as follows. First, to obtain a reasonable reconstruction speed, comparable to our variational method, we use decomposed diffusion sampling (DDS)~\citep{chungDecomposedDiffusionSampler2023}. Second, we train a single model on data from multiple anatomical planes, and cycle between applying DDS slice-wise along each axis.

\paragraph{Quantitative results.} We reconstruct test volumes of the Stanford 3D dataset, and retrospectively implement various acceleration factors by varying the 2D Poisson disk undersampling mask.
The variational approach outperforms posterior sampling and classical baselines both in terms of PSNR (see Figure~\ref{fig:recon_methods_std_performance_comparison}) and in SSIM (see Figure~\ref{fig:appendix_ssim_sensitivities} in the appendix).
 
\paragraph{Visual results.}
By visually inspecting slices of the reconstructed volumes, we see that variational reconstruction tends to produce smoother reconstructions than posterior sampling and compressed sensing. 
Further, reconstructions obtained via posterior sampling exhibit higher pixel noise and are visually less faithful. This can be observed given the accentuated line structures at $48 \times$ accelerated reconstructions (similar to $L_1$ reconstruction).

 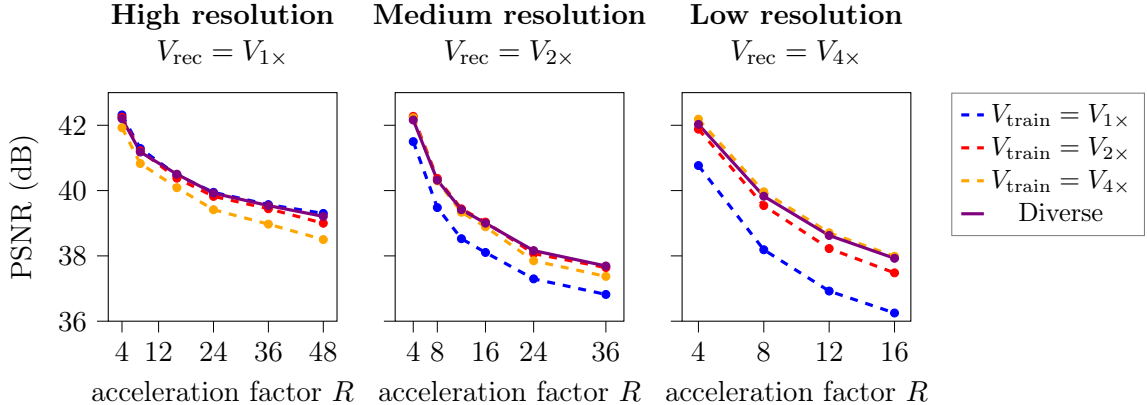
\begin{figure}
    \centering
    \input{figures_onecolumn/resolution_shift_demo_ext.tikz}
    \vspace{0.5cm}
    \caption{
        \textbf{Reconstruction under resolution shifts}.  
        3D reconstruction on the Stanford 3D knee dataset at small, medium, and large voxel-sizes $\Vrecon$ with diffusion models trained at fixed or diverse voxel-sizes $\Vtrain$. 
        Performance drops when the test-voxel size is different to the train-voxel size, with larger gaps for higher accelerations and larger resolution-shifts.
        Training the model on diverse resolutions provides a robust reconstruction method across resolutions.
        }
    \label{fig:resolution_shifts_cross_stanford_3d}
\end{figure}

\subsection{Performance drop due to resolution-shifts}

3D reconstruction with 2D diffusion models performs very well on in-distribution data, as demonstrated in the previous section. 
We now demonstrate that the performance drops when the voxel-size of the training data is different to the measurement's voxel-size. 
We first investigate the robustness to resolution-shifts of the variational approach. 

    We first consider the Calgary-Campinas 3D brain dataset with isotropic voxel-size $V_{1\times} = 1.0$mm, keep the field of view fixed, and generate datasets with coarser resolution via downsampling.
    We obtain volumes at resolutions $V_{2\times} = 2.0 \text{mm}$ and $V_{4\times} = 4.0 \text{mm}$, and train diffusion models for each of those datasets, respectively. We now apply the diffusion model trained at voxel-size $\Vtrain = V_{1\times},V_{2\times},V_{4\times}$ to measurements with voxel-size $\Vrecon = V_{1\times}$.
    As Figure~\ref{fig:main} shows, the performance  drops significantly for larger shifts and higher acceleration factors. The example reconstruction on the left panel illustrate that the resolution shifts induces high-frequent noise-like artifacts.
    In particular, fine-grained structures are not reconstructed well under a resolution-shift.

    We further consider the Stanford 3D knee dataset with voxel size $V_{1 \times} = 0.6  \text{mm} \times 0.5  \text{mm} \times 0.5  \text{mm}$.
    We follow the same procedure as before, but  also reconstruct at larger voxel-sizes $\Vrecon = V_{2\times}$ and $\Vrecon = V_{4\times}$. These cross-evaluation results are depicted in Figure~\ref{fig:resolution_shifts_cross_stanford_3d}, and confirm that none of the diffusion models trained at a fixed resolution provides a robust 3D reconstruction method when evaluated at a larger or smaller voxel-size than during training.
    
    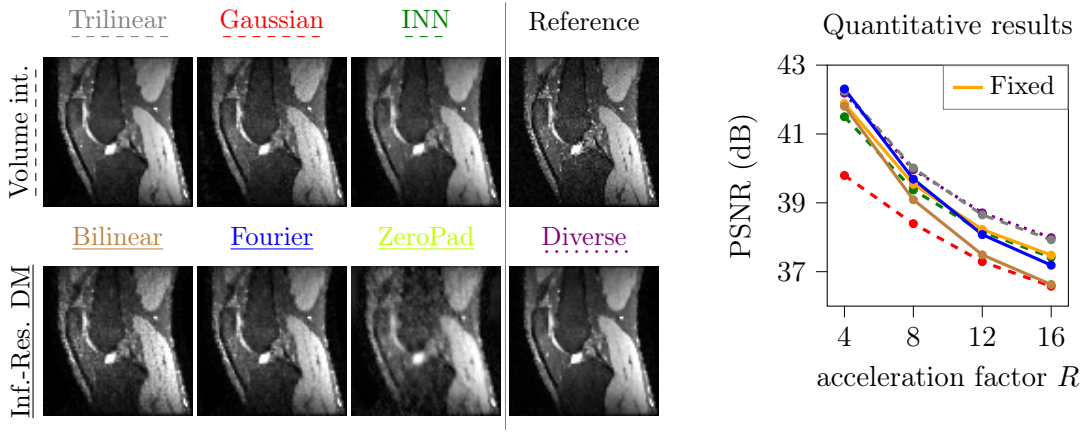
\begin{figure}
        \begin{centering}
            \hspace{0.8cm}
            \input{figures_onecolumn/recon_at_lower_resolution_ext.tikz}
            \vspace{0.5cm}
        \end{centering}
            \caption{
            \textbf{Reconstruction at larger voxel-sizes ($\Vrecon > \Vtrain$).}
            Effectiveness of several volume and kernel interpolation methods, where the diffusion models are trained on $\Vtrain = V_{2 \times}$, and we reconstruct at $\Vrecon = V_{4 \times}$. 
            We observe that simple trilinear resampling mitigates the performance drop completely. 
        }
        \label{fig:resolution_interpolation_lower}
 \end{figure}
    
    We performed similar experiments using the 3D posterior sampling baseline, and observe a comparable drop in performance (see Figure~\ref{fig:appendix_shift_demo_sampling} in the appendix).

\subsection{Achieving resolution robustness: Evaluating Approaches 1-3}

    We now study the approaches introduced in Section~\ref{sec:resrobust} and find training on diverse resolutions (Approach 1) to perform best.

    \paragraph{Approach 1: Diverse training.} 
    Approach 1 generates the training data via trilinear downsampling.
    Specifically, we take full-sized volumes, keep the field of view constant, and change the resolution by a factor sampled uniformly from $[0.1, 1.0]$. 
    We then train a diffusion model with the same architecture and training setup, and perform 3D reconstruction with this 2D diffusion prior. 
    The results in Figures~\ref{fig:main} and~\ref{fig:resolution_shifts_cross_stanford_3d} show that training on diverse resolutions effectively closes the performance-gap, without notably compromising standard accuracy.

\begin{figure}
    \begin{centering}
        \hspace{0.8cm}
        \input{figures_onecolumn/recon_at_higher_resolution_ext.tikz}
        \vspace{0.5cm}
    \end{centering}
        \caption{
            \textbf{Reconstruction at smaller voxel-sizes ($\Vrecon < \Vtrain$).} 
            The left panel shows example reconstructions and the right panel numerical results for reconstructing measurement at high-resolution $\Vrecon = V_{1 \times}$, with a diffusion model that is trained at $\Vtrain = V_{2 \times}$. 
            The interpolation methods, in particular grid-resampling, do not provide performance improvements compared to variational reconstruction with fixed resolution. 
        }
        \label{fig:resolution_interpolation_higher}
 \end{figure}
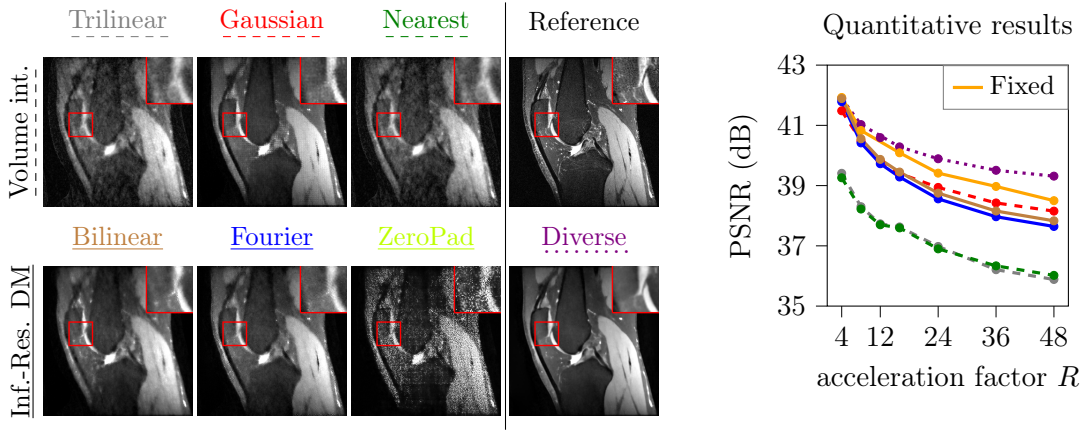

    \paragraph{Approaches 2.1-2.3: Volume interpolation.}  
    Our experimental results, depicted in Figures~\ref{fig:resolution_interpolation_lower} and ~\ref{fig:resolution_interpolation_higher}, show that simple trilinear interpolation closes the performance gap when reconstructing at larger voxel sizes, but that even state-of-the-art interpolation methods fall short when reconstructing at smaller voxel sizes. Regarding the continuous representations, we observe the following.  \\
    \underline{Implicit neural networks:} We estimated the size of the INN architecture required such that we obtain a sufficiently small approximation error on the 3D volumes, and observe a high GPU memory and compute cost (see Figure~\ref{fig:appendix_inn_main}). 
    Our results on reconstructing $4 \times$ downsampled volumes show reasonable performance, but fitting takes multiple hours, rendering INNs infeasible for high-resolution 3D MRI. \\
    \underline{Gaussian interpolation:} We investigated the in-distribution performance using Gaussian splatting, and note that it provides competitive performance, compared to the voxel-based representation (see Figure~\ref{fig:appendix_gaussian_vs_voxelrep}).
    The experiments on reconstructing at smaller voxel-sizes demonstrate that Gaussian splatting outperforms grid-resampling interpolation by a large margin, but is less effective than training on diverse resolutions.

    \paragraph{Approaches 3.1-3.3: Infinite-resolution diffusion models.} 
    We examine the effectiveness of interpolating the depthwise convolutional kernels of the infinite diffusion model architecture discussed previously. 
    The results, depicted in Figures~\ref{fig:resolution_interpolation_lower} and~\ref{fig:resolution_interpolation_higher}, 
    show that zero-padding in Fourier space performs similar or better than bilinear interpolation. 
    Both methods outperform zero-padding in image space, with the results stated in the appendix. 
    Interestingly, we find that infinite resolution diffusion models using kernel interpolation methods are more robust compared to volume interpolation methods, but are still outperformed by diverse-resolution training.

\section{Conclusion}

In this paper we first demonstrated that 3D diffusion-based MRI reconstruction is quite sensitive to shifts in resolution that occur in practice. 
That is, the performance degrades when the voxel size of the measurement is different to the voxel size of the training data.

As a solution, we proposed a variational 3D MRI reconstruction method, which leverages a 2D diffusion prior to regularize randomly sampled slices from all anatomical planes during reconstruction.  
The method outperforms posterior sampling and classical baselines in our experiments, and is further shown to be robust to resolution-shifts by training the diffusion model on diverse resolutions. 

We also propose and study a variety of state-of-the-art interpolation methods based on as 3D complex Gaussian splatting, INNs, and infinite dimensional diffusion models to deal with resolution shifts, and interestingly find that the data-centric approach of training on diverse data outperforms those state-of-the-art interpolation methods. 

Resolution shifts as considered here can be viewed as a type of distribution shift since the distribution from train to test distribution changes. Our results are consistent with deep learning and conventional methods beeing sensitive to distribution shifts~\cite{darestani_MeasuringRobustnessDeep_2021}. Depending on the type of distribution shifts, model-centric~\cite{darestani_TestTimeTrainingCan_2022}  or data-centric approaches~\cite{linRobustnessDeepLearning2024} can be effective for mitigating distribution shifts; in this work we find the data-centric approach to be most efficient for resolution shifts.   

\paragraph{Reproducability} The repository at \url{https://github.com/MLI-lab/resolution_robust_3d_mri} contains the code to reproduce results in the main body of this paper.

\paragraph{Acknowledgments} A.K. and R.H. are supported by the Deutsche Forschungsgemeinschaft (DFG, German Research Foundation) - 456465471,
464123524 and the German Federal Ministry of Education and Research, and the Bavarian
State Ministry for Science and the Arts.

{
  \renewcommand{\emph}[1]{\textit{#1}} 
  \printbibliography                  
}

\cleardoublepage

\appendix
\renewcommand\thefigure{\thesection.\arabic{figure}}    
\setcounter{figure}{0} 

\section{Experimental details and further results}
\label{sec:appendix_details_exp_results}
We follow the outline of Section~\ref{sec:experiments}, and start by providing details on training the diffusion models. Then, we discuss 3D reconstruction without resolution-shifts, turn to experimental results on performance drops and methods to achieve resolution robustness.

\subsection{Training 2D diffusion priors}
For the diffusion models, we choose the DDPM parameterization for the diffusion prior~\citep{hoDenoisingDiffusionProbabilistic2020}, where the forward process is described as $p(\vx_{t+1} \mid \vx_t) = \mathcal{N}(\vx_{t+1}; \sqrt{1-\beta_t} \vx_t, \beta_t \mI)$. The stepwise noise levels $\beta_t$ are chosen according to a linear schedule $\beta_t = \beta_{\text{min}} + \frac{t}{T} (\beta_{\text{max}}-\beta_{\text{min}})$, with $\beta_{\text{min}}=0.0001$, $\beta_{\text{max}}=0.02$ and $T = 1000$ steps.

We train the network using the Adam optimizer with learning rate $0.0001$ and batch size $4$. We employ exponential model averaging with rate $0.999$ after $500$ gradient iterations.

\subsection{3D reconstruction without resolution-shifts}
We start by comparing the proposed variational method to posterior-sampling and classical baselines, followed by  hyperparameter studies on the variational method.

\begin{figure}[h!]
    \centering
    \input{figures_onecolumn_appendix/ssim_in_distr_comparision.tikz}
    \vspace{0.5cm}
    \caption{
        \textbf{Performance of 3D reconstruction methods in SSIM.} We reconstruct complex 3D volumes from undersampled multicoil knee MRI acquisitions of the Stanford 3D dataset (without downsampling), and compare the proposed variational method to classical and posterior-sampling baselines. We calculate the SSIM for each image in the sagittal plane and report the average over the volumes. We observe that the variational method outperforms the baselines, which is consistent to the results of Figure~\ref{fig:recon_methods_std_performance_comparison}.
        \label{fig:appendix_recon_methods_without_shift_ssim}
    }
\end{figure}
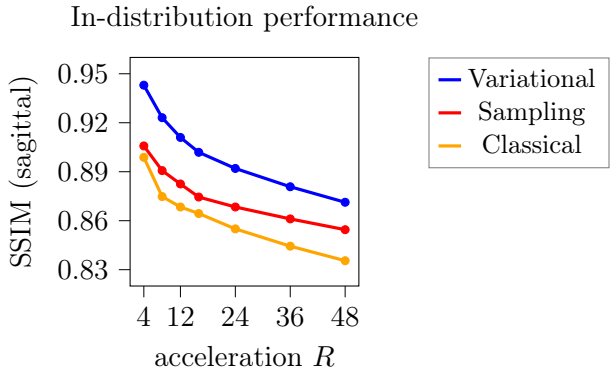

\paragraph{Performance comparison in SSIM.} 
In addition to the results of Figure~\ref{fig:recon_methods_std_performance_comparison}, where we provided visual examples and quantified the performance using PSNR, we calculated the slice-wise SSIM and report the results in  Figure~\ref{fig:appendix_recon_methods_without_shift_ssim}. 

\newpage
\paragraph{Hyperparameters of the variational method.} 
We investigate two main hyperparameters related to our proposed randomized regularization scheme. 
First, we recall that the proposed variational method consists of solving:
\begin{align}
    \begin{split}
        \hat{\vx} (\vy) = \underset{\vx \in \mathbb{C}^N}{\argmin} \sum_{i=1}^C \norm[2]{\vy_i - \mM \mFourier{3} \mS_i \vx}^2
        + \lambda \EX[\mathbf{s} \sim \text{2D-Slices}(\vx)]{\mathcal{R}_{\text{diff}}(\mathbf{s})},
   \end{split}
   \label{eq:appendix_variational_objective}
\end{align}
where the regularization objective is given by:
\begin{equation}
    \mathcal{R}_{\text{diff}}(\mathbf{s}) = \EX[t \sim \mathcal{U}(0, T'), \vep \sim \mathcal{N}(0, \mI)]{w(t) \norm[2]{\vep_{\vphi}(\mathbf{s}_t; t) - \vep}^2}.
    \label{eq:appendix_regularizer}
\end{equation}
Here, $\mathbf{s}_t = \sqrt{1 - \sigma_t^2} \mathbf{s} + \sigma_t \vep$ is a noisy version of the sampled 2D complex slice $\mathbf{s}$. Moreover, we choose the weighting factor $w(t)$ following~\citet{mardaniVariationalPerspectiveSolving2023a}.

We investigate the upper bound $T' < T = 1000$ for sampling the timesteps in the regularization term~\eqref{eq:appendix_regularizer}, and present the results in Figure~\ref{fig:appendix_sweep_diffprior_T}. 
We observe that there is an optimal upper bound $T'$ for the sampled timtestep, which is notably below the number of timesteps used during training the diffusion model.

\begin{figure}
    \centering
    \begin{minipage}[t]{0.48\textwidth} 
        \centering
    \input{figures_onecolumn_appendix/noise_loss_Tprime.tikz}
    \vspace{0.5cm}
    \caption{
        \textbf{Choosing the upper bound for sampling the timesteps is critical for performance.} We perform voxel-based variational reconstruction from $12 \times$ and $48 \times$ accelerated 3D MRI measurements, where we sample timesteps $t \sim \mathcal{U}(0, T')$ for approximating expectation in the regularization objective~\ref{eqn:reg_objective_R_diff}. Here, the diffusion model is trained with $T = 1000$ timesteps, but we observe that choosing a smaller timestep is critical for performance. 
        In all other experiments in this paper, we choose $T' = 400$.
        \label{fig:appendix_sweep_diffprior_T}
        }
    \end{minipage}
    \hfill 
  \begin{minipage}[t]{0.48\textwidth}
    \centering
    \input{figures_onecolumn_appendix/slice_budget.tikz}
    \vspace{0.5cm}
    \caption{
        \textbf{Randomly regularizing a fraction of the slices in each iteration is sufficient.} 
        We perform variational regularization from $48 \times$ accelerated 3D MRI measurements, and very the number of slices extracted along each anatomical plane, that is $3 \cdot S$ slices are extracted per iteration in total. We observe that using sufficiently many slices is critical for performance, but further increasing the number of slices results in marginal performance gains.
        \label{fig:appendix_slice_budget}
    }
  \end{minipage}
\end{figure}

Moreover, we vary the number of randomly selected slices taken from the volume for regularization, to approximate the expectation in~\eqref{eq:appendix_variational_objective}. The results, depicted in Figure~\ref{fig:appendix_slice_budget}, show that there is a certain minimum number of slices required to obtain good performance, and that further increasing the number of slices only provides marginal benefits (e.g. $50$ slices in each epoch and direction are sufficient for a knee volume of dimensions $256 \times 320 \times 320$).

\subsection{Performance drops for variational and posterior sampling methods}
We first study the variational method and posterior-sampling methods individually, and conclude by comparing their sensitivities in terms of SSIM.

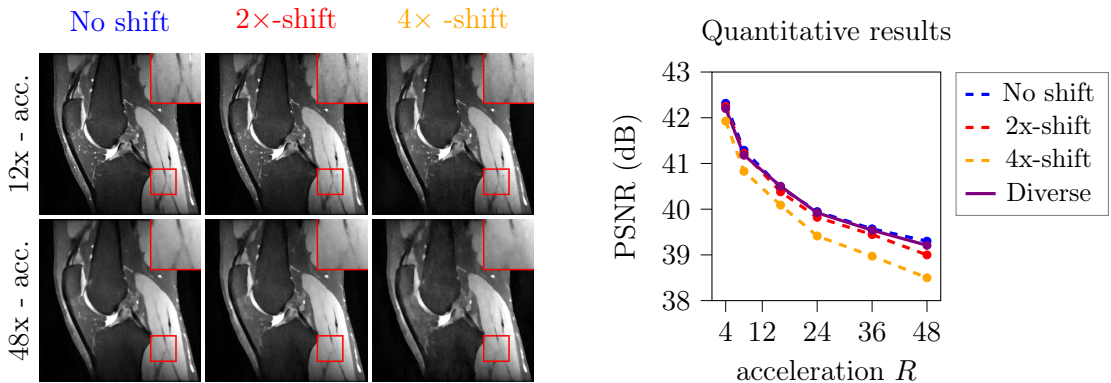
\begin{figure} [h!]
    \begin{centering}
        \hspace{1cm}
        \input{figures_onecolumn_appendix/st3_shift_pngs__sensitivity.tikz}
        \vspace{0.5cm}
    \end{centering}
        \caption{
            \textbf{Variational reconstruction of 3D knee volumes is sensitive to resolution-shifts.}
            We perform reconstruction from undersampled multicoil 3D aquisitions of the Stanford dataset, with diffusion models trained at voxel sizes $0.5$mm (no shift), $1$mm ($2 \times$ -shift) and $2$mm ($4 \times$-shift). 
            Consistent to the findings on brain volumes (see Figure~\ref{fig:main}), we observe the sensitivity to resolution-shifts, and that diverse-resolution training effectively closes the gap.
        }
        \label{fig:mainfig_stanford}
\end{figure}

\paragraph{Sensitivity of the variational approach.}
In Figure~\ref{fig:main} of the main paper, we demonstrated the sensitivity of variational diffusion-based 3D reconstruction on high-field brain volumes.
We performed the same experiments on the high-field knee dataset, with the results depicted Figure~\ref{fig:mainfig_stanford}. We consistently observe that 3D reconstruction is sensitive to resolution-shifts, and that the gap can be effectively closed by diverse training.
  
\begin{figure} [h!]
    \begin{centering}
        \hspace{1cm}
        \input{figures_onecolumn_appendix/st3_shift_pngs__sensitivity_ps.tikz}
        \vspace{0.5cm}
    \end{centering}
        \caption{
            \textbf{Posterior-sampling based 3D reconstruction is sensitive to resolution-shifts.}
            We employ the posterior-sampling baseline for reconstructing complex 3D knee volumes from undersampled multicoil 3D MRI measurements at a fixed voxel-size.
            We use the same diffusion models pretrained on larger voxel sizes as for the variational method (Figure~\ref{fig:mainfig_stanford}), and observe that posterior-sampling based reconstruction is also resolution-sensitive.
        }
        \label{fig:appendix_shift_demo_sampling}
\end{figure}
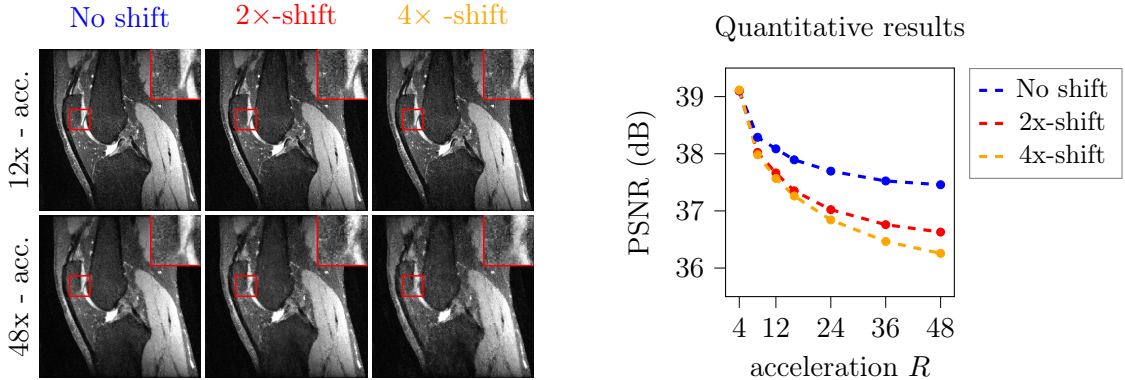

\paragraph{Sensitivity of posterior sampling.}
We use the same diffusion models as for the variational approach, and investigate the sensitivity of posterior sampling-based 3D reconstruction to differences in the measurement and training data's voxel size. 
The results, depicted in Figure~\ref{fig:appendix_shift_demo_sampling}, show that the posterior sampling baseline is also sensitive to resolution-shifts.

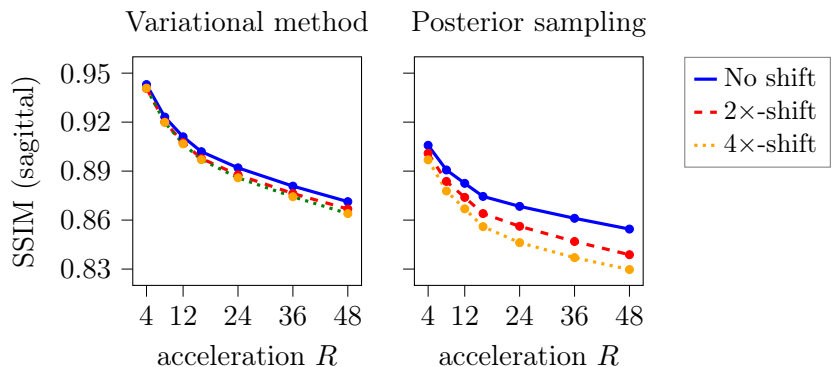
\begin{figure} [h!]
        \centering
        \input{figures_onecolumn_appendix/ssim__comp__sensitivities.tikz}
        \vspace{0.5cm}
        \caption{
            We calculate the SSIM metric for the reconstructions obtained from the experiments performed for the proposed variational approach (Figure~\ref{fig:mainfig_stanford}) and the posterior sampling method (Figure~\ref{fig:appendix_shift_demo_sampling}).
            We find that the performance drop, observed visually and in PSNR, can also be observed in terms of the SSIM, and that the performance drops appears larger for the posterior sampling method in our experiments.
        }
        \label{fig:appendix_ssim_sensitivities}
\end{figure}

\paragraph{Comparing the sensitivities in terms of SSIM.} 
We calculate the structural similarity index measure (SSIM) by taking the magnitude of the complex reconstruction and reference volumes, and then taking the average of the SSIM of each slice in the sagittal plane ($256$ per volume for the knee dataset). 
The results, depicted in Figure~\ref{fig:appendix_ssim_sensitivities}, show that both methods suffer from resolution-shifts in SSIM. 
Relative to the variational method, the performance drop is larger for the posterior-sampling methods in our experiments.

\newpage
\subsection{Achieving resolution robustness}
Towards resolving the sensitivity to resolution-shifts, we investigated an array of data and model-centric approaches, and reported the main results in the paper. 
In the following, we provide results on the in-distribution performance of INNs and Gaussian splatting, state the full results on kernel interpolation and present results on diverse resolution-training on ultra high-field brain volumes. 

\paragraph{Implicit neural representations.} 
We consider INNs as continuous representation for the reconstructed volume, which allows to sample slices at the resolution of the training data. We tune the scale of the randomly sampled weights of the Fourier feature input layer, which determines the resolution of the representation~\citep{tancik_FourierFeaturesLet_2020a}.

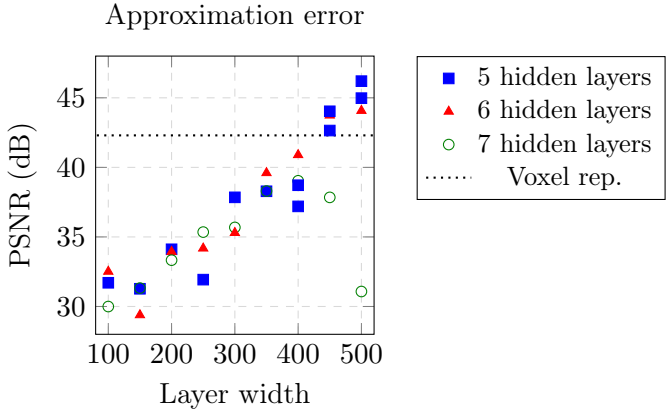
\begin{figure} [h!]
    \begin{centering}
        \hspace{3cm}
        \input{figures_onecolumn_appendix/inn_repr_width.tikz}
        \vspace{0.5cm}
    \end{centering}
        \caption{
            \textbf{Obtaining a sufficiently small approximation error on complex 3D MRI volume requires a wide network.} We consider INNs with an architecture similar to~\citet{tancik_FourierFeaturesLet_2020a} and sweep over the width and number of linear layers. 
            We observe that the network needs to be very wide, entailing high GPU memory and compute costs. As a baseline, we consider the performance of voxel-based variational reconstruction at $4 \times$ acceleration (dotted line).
        }
        \label{fig:appendix_inn_main}
\end{figure}

Furthermore, we sweep over the width and number of the linear layers such that we obtain a sufficiently low approximation error in terms of PSNR. 
We perform this analysis by fitting the INN $f_{\theta} \colon [-1,1]^3 \to \mathbb{C}$ to fully-sampled references volumes of the datasets.
We consider the performance of the proposed variational reconstruction method, at fixed resolution and $4 \times$ acceleration, as a baseline.  
Figure~\ref{fig:appendix_inn_main} shows the results for $4\times$ downsampled volumes of Stanford 3D knee dataset, with matrix size $64 \times 80 \times 80$.
We observe that the linear layers need to be wide such that we obtain a sufficiently small approximation error. This, however, entails a high memory and compute cost. 

We employ INNs with layer width of $450$ for variational reconstruction on the $4 \times$ downsampled datasets, with the results shown in Figure~\ref{fig:resolution_interpolation_lower} of the main paper. Across accelerations, reconstruction with the INN requires over $40$G of GPU memory, and exhibits long reconstruction times exceeding $10$ hours on a A40 Nvidia GPU, rendering INNs infeasible for practical MRI reconstruction. 
Moreover, the number of required INN function evaluation scales cubically with the downsampling factor, in that we need $8 \times$ the evaluations for the $2 \times$ downsampled volumes compared to $4 \times$ downsampled volumes. Due to this scaling INNs are infeasible for high-resolution 3D MRI in practice.

Our findings are consistent with those of \citet{arefeenINFusionDiffusionRegularized2024}, who report that encoding all three spatial dimensions using the INNs entails prohibitively high GPU memory and compute costs for 3D MRI. 
Since the goal of~\citet{arefeenINFusionDiffusionRegularized2024} is to make use of the architectural bias of INNs for in-distribution performance, they propose a hybrid 3D approach, encoding only 2 out of 3 dimensions continuously and obtaining the third dimension as output of the INN. Specifically, they use $g_{\theta} \colon [-1,1]^2 \to \mathbb{C}^{80}$ as representation, for fitting MRI volumes with matrix-size $256 \times 256 \times 80$. 
Therefore, the approach still pertains to a fixed voxel size along the third dimension.

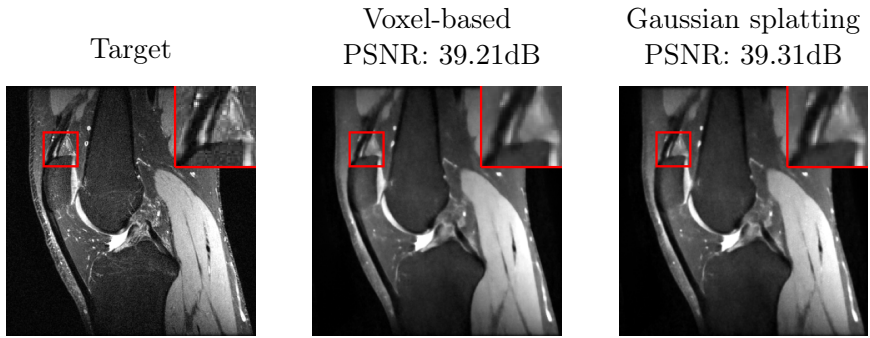
\begin{figure} [h!]
    \begin{centering}
        \hspace{2cm}
        \input{figures_onecolumn_appendix/gaussian_vs_voxelrep.tikz}
        \vspace{0.5cm}
    \end{centering}
    \caption{\textbf{Variational reconstruction with Gaussian splatting exhibits similar results as with voxel-representation.}
        We perform variational 3D MRI reconstruction from $48 \times$ undersampled 3D MRI measurements with Gaussian splatting and voxel-based representation, respectively. For the 3D Gaussian splatting, we adopted the methodology described in the main text, and chose a representation with $12$M Gaussians, initialized using the coarse MVUE estimation. Both methods obtain similar results. From visual inspection, however, we observe that the reconstructions with Gaussian splatting are less smooth compared to those obtained with voxel-representation.
    }
    \label{fig:appendix_gaussian_vs_voxelrep}
\end{figure}

\paragraph{3D Gaussian splatting.} 
We further adapted Gaussian splatting~\citep{kerbl3DGaussianSplatting2023b} to complex 3D imaging for continuously representing the volume. 
In the main paper, we presented results on employing it for volume interpolation, for mitigating performance drops due to resolution-shifts. In the following, we consider its in-distribution performance, i.e. perform reconstruction without any resolution-shift. The results, depicted in Figure~\ref{fig:appendix_gaussian_vs_voxelrep}, indicate that complex 3D Gaussian splatting performs well in-distribution, with a slight improvement over voxel-representation.

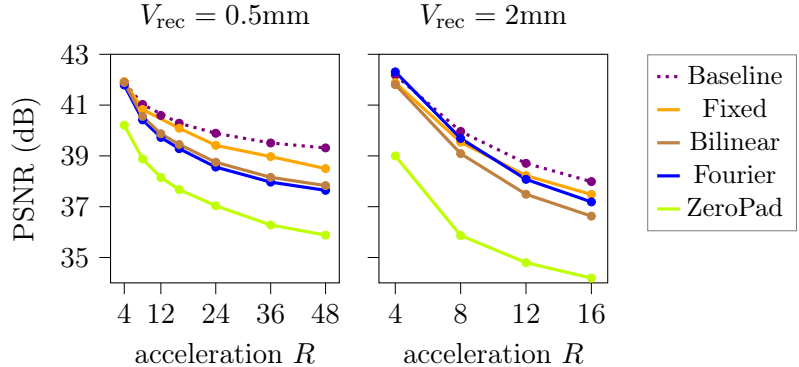
\begin{figure} [h!]
    \begin{centering}
        \hspace{2cm}
        \input{figures_onecolumn_appendix/kernel_interpolations_zero_pad.tikz}
        \vspace{0.5cm}
    \end{centering}
        \caption{
            We perform variational reconstruction on undersampled 3D MRI knee acquisitions at voxel-sizes $\Vrecon = 0.5$mm and $\Vrecon = 2$mm.
            We consider infinite resolution diffusion models pre-trained at resolution $\Vtrain = 1$mm, and employ different methods for interpolating the kernels. We compare to the baseline of reconstructing with a diffusion model trained at $\Vtrain = \Vrecon$, and compare to reconstructing with a standard diffusion model at fixed resolution. We show some visual examples in Figures~\ref{fig:resolution_interpolation_higher} and~\ref{fig:resolution_interpolation_lower} of the main paper.
            \label{fig:appendix_kernel_interpolation_zero_pad}
        }
\end{figure}

\paragraph{Zero-padded kernel interpolation.}
In the main paper, we presented results on performing variational 3D reconstruction using a multiscale U-Net architecture proposed by~\citet{bond-taylorInftyDiffInfinite2023}. 
We investigated several methods for interpolating the depthwise kernels of the first and last residual blocks: Bilinear interpolation, zero-padding in Fourier space and zero-padding in image space. 
To increase visibility, we omitted the results for image-space zero-padding the depthwise kernels in the main paper. 
We included it in Figure~\ref{fig:appendix_kernel_interpolation_zero_pad}. We observe that zero-padding in image space underperforms other interpolation methods by a large margin, which is consistent to findings on other domains such as classification~\citep{kabriResolutionInvariantImageClassification2023}. 

\paragraph{Diverse-resolution training on ultra-high field volumes.}
In our work, we investigate the effectivity of diverse-resolution training on high-field brain and knee volumes, and on ultra-high field brain volumes. 
We presented the results for the high-field brain and knee volumes in the main paper (Figures~\ref{fig:main} and~\ref{fig:resolution_shifts_cross_stanford_3d}). In the following, we state the results on the ultra-high field volumes, where we employed a Gaussian 2D mask for retrospective undersampling. The results, depicted in Figure~\ref{fig:appendix_ahead}, confirm that variational reconstruction is sensitive to resolution-shift and that diverse-resolution training effectively closes the performance gap.
\begin{figure} [h!]
    \begin{centering}
        \hspace{1cm}
        \input{figures_onecolumn_appendix/ahead_shift_pngs__sensitivity.tikz}
        \vspace{0.5cm}
    \end{centering}
        \caption{
            We perform variational 3D MRI reconstruction from undersampled ultra-high-field MRI brain T1-weighted acquisitions. 
            We perform reconstruction at voxel-size $\Vrecon = 1.4$mm, with diffusion-models trained at $\Vtrain = 0.7$mm, $1.4$mm and $2.8$mm. 
            We observe that a shift in voxel-size yields a loss in reconstruction performance, with different artifacts depending on whether the voxel-size during training is lower or larger than during reconstruction. 
            Moreover, diverse-resolution training mitigates the performance drop effectively. 
            Note, that we increased the brightness of the zoomed-in region by a constant factor for visibility.
        }
        \label{fig:appendix_ahead}
\end{figure}

\subsection{Dataset urls and licenses}
\label{subsec:appendix_required_compute}

The experiments are performed using the Stanford 3D dataset~\citep{eppersonCreationFullySampled2013}, the Calgary-Campinas 3D brain dataset~\citep{souzaCC3592018}, and the AHEAD 7T brain dataset~\citep{alkemadeAmsterdamUltrahighField2020}, as described in the main text. The datasets can be downloaded at \href{http://mridata.org/}{http://mridata.org/} (licensed under CC-BY-NC 4.0), \href{https://portal.conp.ca/dataset?id=projects/calgary-campinas}{https://portal.conp.ca/dataset?id=projects/calgary-campinas} (licensed under CC-BY-ND 4.0) and \href{https://dataverse.nl/dataset.xhtml?persistentId=doi:10.34894/IHZGQM}{https://dataverse.nl/dataset.xhtml?persistentId=doi:10.34894/IHZGQM} (licensed under CC-BY 4.0).

\end{document}

%% file: figures_onecolumn/fig1_res_demo_48x_with_plot.tikz
\begin{tikzpicture}

    \definecolor{green}{RGB}{0,128,0}
    \definecolor{orange}{RGB}{255,165,0}
    \definecolor{purple}{RGB}{128,0,128}

    \newcommand{\nodeDistFigOne}{0.0cm}
    \newcommand{\pngWidthFigOne}{0.173\textwidth}

    \node (img_48x_same) {\reflectbox{\rotatebox[origin=c]{180}{\scalebox{0.92}{\includegraphics[width=\pngWidthFigOne]{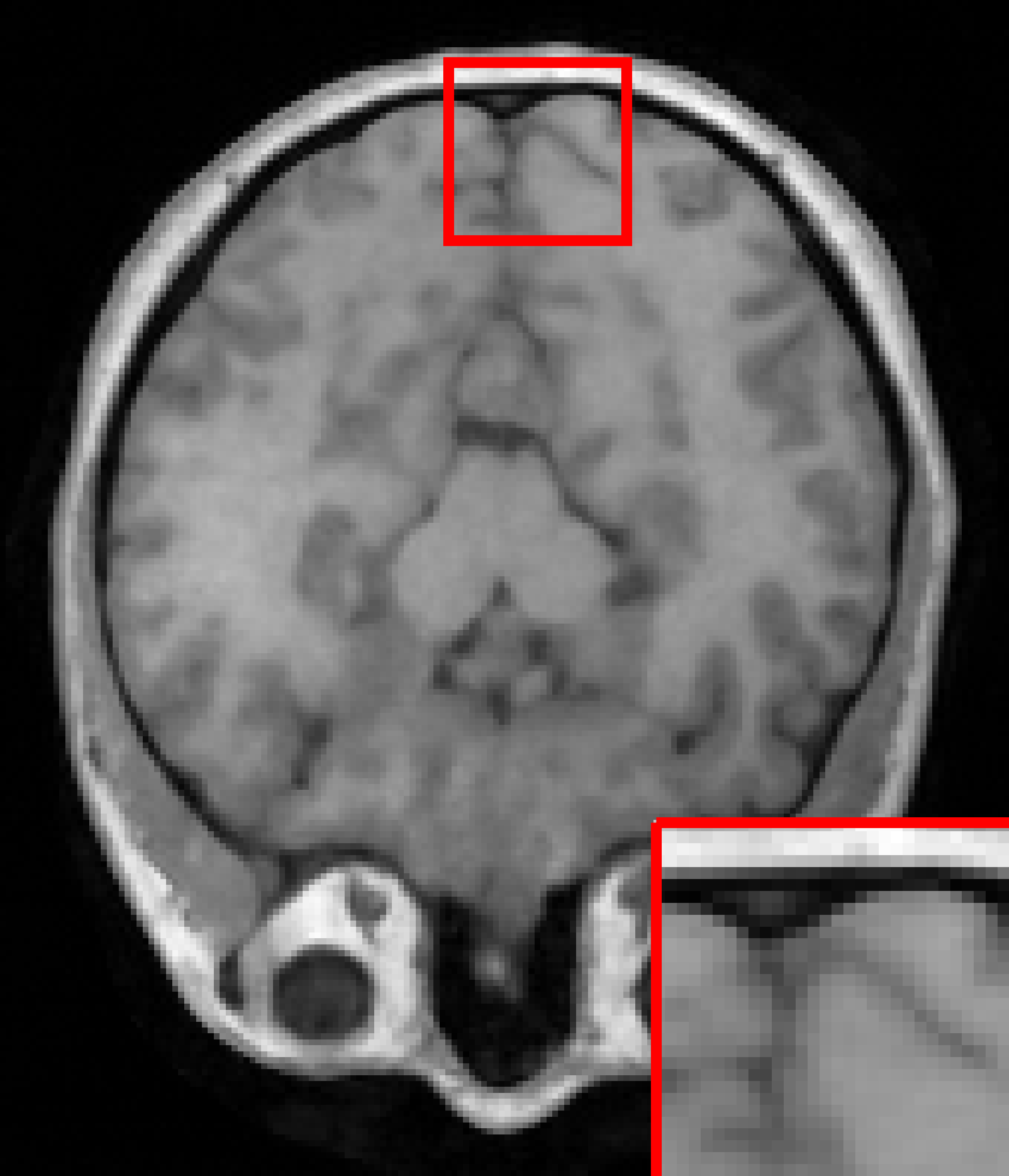}}}}};
    \node [right=\nodeDistFigOne, at=(img_48x_same.east)](img_48x_2x) {\reflectbox{\rotatebox[origin=c]{180}{\scalebox{0.92}{\includegraphics[width=\pngWidthFigOne]{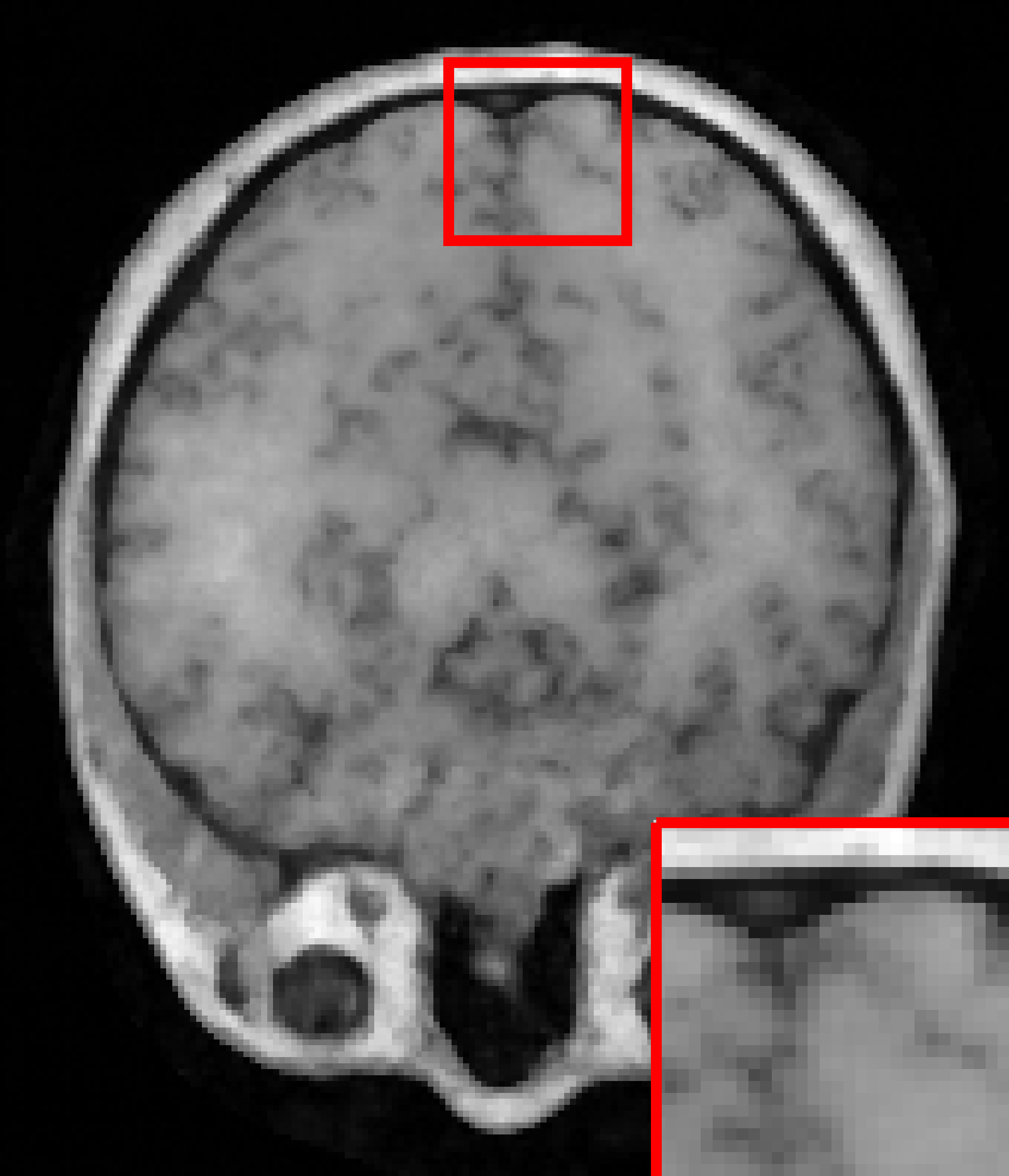}}}}};
    \node [right=\nodeDistFigOne, at=(img_48x_2x.east)](img_48x_4x) {\reflectbox{\rotatebox[origin=c]{180}{\scalebox{0.92}{\includegraphics[width=\pngWidthFigOne]{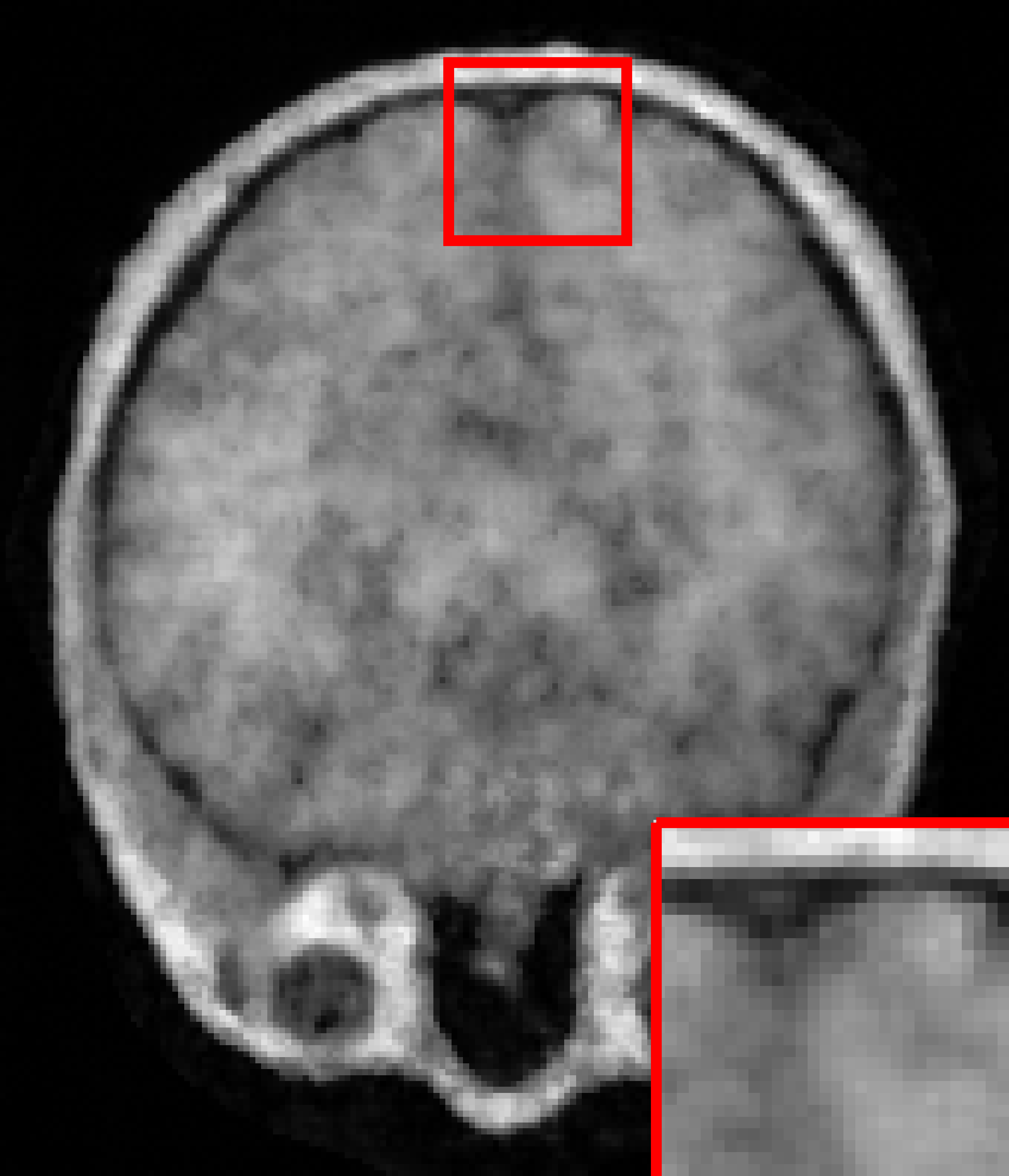}}}}};
    
    \node[above=1pt, at=(img_48x_same.north),font=\color{blue}, align=center] {No shift};
    \node[above=1pt, at=(img_48x_2x.north),font=\color{red}, align=center] {$2 \times$-shift};
    \node[above=1pt, at=(img_48x_4x.north),font=\color{orange}, align=center] {$4 \times$ -shift};

    \node[left=1pt, at=(img_48x_same.west), node distance=0.1cm, rotate=90, anchor=center,yshift=0.1cm,xshift=0.0cm,font=\color{black}] {48x - acc.};
    
    \node[at=(img_48x_4x.east), left=-2.2cm, yshift=-1.5cm]{
        
        \begin{groupplot}[group style={group size=1 by 1},
            width=0.28\columnwidth,
            height=0.28\columnwidth,
            tick align=outside,
            tick pos=left,
            x grid style={darkgray176},
            xtick style={color=black},
            y grid style={darkgray176},
            scaled ticks=false,
            ylabel={PSNR (dB)},
            ytick style={color=black},
            legend image code/.code={
                \draw[#1] (0cm,0.0cm) -- (0.3cm,0.0cm); 
            }, 
            legend style={
              fill opacity=0.8,
              font=\small,
              draw opacity=1,
              text opacity=1,
              at = {(1.4, 1.0)},
              anchor=north,
              draw=gray}
        ]

        \nextgroupplot[xlabel={acceleration factor}, xmin=1, xmax=51, ymin=30, ymax=40, align=center, title={Performance drops}, title style={yshift=-0.08cm},
            xtick={4, 12, 24, 36, 48},
            xticklabels={4, 12, 24, 36, 48},
            ytick={30, 32, 34, 36, 38, 40},
            yticklabels={30, 32, 34, 36, 38, 40},
        ]

            \addplot [draw=purple, fill=purple, forget plot, mark=*, only marks, mark size=1.5]
            table {%
                4 38.705
                8 36.986
                12 35.956
                16 35.227
                24 34.264
                37 33.096
                48 32.384
            };
            \addplot [very thick, purple]
            table {%
                4 38.705
                8 36.986
                12 35.956
                16 35.227
                24 34.264
                37 33.096
                48 32.384
            };
            \addlegendentry{Diverse}
        
            \addplot [draw=blue, fill=blue, forget plot, mark=*, only marks, mark size=1.5]
            table {%
                4 38.787
                8 37.044
                12 36.008
                16 35.335
                24 34.35
                37 32.97
                48 32.307
            };
            \addplot [very thick, blue, dashed]
            table {%
                4 38.787
                8 37.044
                12 36.008
                16 35.335
                24 34.35
                37 32.97
                48 32.307
            };
            \addlegendentry{No shift}
            
            \addplot [draw=red, fill=red, forget plot, mark=*, only marks, mark size=1.5]
            table {%
                4 38.447
                8 36.558
                12 35.381
                16 34.582
                24 33.422
                37 32.08
                48 31.402
            };
            \addplot [very thick, red, dashed]
            table {%
                4 38.447
                8 36.558
                12 35.381
                16 34.582
                24 33.422
                37 32.08
                48 31.402
            };
            \addlegendentry{$2$x-shift}
            
            \addplot [draw=orange, fill=orange, forget plot, mark=*, only marks, mark size=1.5]
            table {%
                4  38.296
                8 36.168
                12 34.805
                16 33.884
                24 32.582
                37 31.01
                48 30.417
            };
            \addplot [very thick, orange, dashed]
            table {%
                4 38.296
                8 36.168
                12 34.805
                16 33.884
                24 32.582
                37 31.01
                48 30.417
            };
            \addlegendentry{$4$x-shift}
    
        \end{groupplot}
    };

\end{tikzpicture}

%% file: figures_onecolumn/3d_reconstruction_comparison_2x3.tikz
\begin{tikzpicture}

    \definecolor{green}{RGB}{0,128,0}
    \definecolor{orange}{RGB}{255,165,0}
    \definecolor{purple}{RGB}{128,0,128}

    \newcommand{\pngwidth}{0.13\textwidth}
    \newcommand{\nodedist}{-0.2cm}

    \node (img_12x_same) {\includegraphics[width=\pngwidth]{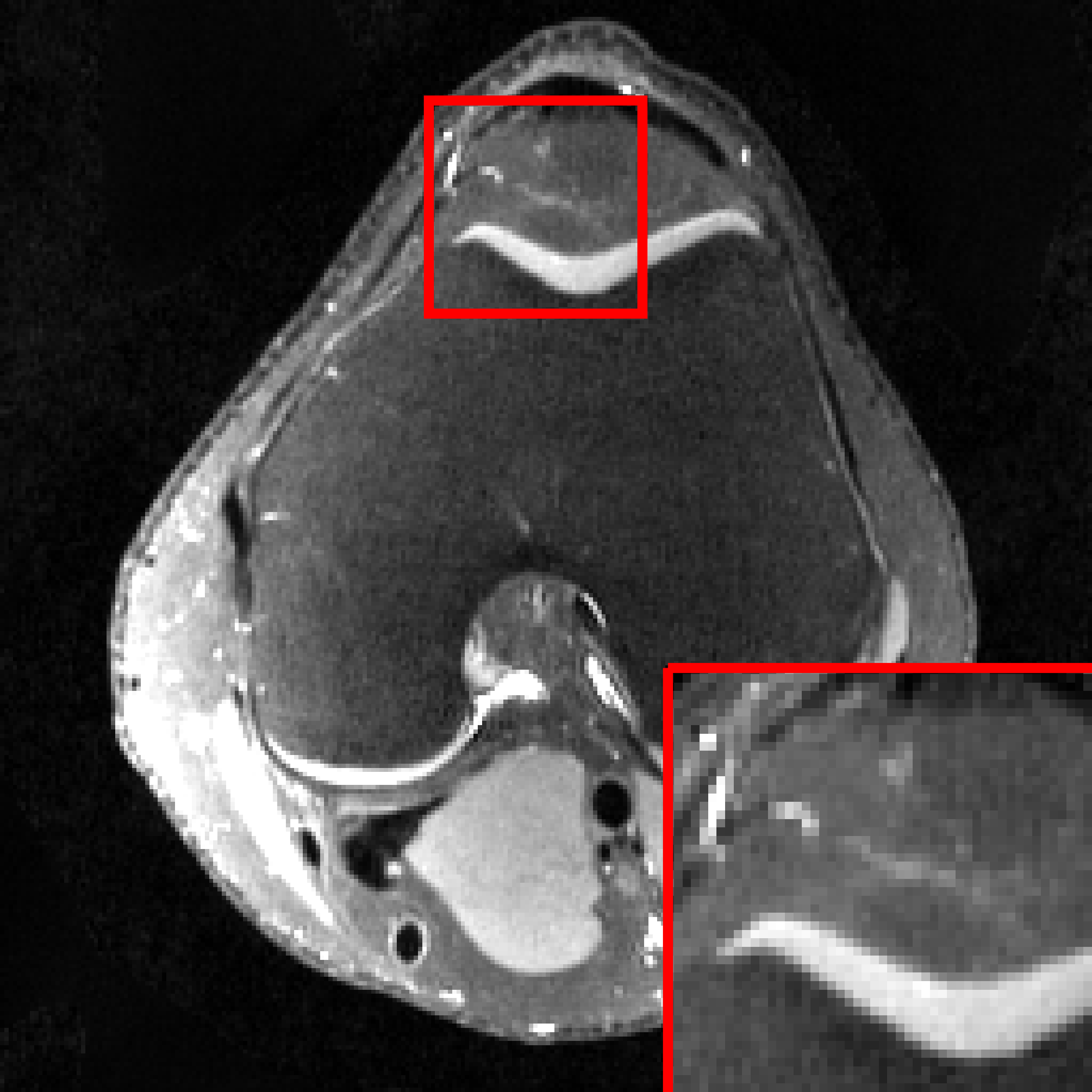}};
    \node [right=\nodedist, at=(img_12x_same.east)](img_12x_sam) {\includegraphics[width=\pngwidth]{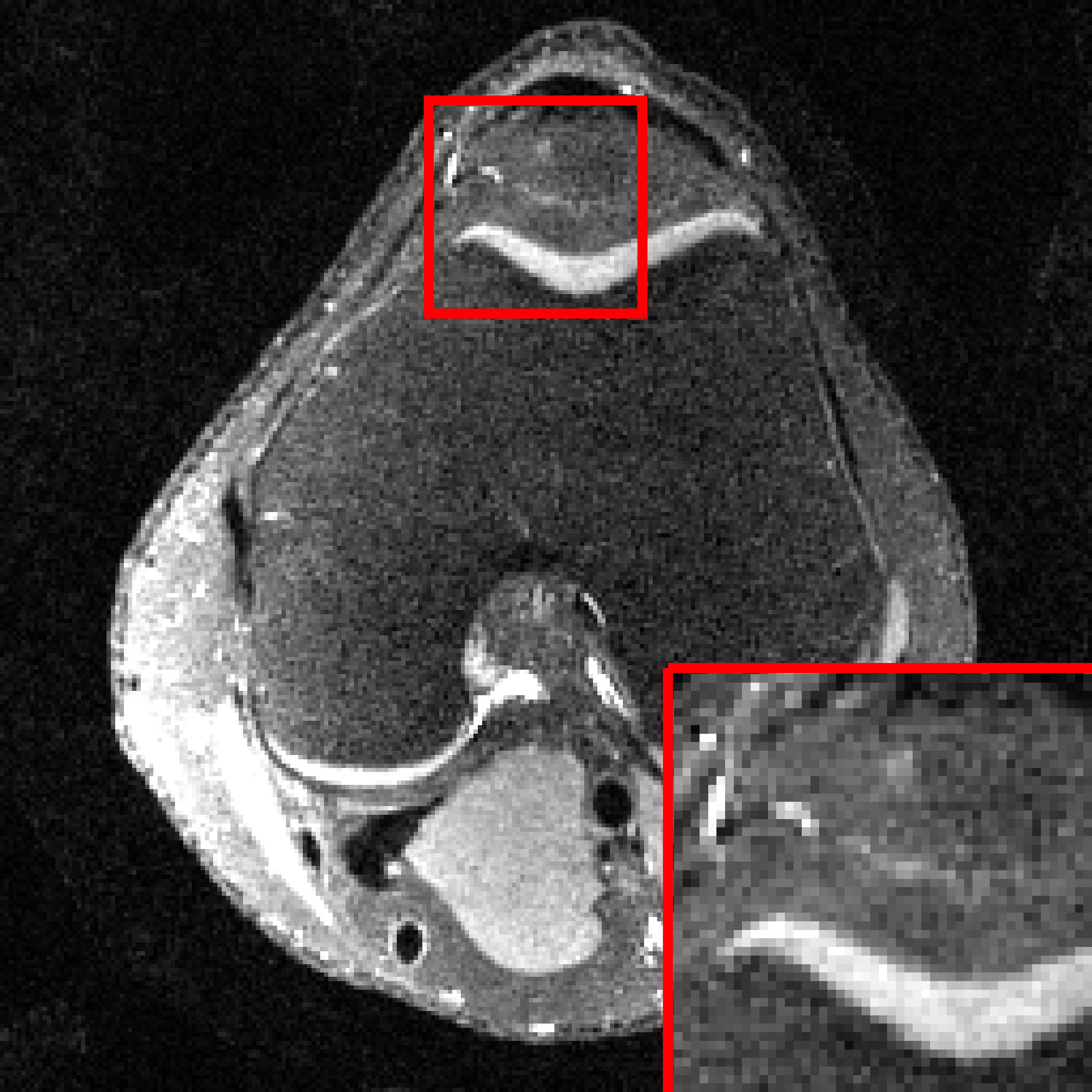}};
    \node [right=\nodedist, at=(img_12x_sam.east)](img_12x_clas) {\includegraphics[width=\pngwidth]{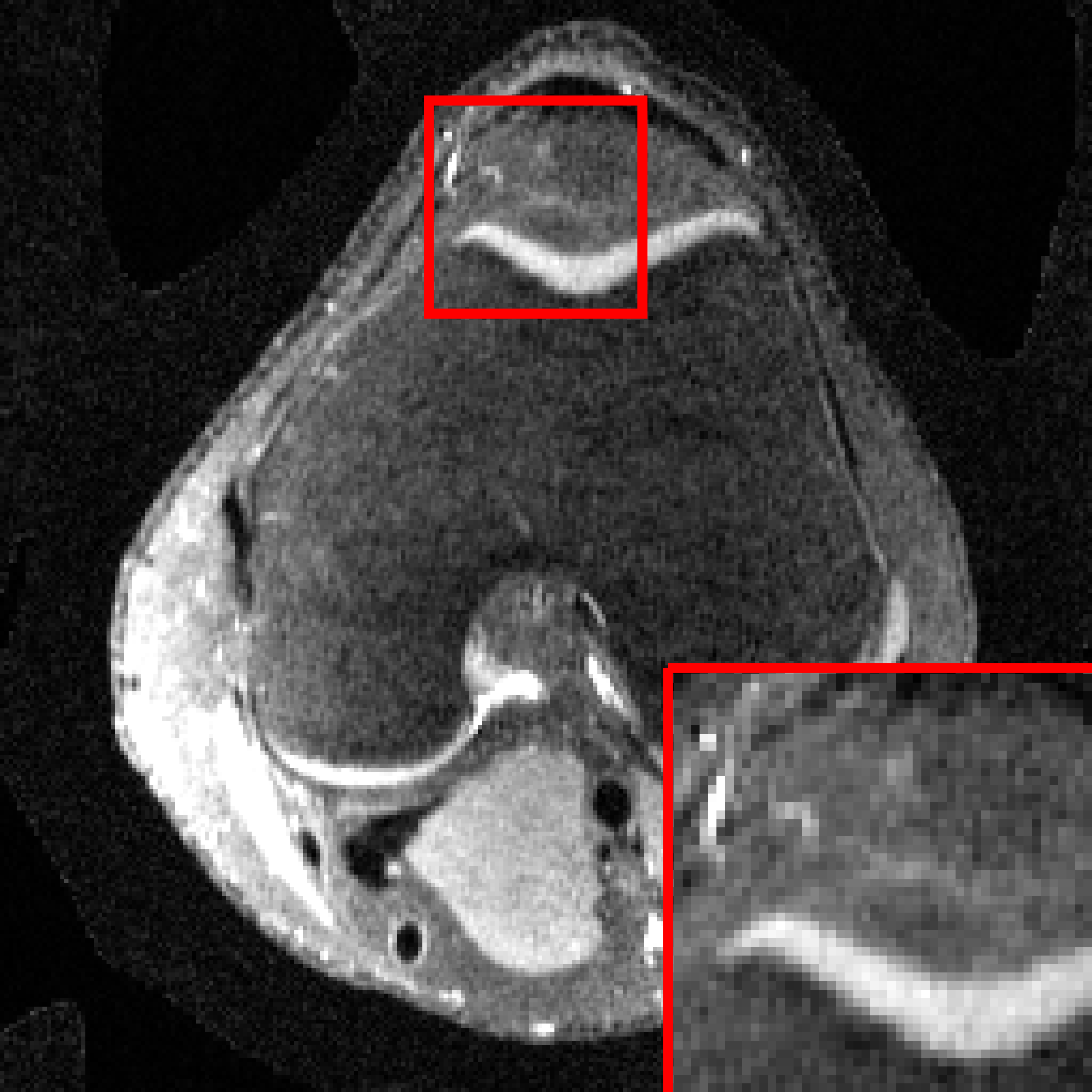}};

    \node [right=\nodedist, yshift=-0.26cm, at=(img_12x_clas.east)](img_target) {\includegraphics[width=\pngwidth]{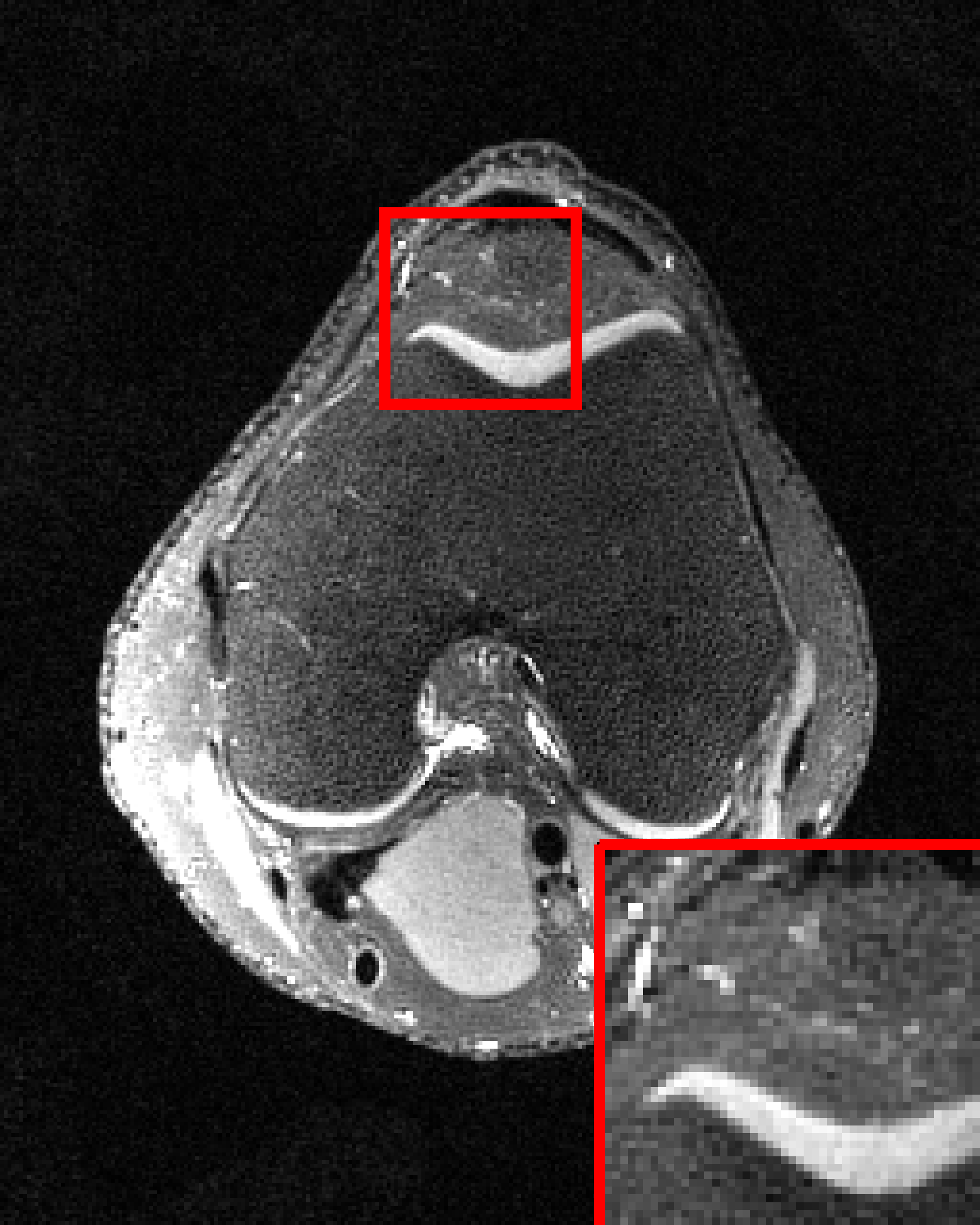}};
    
    \node [below=\nodedist, at=(img_12x_same.south)](img_48x_same) {\includegraphics[width=\pngwidth]{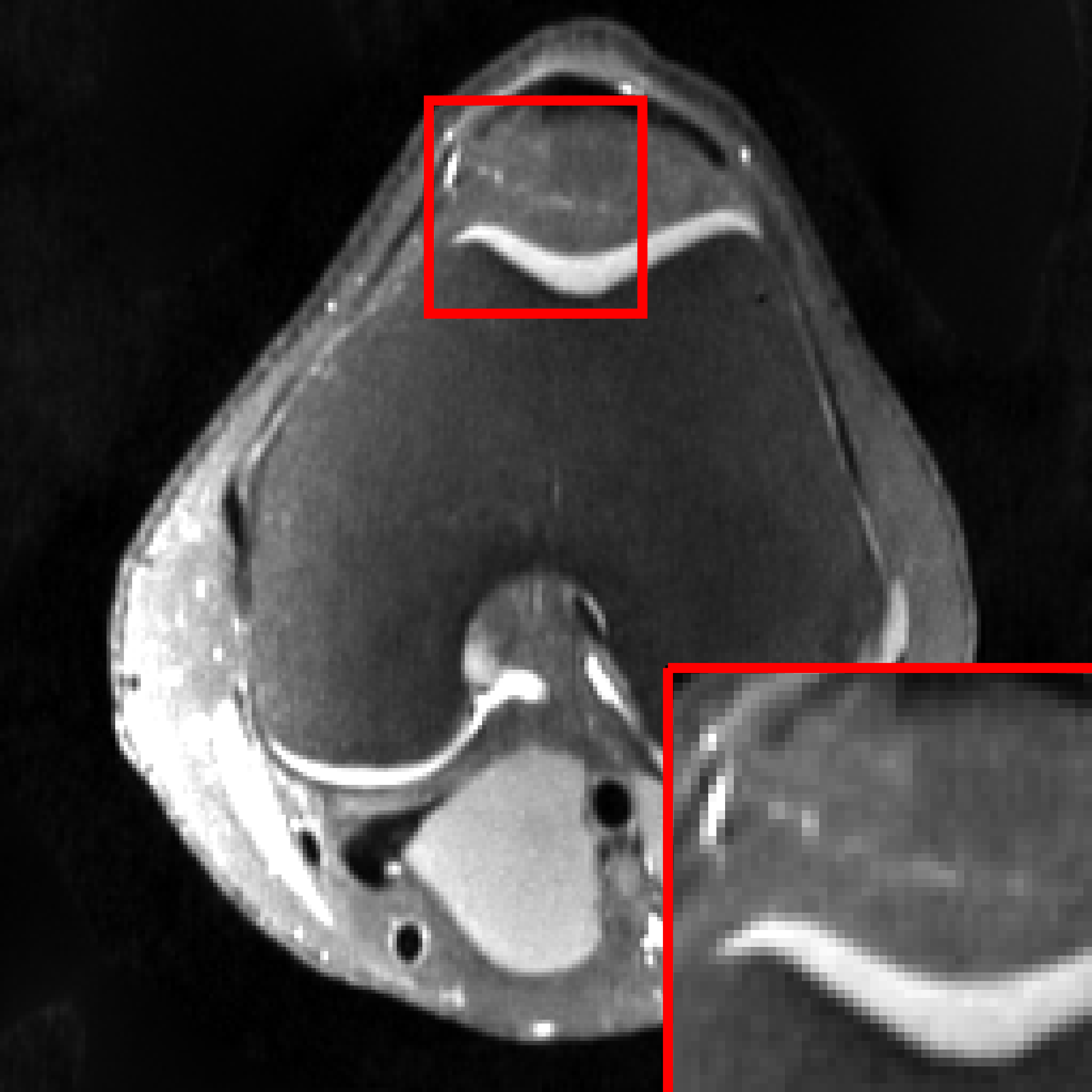}};
    \node [below=\nodedist, at=(img_12x_sam.south)](img_48x_sam) {\includegraphics[width=\pngwidth]{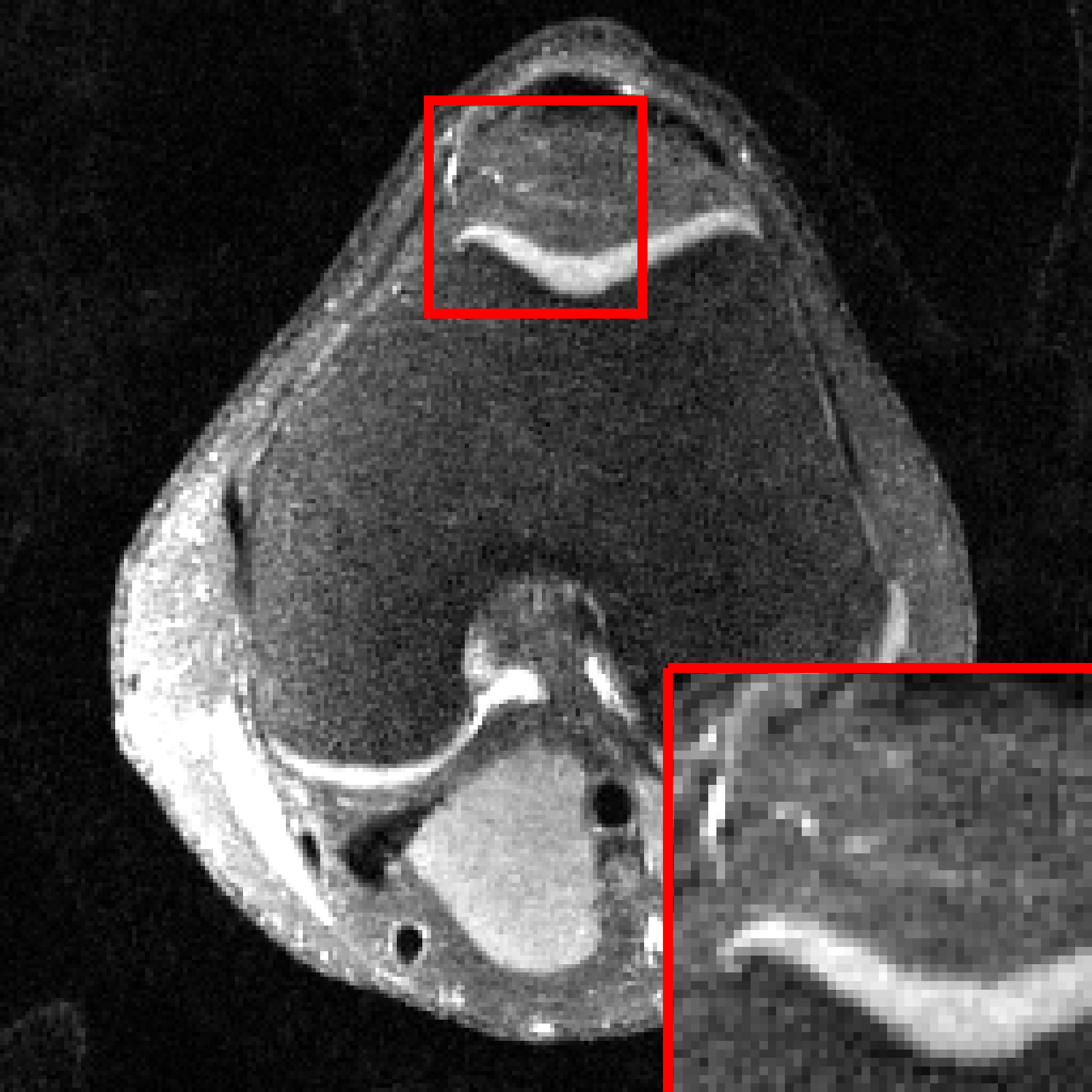}};
    \node [below=\nodedist, at=(img_12x_clas.south)](img_48x_clas) {\includegraphics[width=\pngwidth]{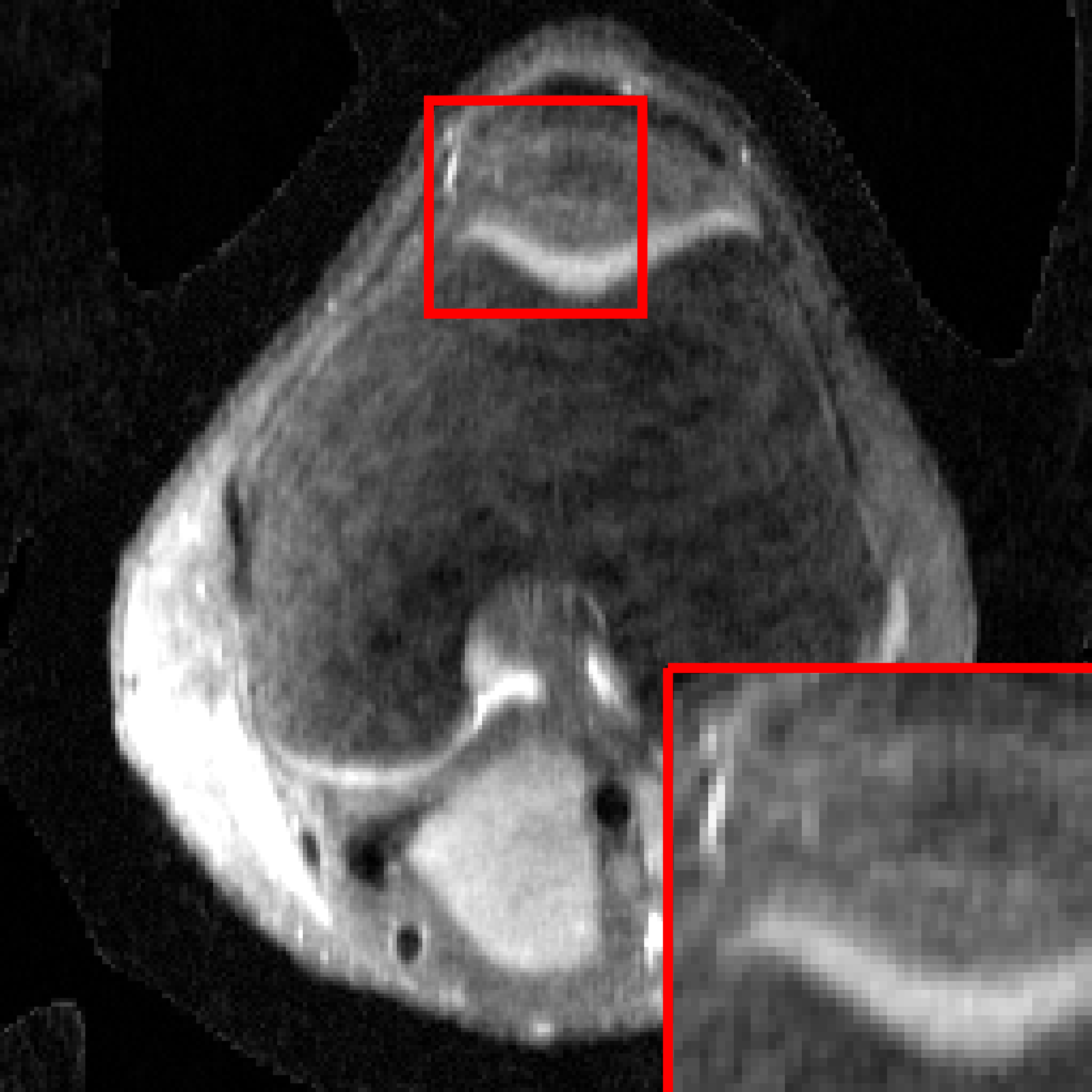}};

    \node[above=1pt, at=(img_12x_same.north),font=\color{blue}, align=center] {Variational};
    \node[above=-1pt, at=(img_12x_sam.north),font=\color{red}, align=center] {Sampling};
    \node[above=1pt, at=(img_12x_clas.north),font=\color{orange}, align=center] {Classical};

    \node[above=1pt, at=(img_target.north),font=\color{black}, align=center] {Reference};

    \node[left=1pt, at=(img_12x_same.west), node distance=0.1cm, rotate=90, anchor=center,yshift=0.1cm,xshift=0.0cm,font=\color{black}] {12x - acc.};
    \node[left=1pt, at=(img_48x_same.west), node distance=0.1cm, rotate=90, anchor=center,yshift=0.1cm,xshift=0.0cm,font=\color{black}] {48x - acc.};

    \node[at=(img_48x_clas.east), right=0.0cm, yshift=0.15cm]{
        
        \begin{groupplot}[group style={group size=1 by 1},
            width=0.28\columnwidth,
            height=0.28\columnwidth,
            tick align=outside,
            tick pos=left,
            x grid style={darkgray176},
            xtick style={color=black},
            y grid style={darkgray176},
            scaled ticks=false,
            ytick style={color=black},
            legend image code/.code={
                \draw[#1] (0cm,0.0cm) -- (0.3cm,0.0cm); 
            }, 
            legend style={
              fill opacity=0.8,
              font=\small,
              draw opacity=1,
              text opacity=1,
              at={(1.4, 1.0)},
              anchor=north,
              draw=gray}
        ]

        \nextgroupplot[ylabel={PSNR (dB)}, xlabel={acceleration factor $R$}, xmin=1, xmax=51, ymin=35.5, ymax=43, title={Quantitative results},
            xtick={4, 12, 24, 36, 48},
            xticklabels={4, 12, 24, 36, 48},
            ytick={36, 38, 40, 42},
            yticklabels={36, 38, 40, 42}
        ]
        
            \addplot [draw=blue, fill=blue, forget plot, mark=*, only marks, mark size=1.5]
            table {%
                4 42.316
                8 41.289
                12 40.875
                16 40.477
                24 39.946
                36 39.57
                48 39.3
            };
            \addplot [very thick, blue, dashed]
            table {%
                4 42.316
                8 41.289
                12 40.875
                16 40.477
                24 39.946
                36 39.57
                48 39.3
            };
            
            \addplot [draw=red, fill=red, forget plot, mark=*, only marks, mark size=1.5]
            table {%
                4 40.715
                8 39.677
                12 39.213
                16 38.872
                24 38.46
                36 38.097
                48 37.899
            };
            \addplot [very thick, red, dashed]
            table {%
                4 40.715
                8 39.677
                12 39.213
                16 38.872
                24 38.46
                36 38.097
                48 37.899
            };
            
            \addplot [draw=orange, fill=orange, forget plot, mark=*, only marks, mark size=1.5]
            table {%
                4 39.001
                8 38.135
                12 37.861
                16 37.76
                24 37.33
                36 36.893
                48 36.58
            }; 
            \addplot [very thick, orange, dashed]
            table {%
                4 39.001
                8 38.135
                12 37.861
                16 37.76
                24 37.33
                36 36.893
                48 36.58
            };
    
        \end{groupplot}
    };

\end{tikzpicture}

%% file: figures_onecolumn/resolution_shift_demo_ext.tikz
\begin{tikzpicture}

    \definecolor{green}{RGB}{0,128,0}
    \definecolor{orange}{RGB}{255,165,0}
    \definecolor{purple}{RGB}{128,0,128}

    \begin{groupplot}[group style={group size=3 by 1,
                horizontal sep=22pt
            },
            width=0.28\columnwidth,
            height=0.28\columnwidth,
            tick align=outside,
            tick pos=left,
            x grid style={darkgray176},
            xtick style={color=black},
            y grid style={darkgray176},
            scaled ticks=false,
            ytick style={color=black},
            legend image code/.code={
                \draw[#1] (0cm,0.0cm) -- (0.4cm,0.0cm); 
            },
            ytick={36, 38, 40, 42},
            yticklabels={36, 38, 40, 42},
            ymin=36,
            ymax=43,
            legend image code/.code={
                \draw[#1] (0cm,0.0cm) -- (0.3cm,0.0cm); 
            }, 
            legend style={
              fill opacity=0.8,
              font=\small,
              draw opacity=1,
              text opacity=1,
              at={(1.6, 1.0)},
              anchor=north,
              draw=gray,
              }
        ]
        
        \nextgroupplot[ylabel={PSNR (dB)}, xlabel={acceleration factor $R$}, xmin=1, xmax=51,
            title={\textbf{High resolution}\\  $V_{\text{rec}} = V_{1\times}$},
            align=center,
            xtick={4, 12, 24, 36, 48},
            xticklabels={4, 12, 24, 36, 48},
        ]
        
            \addplot [draw=blue, fill=blue, forget plot, mark=*, only marks, mark size=1.5]
            table {%
                4 42.316
                8 41.289
                16 40.477
                24 39.946
                36 39.57
                48 39.3
            };
            \addplot [very thick, blue, dashed]
            table {%
                4 42.316
                8 41.289
                16 40.477
                24 39.946
                36 39.57
                48 39.3
            };
            
            \addplot [draw=red, fill=red, forget plot, mark=*, only marks, mark size=1.5]
            table {%
                4 42.25
                8 41.222
                16 40.38
                24 39.8222
                36 39.444
                48 39.0
            };
            \addplot [very thick, red, dashed]
            table {%
                4 42.25
                8 41.222
                16 40.38
                24 39.8222
                36 39.444
                48 39.0
            };
            
            \addplot [draw=orange, fill=orange, forget plot, mark=*, only marks, mark size=1.5]
            table {%
                4 41.926
                8 40.829
                16 40.089
                24 39.414
                36 38.972
                48 38.5
            };
            \addplot [very thick, orange, dashed]
            table {%
                4 41.926
                8 40.829
                16 40.089
                24 39.414
                36 38.972
                48 38.5
            };
            
            \addplot [draw=purple, fill=purple, forget plot, mark=*, only marks, mark size=1.5]
            table {%
                4 42.202
                8 41.181
                16 40.502
                24 39.921
                36 39.54
                48 39.21
            };
            \addplot [very thick, purple]
            table {%
                4 42.202
                8 41.181
                16 40.502
                24 39.921
                36 39.54
                48 39.21
            };

        \nextgroupplot[xlabel={acceleration factor $R$}, xmin=1, xmax=39,
            title={\textbf{Medium resolution} \\ $V_{\text{rec}} = V_{2 \times}$},
            align=center,
            xtick={4, 8, 16, 24, 36},
            xticklabels={4, 8, 16, 24, 36}, 
            yticklabels={}
        ]
        
            \addplot [draw=blue, fill=blue, forget plot, mark=*, only marks, mark size=1.5]
            table {%
                4.0000 41.4994
                8.0000 39.4798
                12.0000 38.5269
                16.0000 38.1047
                24.0000 37.2964
                36.0000 36.8192
            };
            \addplot [very thick, blue, dashed]
            table {%
                4.0000 41.4994
                8.0000 39.4798
                12.0000 38.5269
                16.0000 38.1047
                24.0000 37.2964
                36.0000 36.8192
            };
            
            \addplot [draw=red, fill=red, forget plot, mark=*, only marks, mark size=1.5]
            table {%
                4.0000 42.2735
                8.0000 40.3714
                12.0000 39.4410
                16.0000 39.0353
                24.0000 38.0703
                36.0000 37.6421
            };
            \addplot [very thick, red, dashed]
            table {%
                4.0000 42.2735
                8.0000 40.3714
                12.0000 39.4410
                16.0000 39.0353
                24.0000 38.0703
                36.0000 37.6421
            };
            
            \addplot [draw=orange, fill=orange, forget plot, mark=*, only marks, mark size=1.5]
            table {%
                4.0000 42.2319
                8.0000 40.2996
                12.0000 39.3365
                16.0000 38.8918
                24.0000 37.8509
                36.0000 37.3726
            };
            \addplot [very thick, orange, dashed]
            table {%
                4.0000 42.2319
                8.0000 40.2996
                12.0000 39.3365
                16.0000 38.8918
                24.0000 37.8509
                36.0000 37.3726
            };
            
            \addplot [draw=purple, fill=purple, forget plot, mark=*, only marks, mark size=1.5]
            table {%
                4.0000 42.1611
                8.0000 40.3181
                12.0000 39.4152
                16.0000 39.0028
                24.0000 38.1619
                36.0000 37.6910
            };
            \addplot [very thick, purple]
            table {%
                4.0000 42.1611
                8.0000 40.3181
                12.0000 39.4152
                16.0000 39.0028
                24.0000 38.1619
                36.0000 37.6910
            };

        \nextgroupplot[xlabel={acceleration factor $R$}, xmin=3, xmax=17,
            title={\textbf{Low resolution} \\ $V_{\text{rec}} = V_{4 \times}$},
            align=center,
            xtick={4, 8, 12, 16},
            xticklabels={4, 8, 12, 16},
            yticklabels={}
        ]
        
            \addplot [draw=blue, fill=blue, forget plot, mark=*, only marks, mark size=1.5]
            table {%
                4.0000 40.7661
                8.0000 38.1875
                12.0000 36.9228
                16.0000 36.2511
            };
            \addplot [very thick, blue, dashed]
            table {%
                4.0000 40.7661
                8.0000 38.1875
                12.0000 36.9228
                16.0000 36.2511
            };
            \addlegendentry{$V_{\text{train}} = V_{1 \times}$}
            
            \addplot [draw=red, fill=red, forget plot, mark=*, only marks, mark size=1.5]
            table {%
                4.0000 41.8807
                8.0000 39.5402
                12.0000 38.2246
                16.0000 37.4802
            };
            \addplot [very thick, red, dashed]
            table {%
                4.0000 41.8807
                8.0000 39.5402
                12.0000 38.2246
                16.0000 37.4802
            };
            \addlegendentry{$V_{\text{train}} = V_{2 \times}$}
            
            \addplot [draw=orange, fill=orange, forget plot, mark=*, only marks, mark size=1.5]
            table {%
                4.0000 42.1862
                8.0000 39.9634
                12.0000 38.7077
                16.0000 37.9871
            };
            \addplot [very thick, orange, dashed]
            table {%
                4.0000 42.1862
                8.0000 39.9634
                12.0000 38.7077
                16.0000 37.9871
            };
            \addlegendentry{$V_{\text{train}} = V_{4 \times }$}
            
            \addplot [draw=purple, fill=purple, forget plot, mark=*, only marks, mark size=1.5]
            table {%
                4.0000 42.0268
                8.0000 39.8266
                12.0000 38.6255
                16.0000 37.9266
            };
            \addplot [very thick, purple]
            table {%
                4.0000 42.0268
                8.0000 39.8266
                12.0000 38.6255
                16.0000 37.9266
            };
            \addlegendentry{Diverse}
    
    \end{groupplot}

\end{tikzpicture}

%% file: figures_onecolumn/recon_at_lower_resolution_ext.tikz
\begin{tikzpicture}

    \definecolor{green}{RGB}{0,128,0}
    \definecolor{orange}{RGB}{255,165,0}
    \definecolor{purple}{RGB}{128,0,128}
    
    \newcommand{\pngWidthLowerRes}{0.12\textwidth}

    \node (img_volume_trilinear) {\reflectbox{\rotatebox[origin=c]{180}{\includegraphics[width=\pngWidthLowerRes]{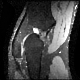}}}};
    \node [right=-0.2cm, at=(img_volume_trilinear.east)](img_volume_Gaussian) {\reflectbox{\rotatebox[origin=c]{180}{\includegraphics[width=\pngWidthLowerRes]{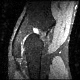}}}};
    \node [right=-0.2cm, at=(img_volume_Gaussian.east)](img_volume_inn) {\reflectbox{\rotatebox[origin=c]{180}{\includegraphics[width=\pngWidthLowerRes]{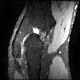}}}};
    \node [right=-0.15cm, at=(img_volume_inn.east)](img_volume_ref) {\reflectbox{\rotatebox[origin=c]{180}{\includegraphics[width=\pngWidthLowerRes]{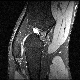}}}};
    
    \node[above=0.1pt, at=(img_volume_trilinear.north),font=\color{gray}, align=center] {\small \dashuline{Trilinear}};
    \node[above=0.1pt, at=(img_volume_Gaussian.north),font=\color{red}, align=center] {\small \dashuline{Gaussian}};
    \node[above=0.1pt, at=(img_volume_inn.north),font=\color{green}, align=center] {\small \dashuline{INN}};
    
    \node[above=0.1pt, yshift=0.1cm, at=(img_volume_ref.north),font=\color{black}, align=center] (labelRef) {\small Reference};
    
    \node[below=0.1pt, at=(img_volume_trilinear.south),font=\color{brown}, align=center] (labelZeroPadTrilinear) {\small \underline{Bilinear}};
    \node[below=0.1pt, at=(img_volume_Gaussian.south),font=\color{blue}, align=center] (labelZeroPadFourier) {\small \underline{Fourier}};
    \node[below=0.1pt, at=(img_volume_inn.south),font=\color{lime}, align=center] (labelZeroPadImg) {\small \underline{ZeroPad}};
    \node[below=0.1pt, at=(img_volume_ref.south),font=\color{purple}, align=center] (labelDiv) {\small \dotuline{Diverse}};

    \node [below=-0.05cm, at=(labelZeroPadTrilinear.south)](img_kernel_trilinear) {\reflectbox{\rotatebox[origin=c]{180}{\includegraphics[width=\pngWidthLowerRes]{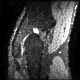}}}};
    \node [right=-0.2cm, at=(img_kernel_trilinear.east)](img_kernel_Fourier) {\reflectbox{\rotatebox[origin=c]{180}{\includegraphics[width=\pngWidthLowerRes]{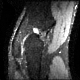}}}};
    \node [right=-0.2cm, at=(img_kernel_Fourier.east)](img_kernel_zerp_pad) {\reflectbox{\rotatebox[origin=c]{180}{\includegraphics[width=\pngWidthLowerRes]{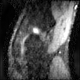}}}};
    
    \node [right=-0.15cm, at=(img_kernel_zerp_pad.east)](img_volume_div) {\reflectbox{\rotatebox[origin=c]{180}{\includegraphics[width=\pngWidthLowerRes]{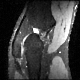}}}};

    \draw[gray] ($(img_volume_inn.north east)+(-0.07,0.6)$)  rectangle ($(img_kernel_zerp_pad.south east)+(-0.07,-0.05)$);

    \node[left=1pt, at=(img_volume_trilinear.west), node distance=0.1cm, rotate=90, anchor=center,yshift=0.1cm,xshift=0.0cm,font=\color{black}] {\small \dashuline{Volume int.}};
    
    \node[left=1pt, at=(img_kernel_trilinear.west), node distance=0.1cm, rotate=90, anchor=center,yshift=0.1cm,xshift=0.0cm,font=\color{black}] {\small \underline{Inf.-Res. DM}};

    \node[at=(img_kernel_zerp_pad.east), xshift=1.4cm, yshift=0.47cm]{
        
        \begin{groupplot}[group style={group size=1 by 1},
            width=0.29\columnwidth,
            height=0.29\columnwidth,
            tick align=outside,
            tick pos=left,
            x grid style={black},
            xtick style={color=black},
            y grid style={black},
            scaled ticks=false,
            ytick style={color=black},
            legend image code/.code={
                \draw[#1] (0cm,0.0cm) -- (0.4cm,0.0cm); 
            }, 
            legend style={
              fill opacity=0.8,
              font=\small,
              draw opacity=1,
              text opacity=1,
              at = {(0.74, 1.0)},
              anchor=north,
              draw=gray}
        ]
            
        \nextgroupplot[xlabel={acceleration factor $R$}, ylabel={PSNR (dB)}, xmin=3, xmax=17, ymin=36, ymax=43, title={Quantitative results}, align=center,
            xtick={4, 8, 12, 16},
            xticklabels={4, 8, 12, 16},
            ytick={37, 39, 41, 43},
            yticklabels={37,39,41,43}
        ]

            \addplot [draw=purple, fill=purple, forget plot, mark=*, only marks, mark size=1.5]
            table {%
                4.0000 42.1862
                8.0000 39.9634
                12.0000 38.7077
                16.0000 37.9871
            };
            \addplot [very thick, purple, dotted, forget plot]
            table {%
                4.0000 42.1862
                8.0000 39.9634
                12.0000 38.7077
                16.0000 37.9871
            };
            
            \addplot [draw=red, fill=red, forget plot, mark=*, only marks, mark size=1.5]
            table {%
                4.0000 39.7949
                8.0000 38.3942
                12.0000 37.2903
                16.0000 36.5774
            };
            \addplot [very thick, red, dashed, forget plot]
            table {%
                4.0000 39.7949
                8.0000 38.3942
                12.0000 37.2903
                16.0000 36.5774
            };

            \addplot[draw=gray, fill=gray, forget plot, mark=*, only marks, mark size=1.5] 
            table {%
                4.0 42.245
                8.0 40.0136
                12.0 38.65
                16.0 37.928
            };
            \addplot [very thick, gray, dashed, forget plot]
            table {%
                4.0 42.245
                8.0 40.0136
                12.0 38.65
                16.0 37.928
            };
            
            \addplot[draw=green, fill=green, forget plot, mark=*, only marks, mark size=1.5] 
            table {%
                4.0 41.4992
                8.0 39.3729
                12.0 38.1626
                16.0 37.4171
            };
            \addplot [very thick, green, dashed, forget plot]
            table {%
                4.0 41.4992
                8.0 39.3729
                12.0 38.1626
                16.0 37.4171
            };

            \addplot [draw=orange, fill=orange, forget plot, mark=*, only marks, mark size=1.5]
            table {%
                4.0000 41.8807
                8.0000 39.5402
                12.0000 38.2246
                16.0000 37.4802
            };
            \addplot [very thick, orange]
            table {%
                4.0000 41.8807
                8.0000 39.5402
                12.0000 38.2246
                16.0000 37.4802
            };
            \addlegendentry{Fixed}

            \addplot [draw=brown, fill=brown, forget plot, mark=*, only marks, mark size=1.5]
            table {%
                4.0000 41.805
                8.0000 39.08981
                12.0000 37.48902
                16.0000 36.62812
            };
            \addplot [very thick, brown, forget plot]
            table {%
                4.0000 41.805
                8.0000 39.08981
                12.0000 37.48902
                16.0000 36.62812
            };

            \addplot [draw=blue, fill=blue, forget plot, mark=*, only marks, mark size=1.5]
            table {%
                4.0000 42.30799
                8.0000 39.68783
                12.0000 38.07757
                16.0000 37.18916
            };
            \addplot [very thick, blue, forget plot]
            table {%
                4.0000 42.30799
                8.0000 39.68783
                12.0000 38.07757
                16.0000 37.18916
            };
    
        \end{groupplot}
    };

\end{tikzpicture}

%% file: figures_onecolumn/recon_at_higher_resolution_ext.tikz
\begin{tikzpicture}

    \definecolor{green}{RGB}{0,128,0}
    \definecolor{orange}{RGB}{255,165,0}
    \definecolor{purple}{RGB}{128,0,128}

    \newcommand{\pngWidthHigherRes}{0.12\textwidth}
    \newcommand{\horizDistance}{-0.2cm}

    \node (img_volume_trilinear) {\includegraphics[width=\pngWidthHigherRes]{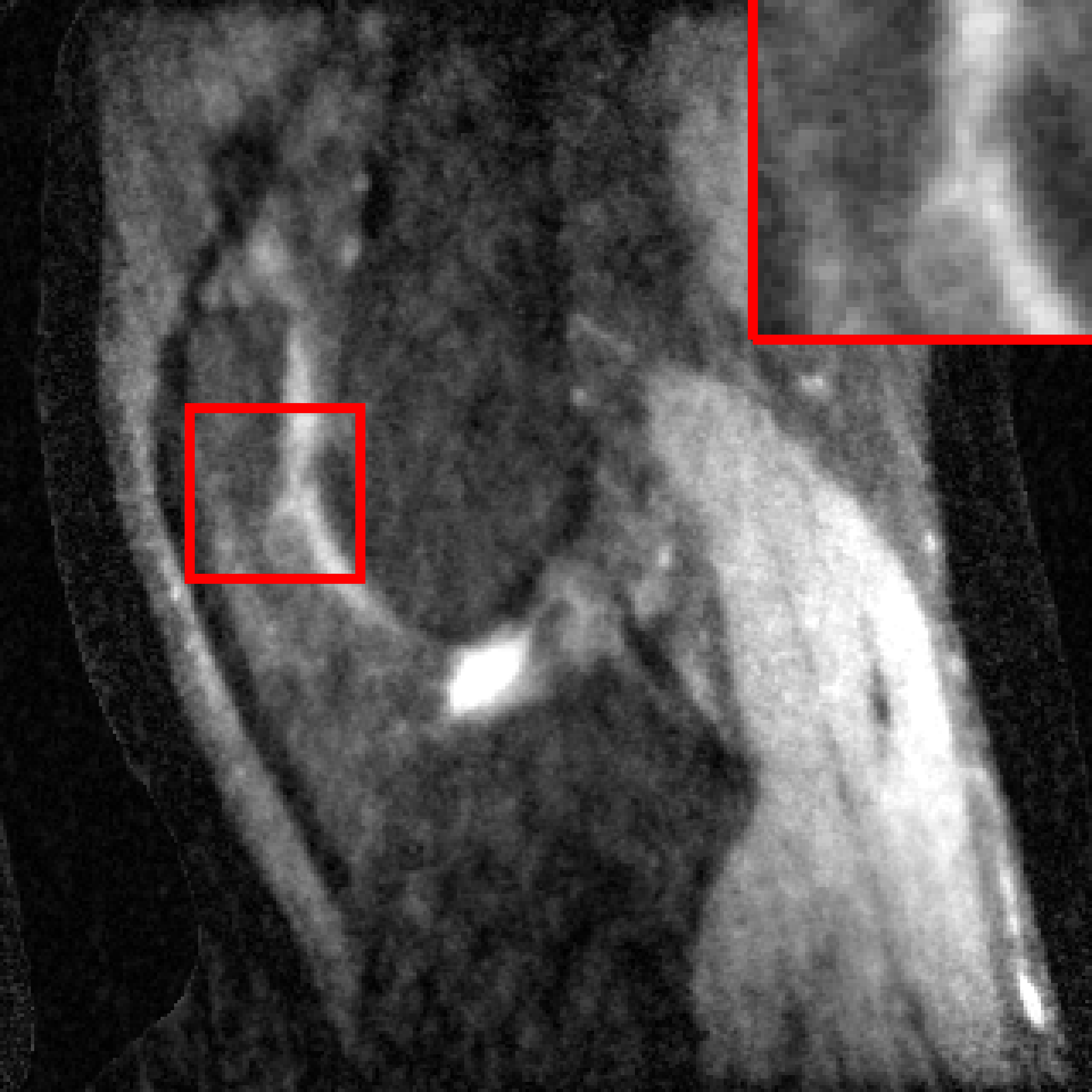}};
    \node [right=-0.2cm, at=(img_volume_trilinear.east)](img_volume_Gaussian) {\includegraphics[width=\pngWidthHigherRes]{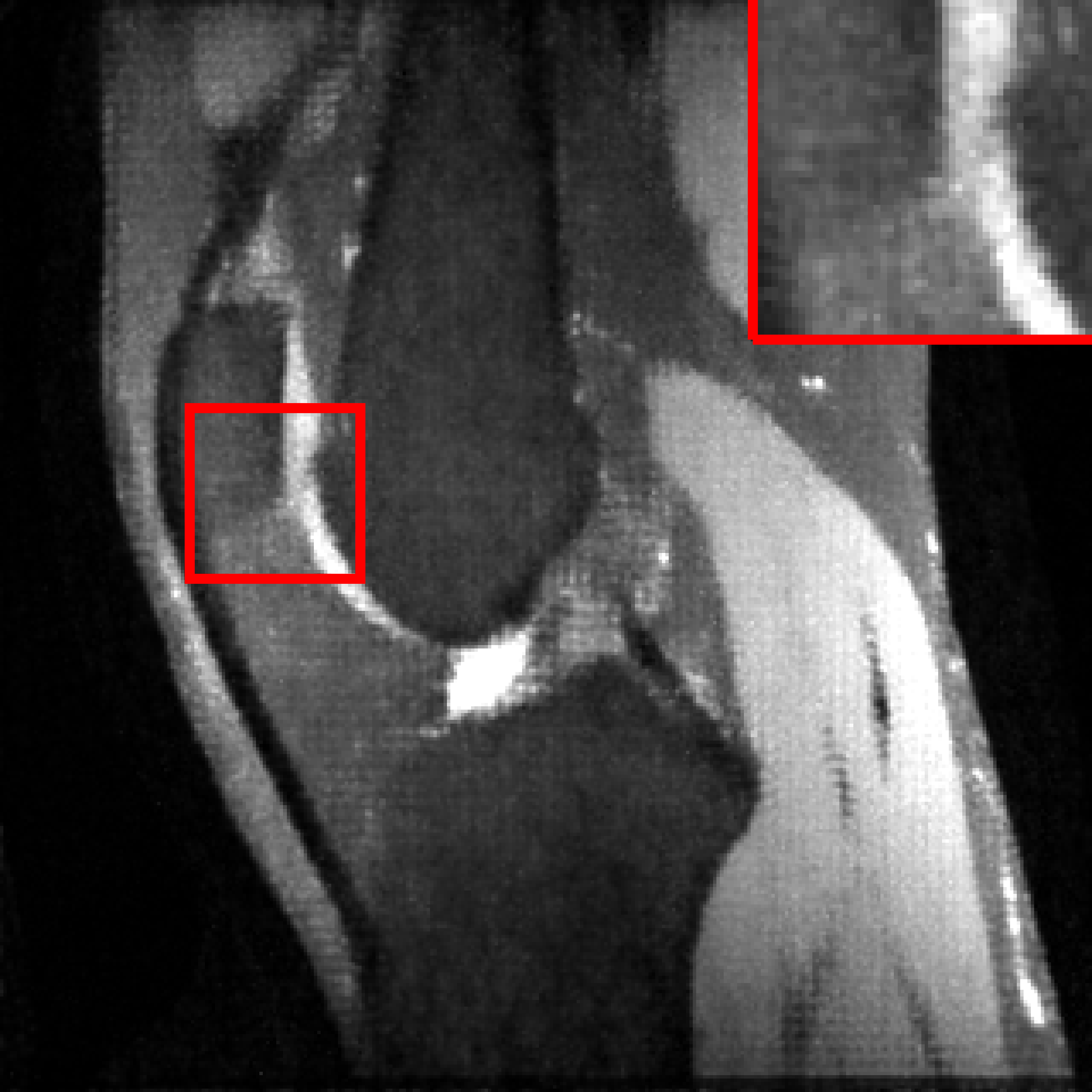}};
    \node [right=-0.2cm, at=(img_volume_Gaussian.east)](img_volume_nearest) {\includegraphics[width=\pngWidthHigherRes]{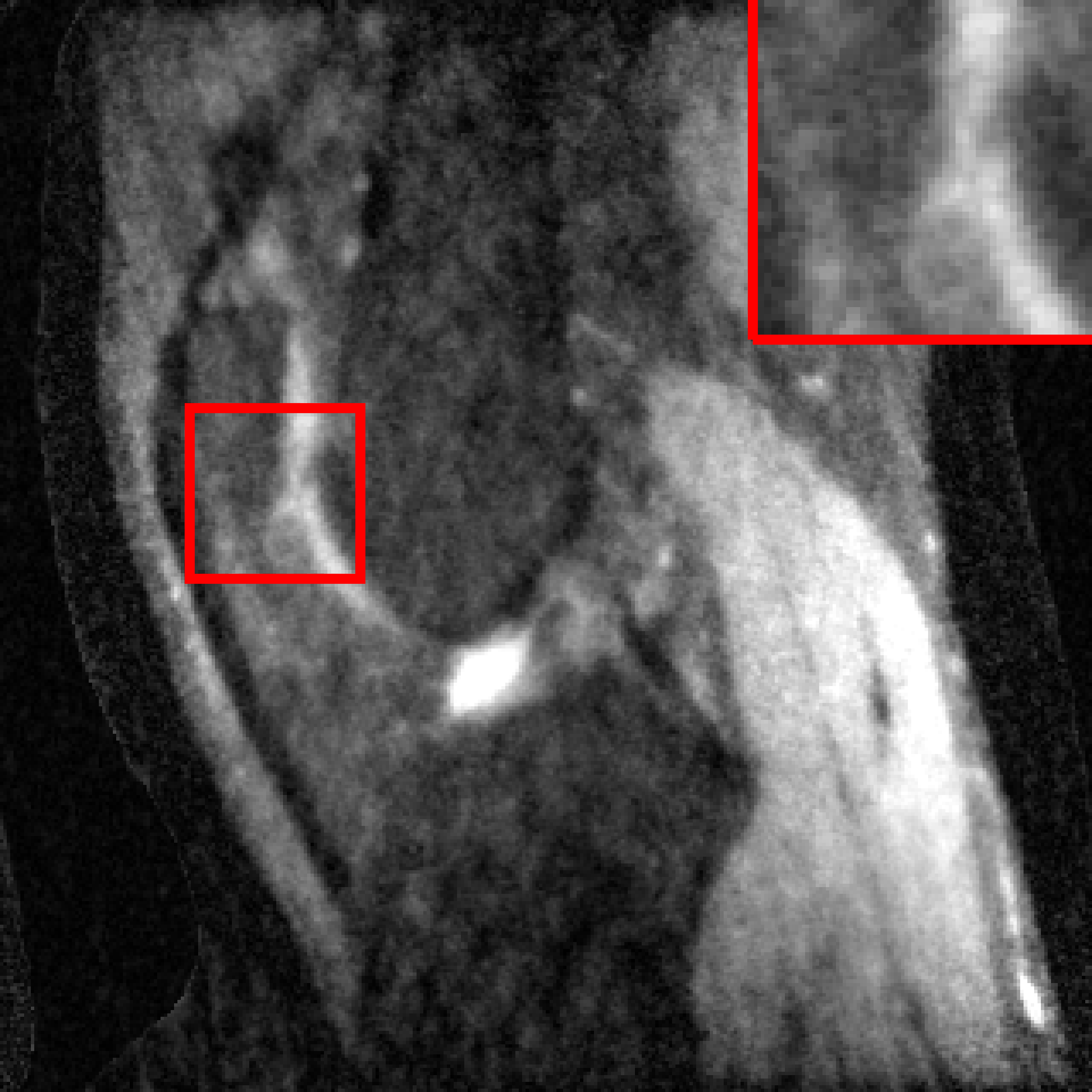}};
    \node [right=-0.15cm, at=(img_volume_nearest.east)](img_volume_ref) {\includegraphics[width=\pngWidthHigherRes]{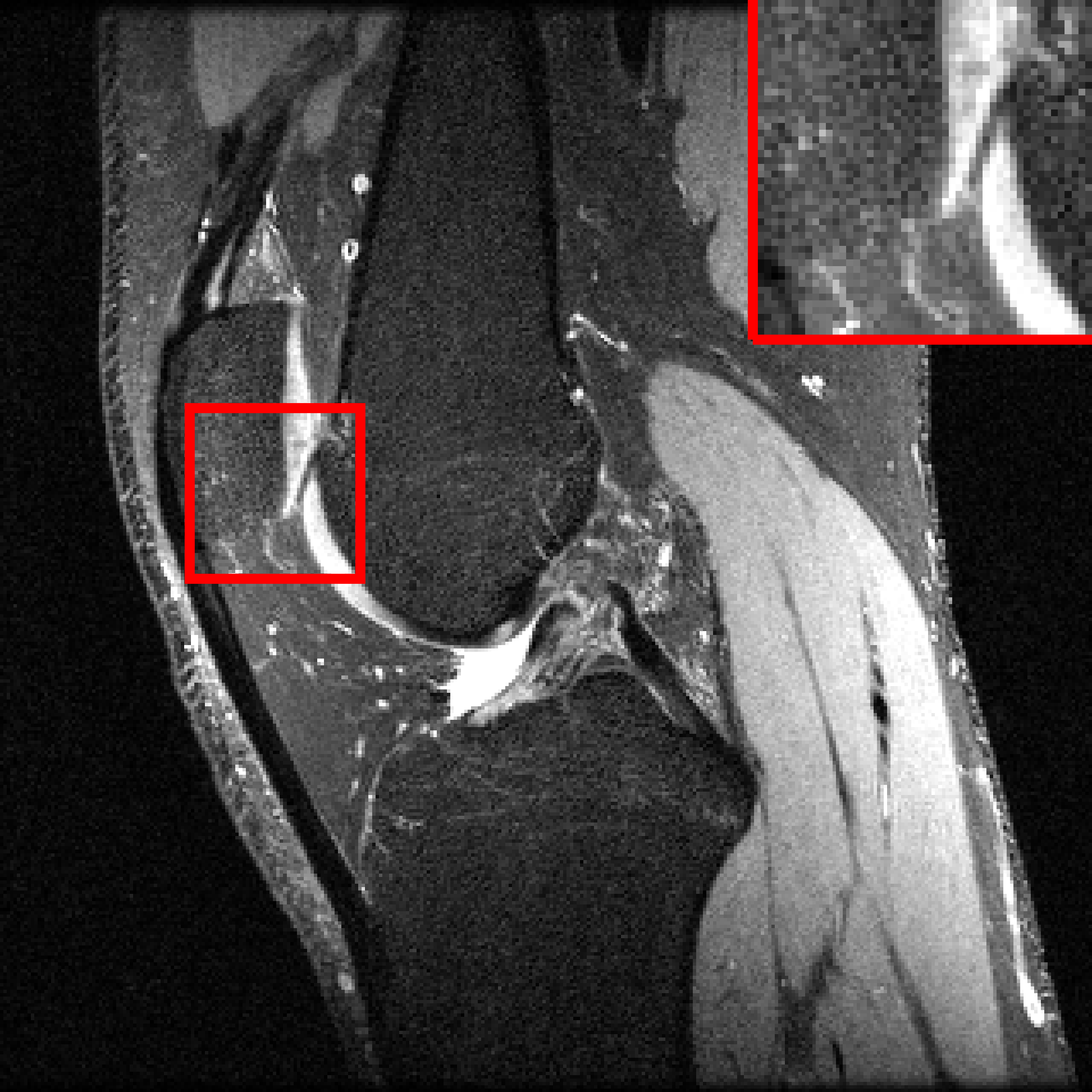}};

    \node[above=0.1pt, at=(img_volume_trilinear.north),font=\color{gray}, align=center] {\small \dashuline{Trilinear}};
    \node[above=0.1pt, at=(img_volume_Gaussian.north),font=\color{red}, align=center] {\small \dashuline{Gaussian}};
    \node[above=0.1pt, at=(img_volume_nearest.north),font=\color{green}, align=center] {\small \dashuline{Nearest}};
    \node[above=0.1pt, yshift=0.1cm, at=(img_volume_ref.north),font=\color{black}, align=center] (labelZeroPadImg) {\small Reference};

    \node[below=0.1pt, at=(img_volume_trilinear.south),font=\color{brown}, align=center] (labelZeroPadTrilinear) {\small \underline{Bilinear}};
    \node[below=0.1pt, at=(img_volume_Gaussian.south),font=\color{blue}, align=center] (labelZeroPadFourier) {\small \underline{Fourier}};
    \node[below=0.1pt, at=(img_volume_nearest.south),font=\color{lime}, align=center] (labelZeroPadImg) {\small \underline{ZeroPad}};
    \node[below=0.1pt, at=(img_volume_ref.south),font=\color{purple}, align=center] (labelZeroPadImg) {\small \dotuline{Diverse}};
    
    \node [below=-0.05cm, at=(labelZeroPadTrilinear.south)](img_kernel_trilinear) {\includegraphics[width=\pngWidthHigherRes]{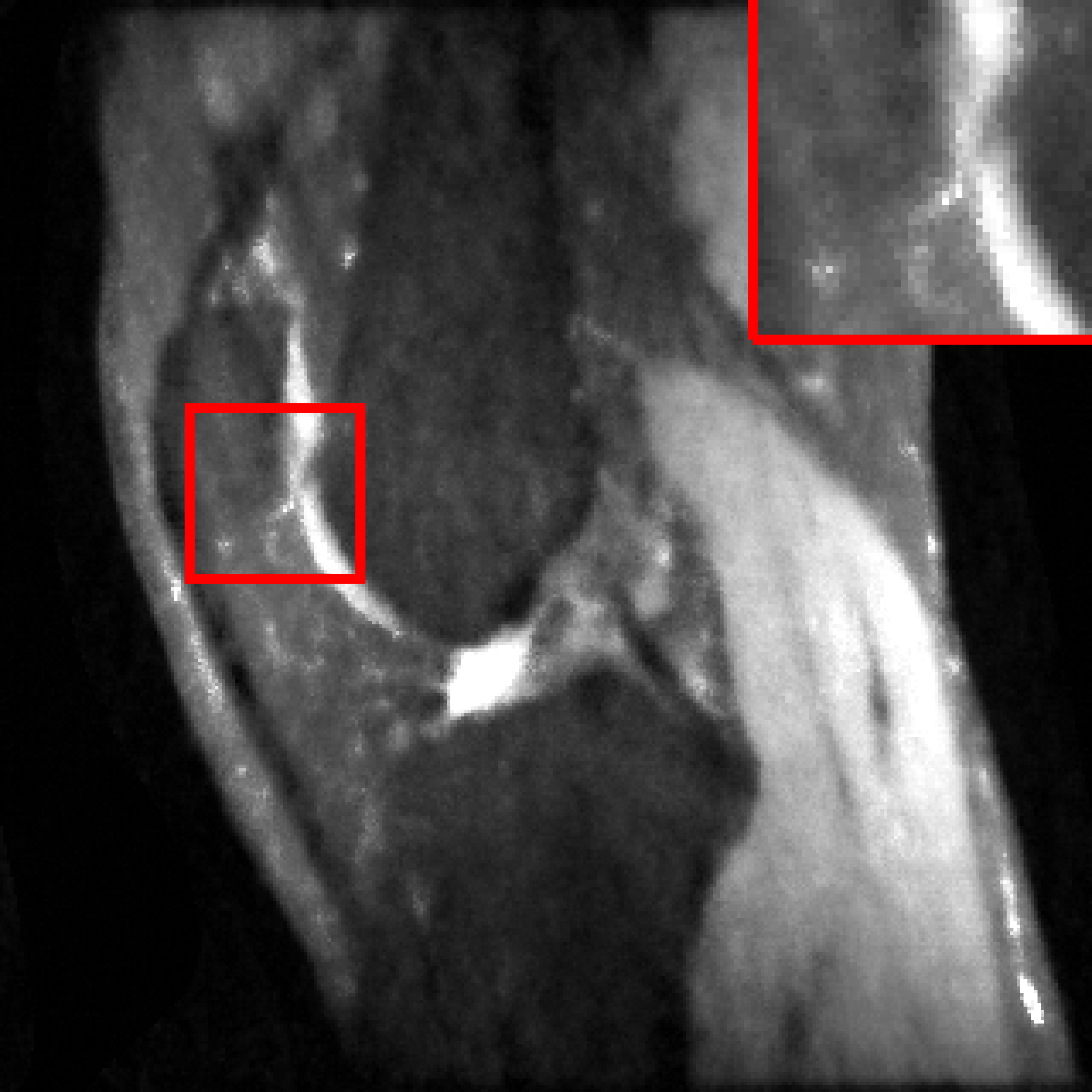}};
    \node [right=-0.2cm, at=(img_kernel_trilinear.east)](img_kernel_Fourier) {\includegraphics[width=\pngWidthHigherRes]{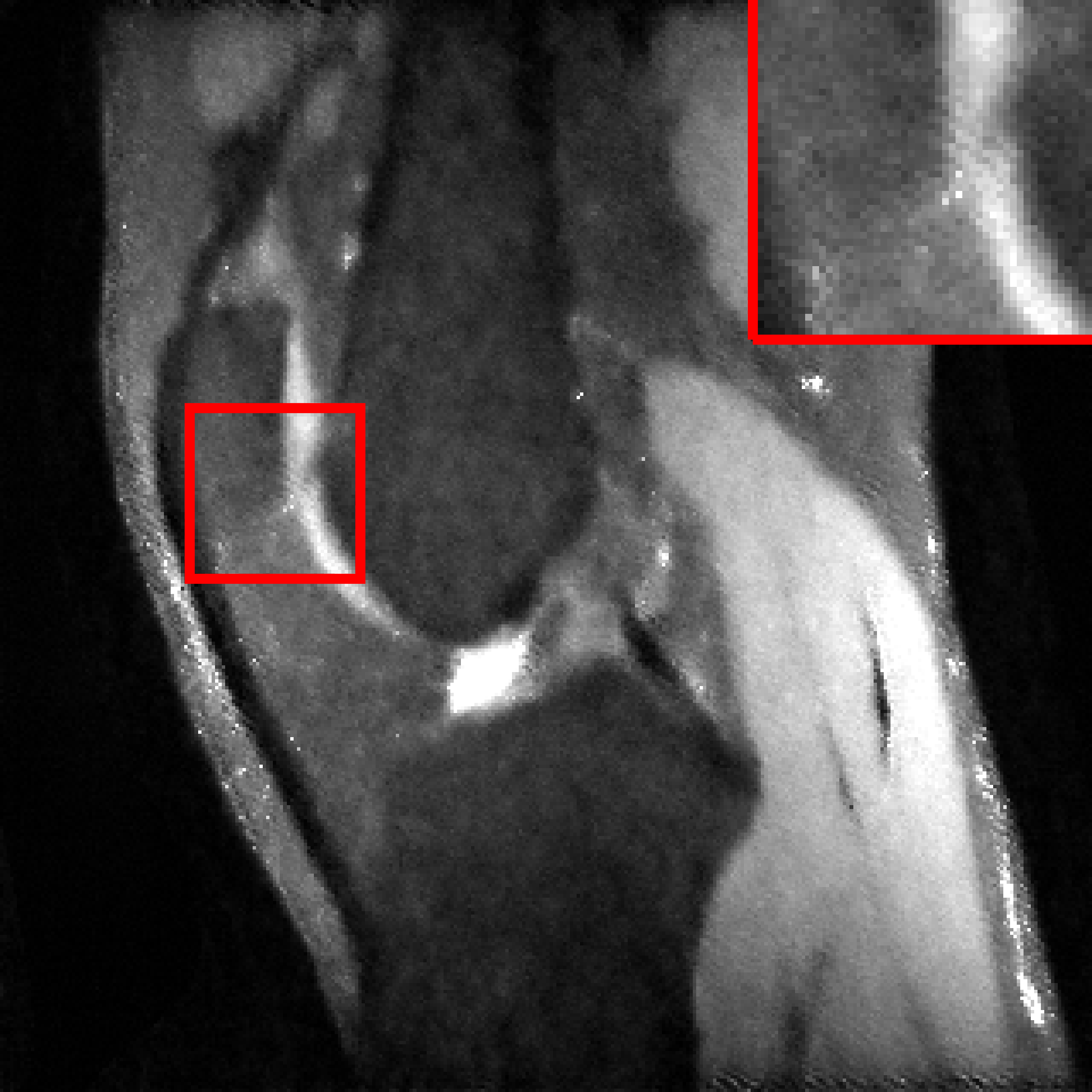}};
    
    \node [right=-0.2cm, at=(img_kernel_Fourier.east)](img_kernel_zerp_pad) {\includegraphics[width=\pngWidthHigherRes]{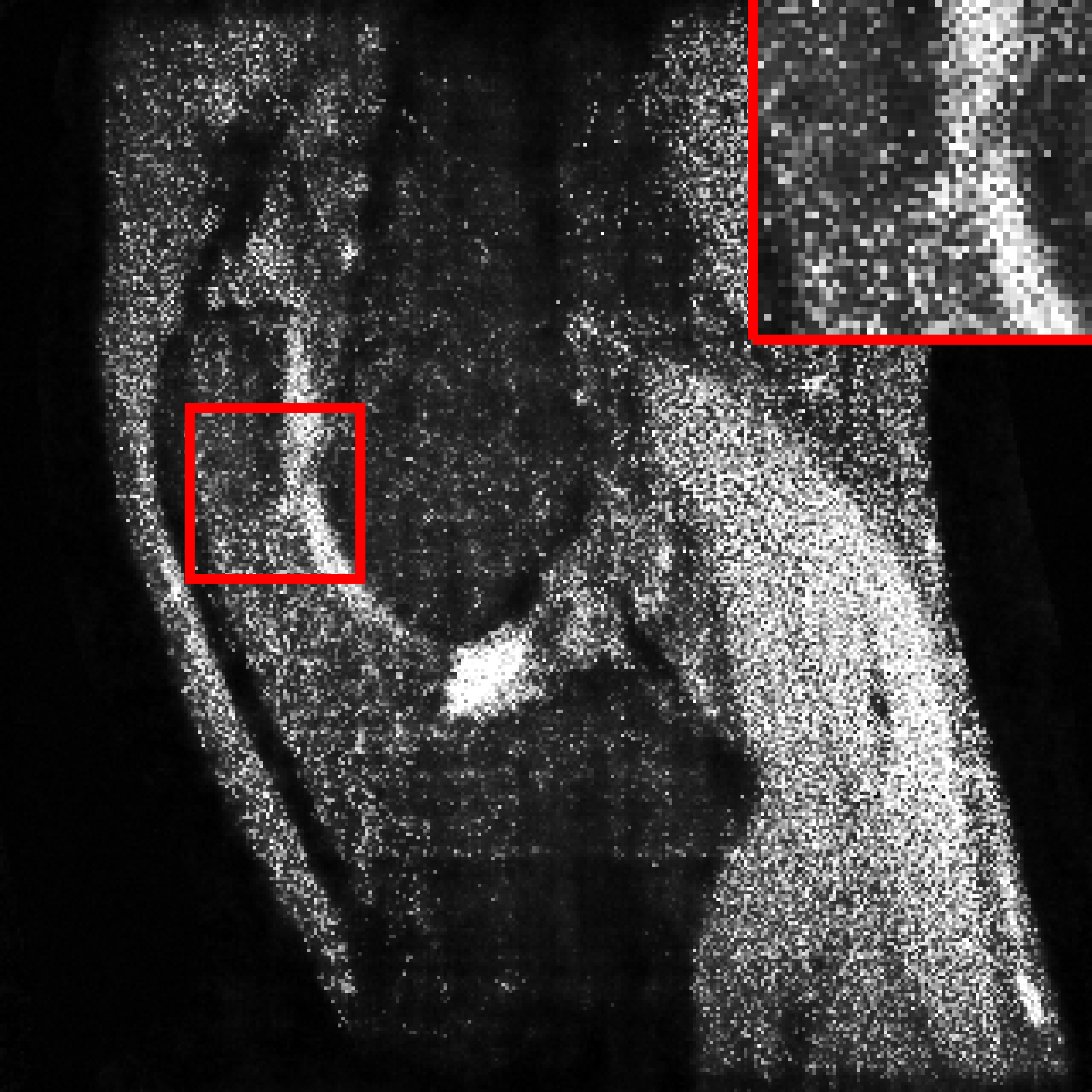}};

    \draw[black] ($(img_volume_nearest.north east)+(-0.07,0.6)$)  rectangle ($(img_kernel_zerp_pad.south east)+(-0.07,-0.05)$);
    \node [right=-0.15cm, at=(img_kernel_zerp_pad.east)](img_kernel_diversed) {\includegraphics[width=\pngWidthHigherRes]{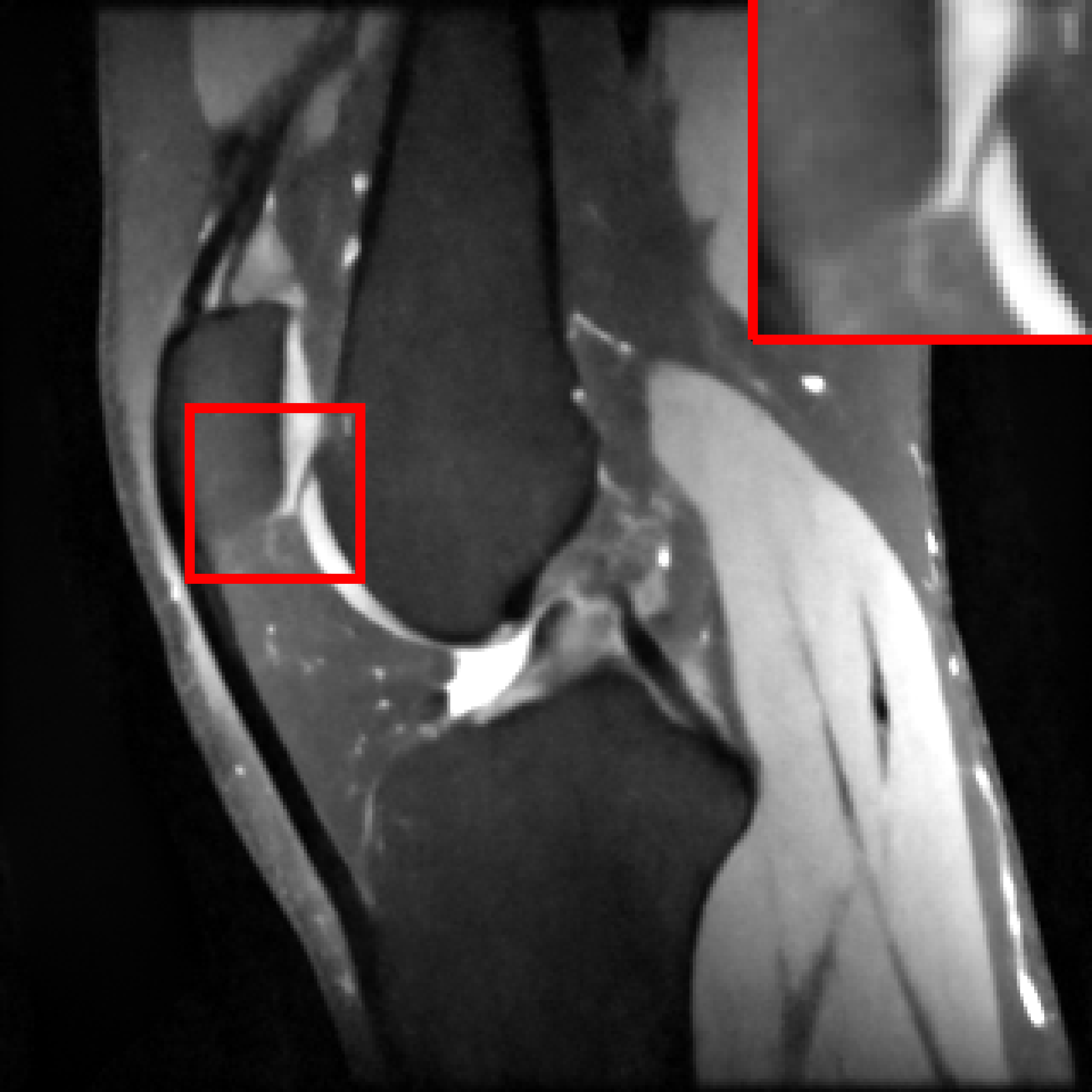}};

    \node[left=1pt, at=(img_volume_trilinear.west), node distance=0.1cm, rotate=90, anchor=center,yshift=0.1cm,xshift=0.0cm,font=\color{black}] {\small \dashuline{Volume int.}};
    
    \node[left=1pt, at=(img_kernel_trilinear.west), node distance=0.1cm, rotate=90, anchor=center,yshift=0.1cm,xshift=0.0cm,font=\color{black}] {\small \underline{Inf.-Res. DM}};

    \node[at=(img_kernel_zerp_pad.east), xshift=1.4cm, yshift=0.47cm]{
        
        \begin{groupplot}[group style={group size=1 by 1},
            width=0.29\columnwidth,
            height=0.29\columnwidth,
            tick align=outside,
            tick pos=left,
            x grid style={darkgray176},
            xtick style={color=black},
            y grid style={darkgray176},
            scaled ticks=false,
            ytick style={color=black},
            legend image code/.code={
                \draw[#1] (0cm,0.0cm) -- (0.4cm,0.0cm); 
            }, 
            legend style={
              fill opacity=0.8,
              font=\small,
              draw opacity=1,
              text opacity=1,
              at = {(0.74, 1.0)},
              anchor=north,
              draw=gray}
        ]

        \nextgroupplot[ylabel={PSNR (dB)}, xlabel={acceleration factor $R$}, xmin=1, xmax=51, ymin=35, ymax=43, title={Quantitative results},
            xtick={4, 12, 24, 36, 48},
            xticklabels={4, 12, 24, 36, 48},
            ytick={35, 37, 39, 41, 43},
            yticklabels={35, 37, 39, 41, 43}
        ]

            \addplot [draw=purple, fill=purple, forget plot, mark=*, only marks, mark size=1.5]
            table {%
            4 41.892
            8 41.026
            12 40.593
            16 40.286
            24 39.891
            36 39.508
            48 39.314
            };
            \addplot [very thick, purple, dotted, forget plot]
            table {%
            4 41.892
            8 41.026
            16 40.286
            24 39.891
            36 39.508
            48 39.314
            };
            
            \addplot [draw=red, fill=red, forget plot, mark=*, only marks, mark size=1.5]
            table {%
            4 41.484
            8 40.488
            12 39.816
            16 39.403
            24 38.94
            36 38.424
            48 38.153
            };
            \addplot [very thick, red, dashed, forget plot]
            table {%
            4 41.484
            8 40.488
            12 39.816
            16 39.403
            24 38.94
            36 38.424
            48 38.153
            };
            
            \addplot [draw=gray, fill=gray, forget plot, mark=*, only marks, mark size=1.5]
            table {%
                4 39.408
                8 38.3
                12 37.731
                16 37.628
                24 36.98
                36 36.208
                48 35.884                
            };
            \addplot [very thick, gray, dashed, forget plot]
            table {%
                4 39.408
                8 38.3
                12 37.731
                16 37.628
                24 36.98
                36 36.208
                48 35.884                
            };

            \addplot [draw=green, fill=green, forget plot, mark=*, only marks, mark size=1.5]
            table {%
            4 39.26
            8 38.222
            12 37.7
            16 37.596
            24 36.9
            36 36.336
            48 36.02
            };
            \addplot [very thick, green, dashed, forget plot]
            table {%
            4 39.26
            8 38.222
            12 37.7
            16 37.596
            24 36.9
            36 36.336
            48 36.02
            };

            \addplot [draw=orange, fill=orange, forget plot, mark=*, only marks, mark size=1.5]
            table {%
            4 41.926
            8 40.829
            16 40.089
            24 39.414
            36 38.972
            48 38.5
            };
            \addplot [very thick, orange]
            table {%
            4 41.926
            8 40.829
            16 40.089
            24 39.414
            36 38.972
            48 38.5
            };
            \addlegendentry{Fixed}

             \addplot [draw=blue, fill=blue, forget plot, mark=*, only marks, mark size=1.5]
            table {%
                4 41.7879
                8 40.41799
                12 39.7262
                16 39.2855
                24 38.5595
                36 37.967
                48 37.644
            };
            \addplot [very thick, blue, forget plot]
            table {%
                4 41.7879
                8 40.41799
                12 39.7262
                16 39.2855
                24 38.5595
                36 37.967
                48 37.644
            };
        
            \addplot [draw=brown, fill=brown, forget plot, mark=*, only marks, mark size=1.5]
            table {%
                4 41.8891
                8 40.5543
                12 39.87
                16 39.45
                24 38.75
                36 38.1560
                48 37.8307
            };
            \addplot [very thick, brown, forget plot]
            table {%
                4 41.8891
                8 40.5543
                12 39.87
                16 39.45
                24 38.75
                36 38.1560
                48 37.8307
            };
        \end{groupplot}
    };

\end{tikzpicture}

%% file: figures_onecolumn_appendix/ssim_in_distr_comparision.tikz
\begin{tikzpicture}

    \definecolor{green}{RGB}{0,128,0}
    \definecolor{orange}{RGB}{255,165,0}
    \definecolor{purple}{RGB}{128,0,128}

    \begin{groupplot}[group style={group size=1 by 1,
            horizontal sep=10pt},
            width=0.28\textwidth,
            height=0.28\textwidth,
            tick align=outside,
            tick pos=left,
            x grid style={darkgray176},
            xtick style={color=black},
            y grid style={darkgray176},
            scaled ticks=false,
            ylabel={SSIM (sagittal)},
            xlabel={acceleration $R$},
            ytick style={color=black},
            xmin=1, xmax=51,
            ymin=0.82, ymax=0.96,
            ytick={0.83, 0.86, 0.89, 0.92, 0.95},
            yticklabels={0.83, 0.86, 0.89, 0.92, 0.95},
            xtick={4, 12, 24, 36, 48},
            xticklabels={4, 12, 24, 36, 48},
            legend image code/.code={
                \draw[#1] (0cm,0.0cm) -- (0.3cm,0.0cm); 
            }, 
            legend style={
              legend columns=1,
              fill opacity=0.8,
              font=\small,
              draw opacity=1,
              text opacity=1,
              at={(1.7, 1.0)},
              anchor=north,
              draw=gray}
        ]

        \nextgroupplot[
            ytick style={color=black},
            title={In-distribution performance}
        ]
        
            \addplot [draw=blue, fill=blue, forget plot, mark=*, only marks, mark size=1.5]
            table {%
                4 0.943
                8 0.9231
                12 0.911
                16 0.9019
                24 0.892
                36 0.8808
                48 0.8713
            };
            \addplot [very thick, blue]
            table {%
                4 0.943
                8 0.9231
                12 0.911
                16 0.9019
                24 0.892
                36 0.8808
                48 0.8713
            };
            \addlegendentry{Variational}
        
            \addplot [draw=red, fill=red, forget plot, mark=*, only marks, mark size=1.5]
            table {%
                4 0.9058
                8 0.8907
                12 0.8825
                16 0.8745
                24 0.8684
                36 0.8611
                48 0.8545
            };
            \addplot [very thick, red]
            table {%
                4 0.9058
                8 0.8907
                12 0.8825
                16 0.8745
                24 0.8684
                36 0.8611
                48 0.8545
            };
            \addlegendentry{Sampling}

            \addplot [draw=orange, fill=orange, forget plot, mark=*, only marks, mark size=1.5]
            table {%
                4 0.8988
                8 0.8748
                12 0.8684
                16 0.8644
                24 0.855
                36 0.8444
                48 0.8355
            };
            \addplot [very thick, orange]
            table {%
                4 0.8988
                8 0.8748
                12 0.8684
                16 0.8644
                24 0.855
                36 0.8444
                48 0.8355
            };
            \addlegendentry{Classical}

    \end{groupplot}

\end{tikzpicture}

%% file: figures_onecolumn_appendix/noise_loss_Tprime.tikz
\begin{tikzpicture}

    \definecolor{green}{RGB}{0,128,0}
    \definecolor{orange}{RGB}{255,165,0}
    \definecolor{purple}{RGB}{128,0,128}

    \begin{groupplot}[group style={group size=1 by 1,
            horizontal sep=30pt},
            width=0.6\columnwidth,
            height=0.6\columnwidth,
            tick align=outside,
            tick pos=left,
            x grid style={darkgray176},
            xtick style={color=black},
            y grid style={darkgray176},
            scaled ticks=false,
            ylabel={SSIM (sagittal)},
            xlabel={acceleration factor $R$},
            ytick style={color=black},
            xmin=0, xmax=1,
            legend cell align={left},
            legend image code/.code={
                \draw[#1] (0cm,0.0cm) -- (0.3cm,0.0cm); 
            }, 
            legend style={
              fill opacity=0.8,
              font=\small,
             draw opacity=1,
              text opacity=1,
              at={(1.5, 1.0)},
              anchor=north,
              draw=gray}
        ]

        \nextgroupplot[title={Timestep sampling $t \sim \mathcal{U}(0, T')$}, xlabel={Maximum timestep $T'$}, ylabel={PSNR (dB)},
        xtick={0.1, 0.4, 0.7, 1.0},
        xticklabels={100, 400, 700, 1000}
        ]

	\addplot [draw=orange,fill=orange, forget plot, mark=*, only marks, mark size=1.5]
	table {
	0.1000 39.3838
	0.2000 40.8001
	0.3000 40.9663
	0.4000 40.9797
	0.5000 40.8869
	0.6000 40.8345
	0.7000 40.6673
	0.8000 40.4886
	0.9000 39.8944
	1.0000 39.4251
	};
	\addplot [very thick, orange]
	table {
	0.1000 39.3838
	0.2000 40.8001
	0.3000 40.9663
	0.4000 40.9797
	0.5000 40.8869
	0.6000 40.8345
	0.7000 40.6673
	0.8000 40.4886
	0.9000 39.8944
	1.0000 39.4251
	};
	\addlegendentry{$12 \times$-acc.};
	\path [draw=orange, fill=orange, opacity=0.2]
	(axis cs:0.1000, 39.4015)
	--(axis cs:0.2000, 40.8148)
	--(axis cs:0.3000, 40.9816)
	--(axis cs:0.4000, 40.9901)
	--(axis cs:0.5000, 40.9310)
	--(axis cs:0.6000, 40.8572)
	--(axis cs:0.7000, 40.7195)
	--(axis cs:0.8000, 40.5017)
	--(axis cs:0.9000, 40.1158)
	--(axis cs:1.0000, 39.4415)
	--(axis cs:1.0000, 39.4088)
	--(axis cs:0.9000, 39.6730)
	--(axis cs:0.8000, 40.4755)
	--(axis cs:0.7000, 40.6151)
	--(axis cs:0.6000, 40.8117)
	--(axis cs:0.5000, 40.8428)
	--(axis cs:0.4000, 40.9692)
	--(axis cs:0.3000, 40.9511)
	--(axis cs:0.2000, 40.7854)
	--(axis cs:0.1000, 39.3660)
	--cycle;
	\addplot [draw=blue,fill=blue, forget plot, mark=*, only marks, mark size=1.5]
	table {
	0.1000 38.0581
	0.2000 39.1258
	0.3000 39.3006
	0.4000 39.2703
	0.5000 39.2151
	0.6000 39.1163
	0.7000 38.9338
	0.8000 38.6754
	0.9000 38.3454
	1.0000 37.8172
	};
	\addplot [very thick, blue]
	table {
	0.1000 38.0581
	0.2000 39.1258
	0.3000 39.3006
	0.4000 39.2703
	0.5000 39.2151
	0.6000 39.1163
	0.7000 38.9338
	0.8000 38.6754
	0.9000 38.3454
	1.0000 37.8172
	};
	\addlegendentry{$48 \times$-acc.};
	\path [draw=blue, fill=blue, opacity=0.2]
	(axis cs:0.1000, 38.1570)
	--(axis cs:0.2000, 39.2184)
	--(axis cs:0.3000, 39.3544)
	--(axis cs:0.4000, 39.2958)
	--(axis cs:0.5000, 39.2424)
	--(axis cs:0.6000, 39.1370)
	--(axis cs:0.7000, 38.9659)
	--(axis cs:0.8000, 38.7136)
	--(axis cs:0.9000, 38.4073)
	--(axis cs:1.0000, 37.8542)
	--(axis cs:1.0000, 37.7803)
	--(axis cs:0.9000, 38.2834)
	--(axis cs:0.8000, 38.6371)
	--(axis cs:0.7000, 38.9017)
	--(axis cs:0.6000, 39.0955)
	--(axis cs:0.5000, 39.1878)
	--(axis cs:0.4000, 39.2449)
	--(axis cs:0.3000, 39.2468)
	--(axis cs:0.2000, 39.0332)
	--(axis cs:0.1000, 37.9591)
	--cycle;
        
    \end{groupplot}

\end{tikzpicture}

%% file: figures_onecolumn_appendix/slice_budget.tikz
\begin{tikzpicture}

    \definecolor{green}{RGB}{0,128,0}
    \definecolor{orange}{RGB}{255,165,0}
    \definecolor{purple}{RGB}{128,0,128}

    \begin{groupplot}[group style={group size=1 by 1,
            horizontal sep=30pt},
            width=0.6\columnwidth,
            height=0.6\columnwidth,
            tick align=outside,
            tick pos=left,
            x grid style={darkgray176},
            xtick style={color=black},
            y grid style={darkgray176},
            scaled ticks=false,
            ylabel={SSIM (sagittal)},
            xlabel={acceleration factor $R$},
            ytick style={color=black},
            legend cell align={left},
            legend image code/.code={
                \draw[#1] (0cm,0.0cm) -- (0.3cm,0.0cm); 
            }, 
            legend style={
              fill opacity=0.8,
              font=\small,
             draw opacity=1,
              text opacity=1,
              at={(1.5, 1.0)},
              anchor=north,
              draw=gray}
        ]

        \nextgroupplot[align=center, title={Nr. regularized slices $S$}, xlabel={Slice budget $S$}, ylabel={PSNR (dB)},
            xtick={5, 25, 50, 75, 100},
            xticklabels={5, 25, 50, 75, 100},
            ytick={34, 35, 36, 37, 38, 39},
            yticklabels={34, 35, 36, 37, 38, 39}
        ]

	\addplot [draw=orange,fill=orange, forget plot, mark=*, only marks, mark size=1.5]
	table {
	5.0000 33.6618
	10.0000 37.9992
	15.0000 38.6024
	20.0000 38.7696
	25.0000 38.8714
	30.0000 39.0777
	35.0000 39.1231
	40.0000 39.0944
	45.0000 39.1911
	50.0000 39.2258
	55.0000 39.1858
	60.0000 39.2675
	65.0000 39.2757
	70.0000 39.2649
	75.0000 39.2811
	80.0000 39.2778
	85.0000 39.3345
	90.0000 39.3070
	95.0000 39.3035
	100.0000 39.3319
	};

	\addplot [very thick, orange]
	table {
	5.0000 33.6618
	10.0000 37.9992
	15.0000 38.6024
	20.0000 38.7696
	25.0000 38.8714
	30.0000 39.0777
	35.0000 39.1231
	40.0000 39.0944
	45.0000 39.1911
	50.0000 39.2258
	55.0000 39.1858
	60.0000 39.2675
	65.0000 39.2757
	70.0000 39.2649
	75.0000 39.2811
	80.0000 39.2778
	85.0000 39.3345
	90.0000 39.3070
	95.0000 39.3035
	100.0000 39.3319
	};
	\addlegendentry{$48 \times$ acc.};
	\path [draw=orange, fill=orange, opacity=0.2]
	(axis cs:5.0000, 34.3137)
	--(axis cs:10.0000, 38.1500)
	--(axis cs:15.0000, 38.6461)
	--(axis cs:20.0000, 38.8207)
	--(axis cs:25.0000, 38.9395)
	--(axis cs:30.0000, 39.0995)
	--(axis cs:35.0000, 39.1396)
	--(axis cs:40.0000, 39.1689)
	--(axis cs:45.0000, 39.1966)
	--(axis cs:50.0000, 39.2363)
	--(axis cs:55.0000, 39.2394)
	--(axis cs:60.0000, 39.2927)
	--(axis cs:65.0000, 39.2985)
	--(axis cs:70.0000, 39.3098)
	--(axis cs:75.0000, 39.2861)
	--(axis cs:80.0000, 39.3396)
	--(axis cs:85.0000, 39.3661)
	--(axis cs:90.0000, 39.3451)
	--(axis cs:95.0000, 39.4026)
	--(axis cs:100.0000, 39.3735)
	--(axis cs:100.0000, 39.2903)
	--(axis cs:95.0000, 39.2045)
	--(axis cs:90.0000, 39.2689)
	--(axis cs:85.0000, 39.3030)
	--(axis cs:80.0000, 39.2160)
	--(axis cs:75.0000, 39.2760)
	--(axis cs:70.0000, 39.2199)
	--(axis cs:65.0000, 39.2530)
	--(axis cs:60.0000, 39.2422)
	--(axis cs:55.0000, 39.1323)
	--(axis cs:50.0000, 39.2152)
	--(axis cs:45.0000, 39.1856)
	--(axis cs:40.0000, 39.0199)
	--(axis cs:35.0000, 39.1067)
	--(axis cs:30.0000, 39.0559)
	--(axis cs:25.0000, 38.8034)
	--(axis cs:20.0000, 38.7185)
	--(axis cs:15.0000, 38.5586)
	--(axis cs:10.0000, 37.8484)
	--(axis cs:5.0000, 33.0098)
	--cycle;

    \addplot[draw=red,fill=red, mark=*, mark size=2.0]
    table {
	   50.0000 39.2258
    };
    \addlegendentry{Selected}
        
    \end{groupplot}

\end{tikzpicture}

%% file: figures_onecolumn_appendix/st3_shift_pngs__sensitivity.tikz
\begin{tikzpicture}

    \definecolor{green}{RGB}{0,128,0}
    \definecolor{orange}{RGB}{255,165,0}
    \definecolor{purple}{RGB}{128,0,128}

    \newcommand{\pngwidth}{0.13\textwidth}

    \node (img_12x_same) {\reflectbox{\rotatebox[origin=c]{180}{\includegraphics[width=\pngwidth]{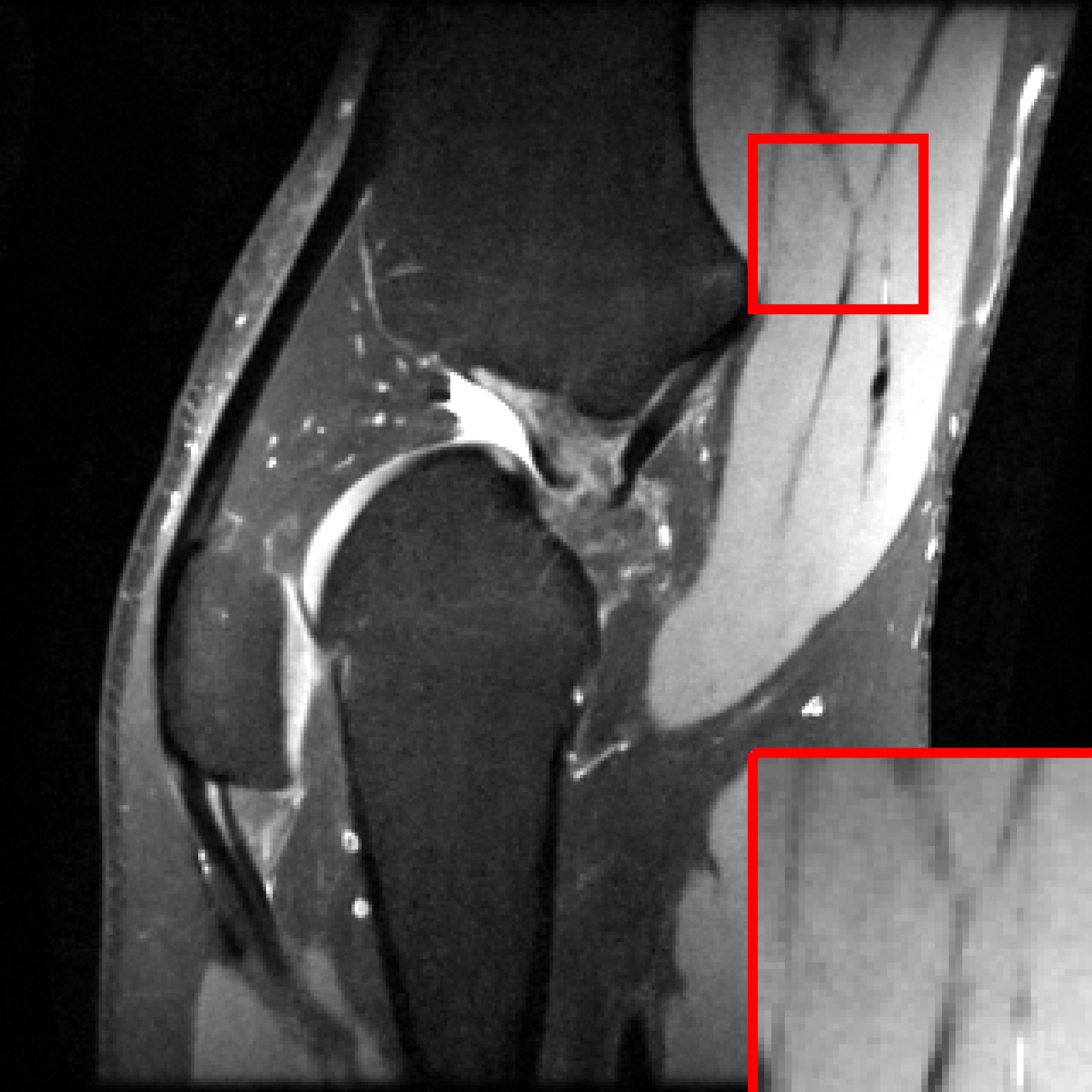}}}};
    \node [right=-0.2cm, at=(img_12x_same.east)](img_12x_2x) {\reflectbox{\rotatebox[origin=c]{180}{\includegraphics[width=\pngwidth]{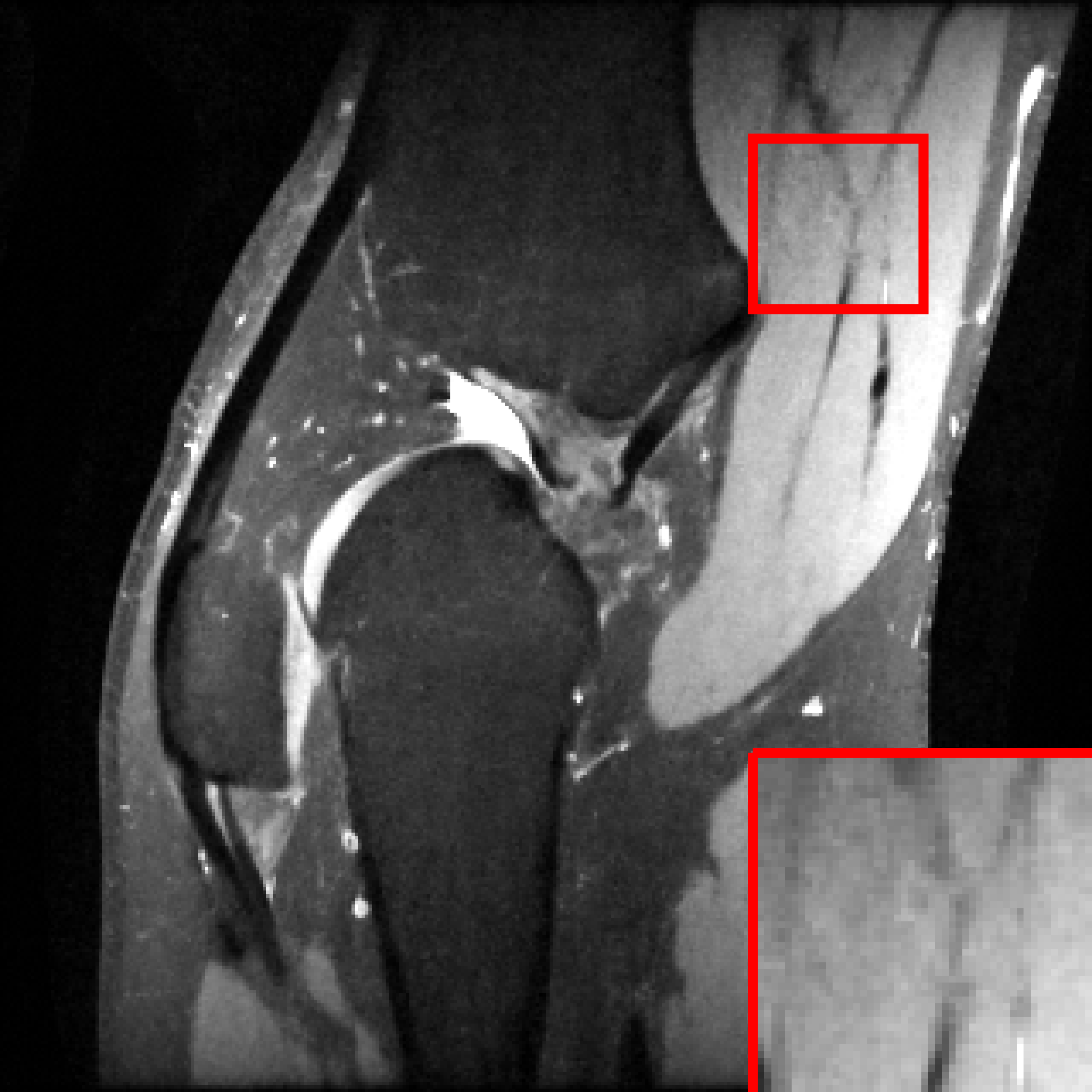}}}};
    \node [right=-0.2cm, at=(img_12x_2x.east)](img_12x_4x) {\reflectbox{\rotatebox[origin=c]{180}{\includegraphics[width=\pngwidth]{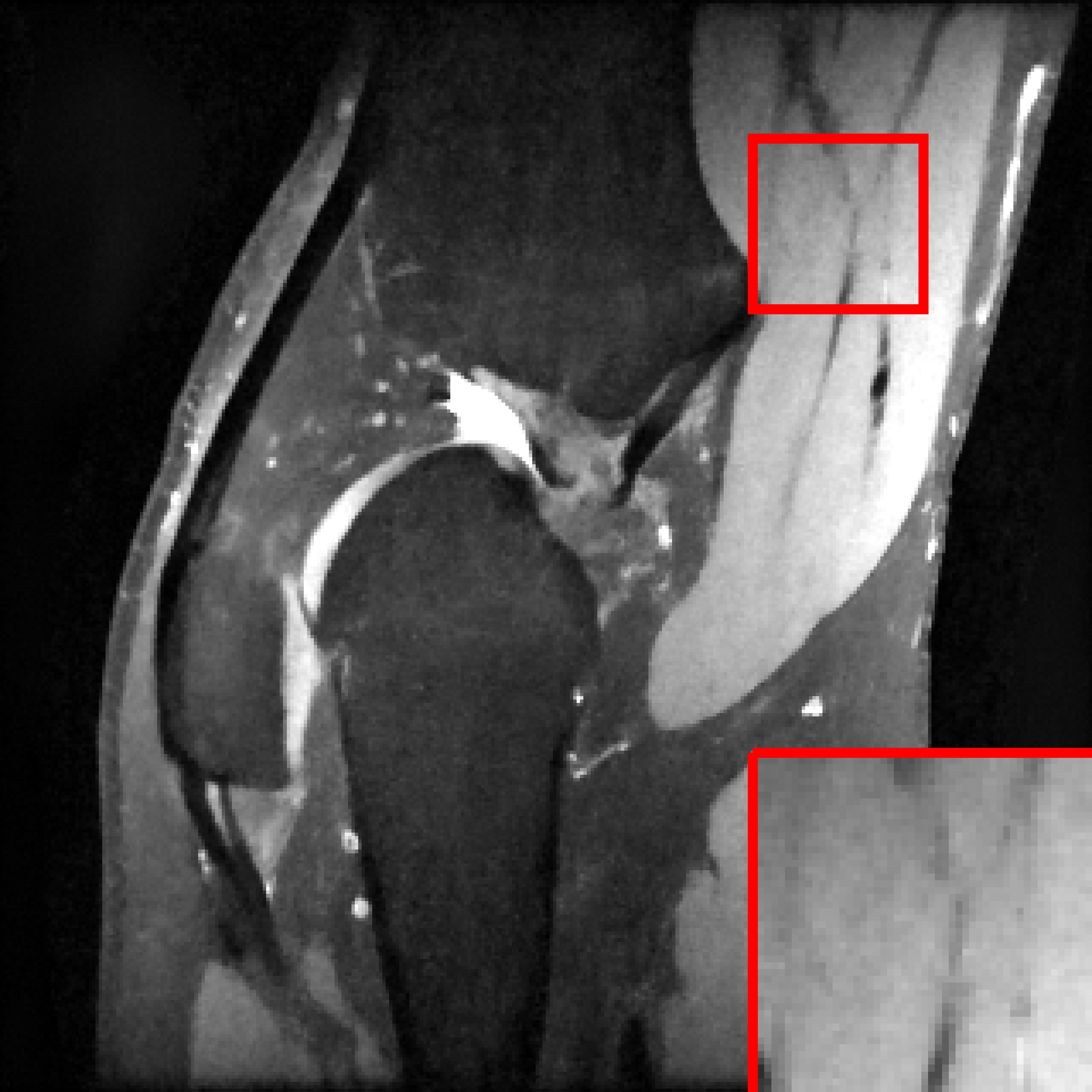}}}};
    
    \node [below=-0.2cm, at=(img_12x_same.south)](img_48x_same) {\reflectbox{\rotatebox[origin=c]{180}{\includegraphics[width=\pngwidth]{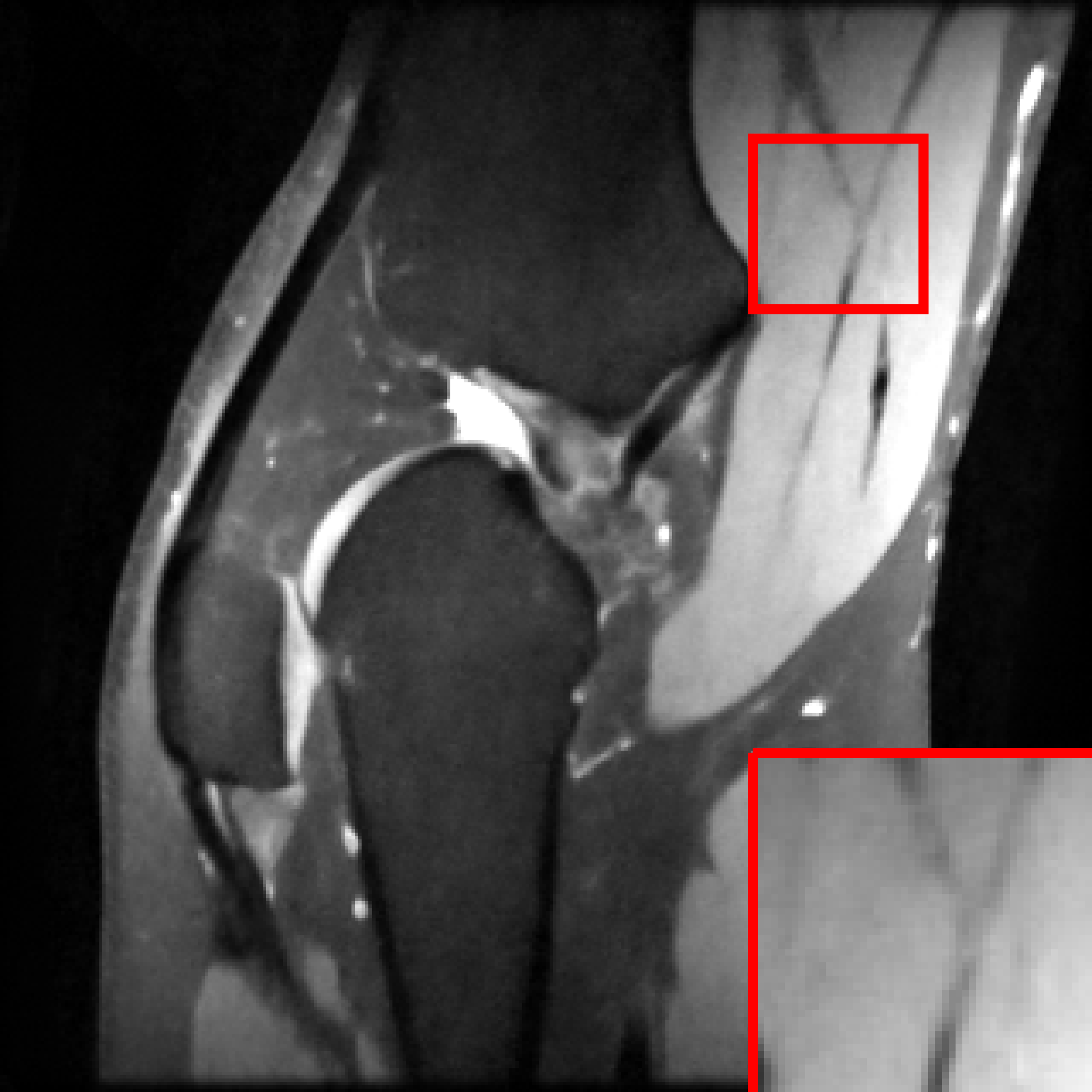}}}};
    \node [below=-0.2cm, at=(img_12x_2x.south)](img_48x_2x) {\reflectbox{\rotatebox[origin=c]{180}{\includegraphics[width=\pngwidth]{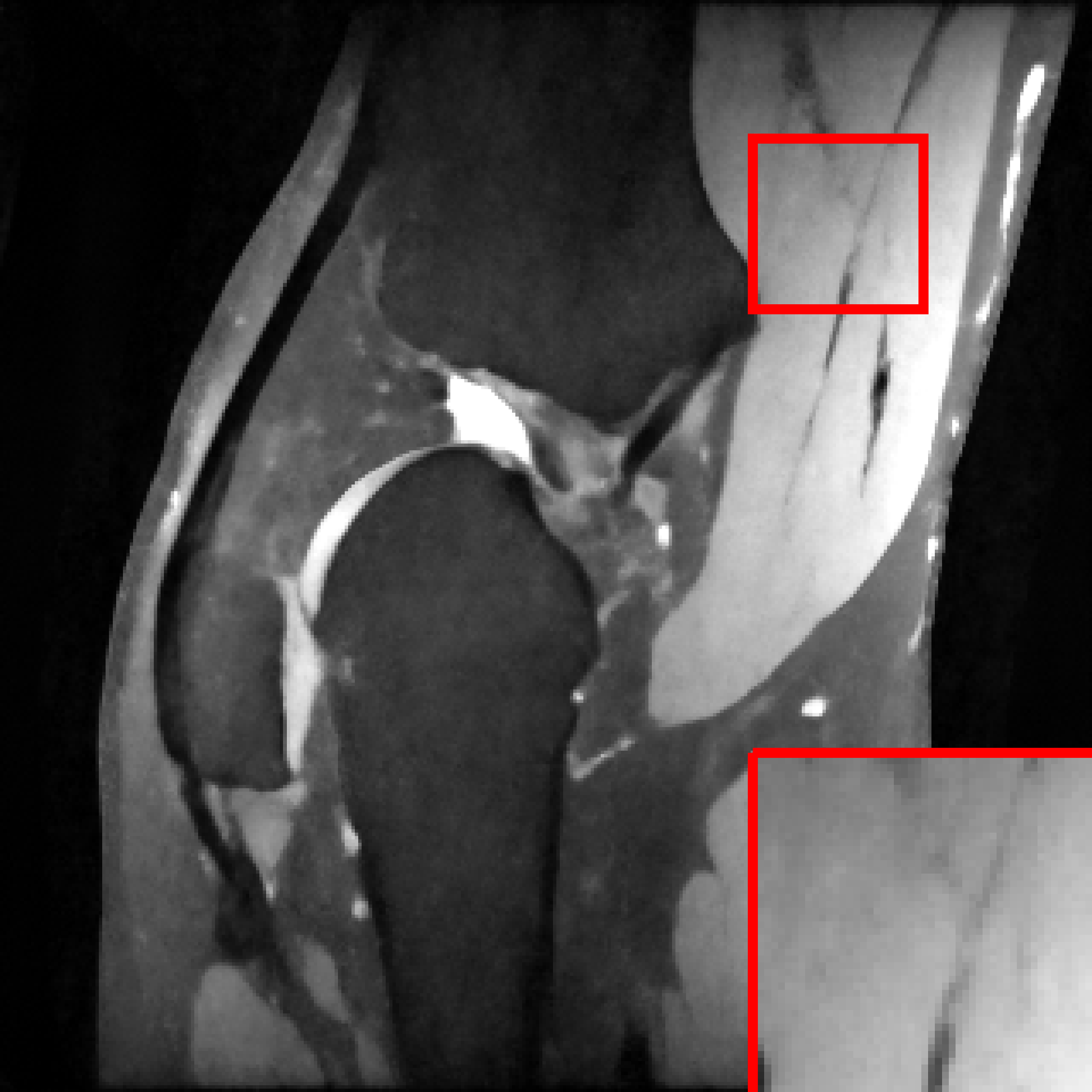}}}};
    \node [below=-0.2cm, at=(img_12x_4x.south)](img_48x_4x) {\reflectbox{\rotatebox[origin=c]{180}{\includegraphics[width=\pngwidth]{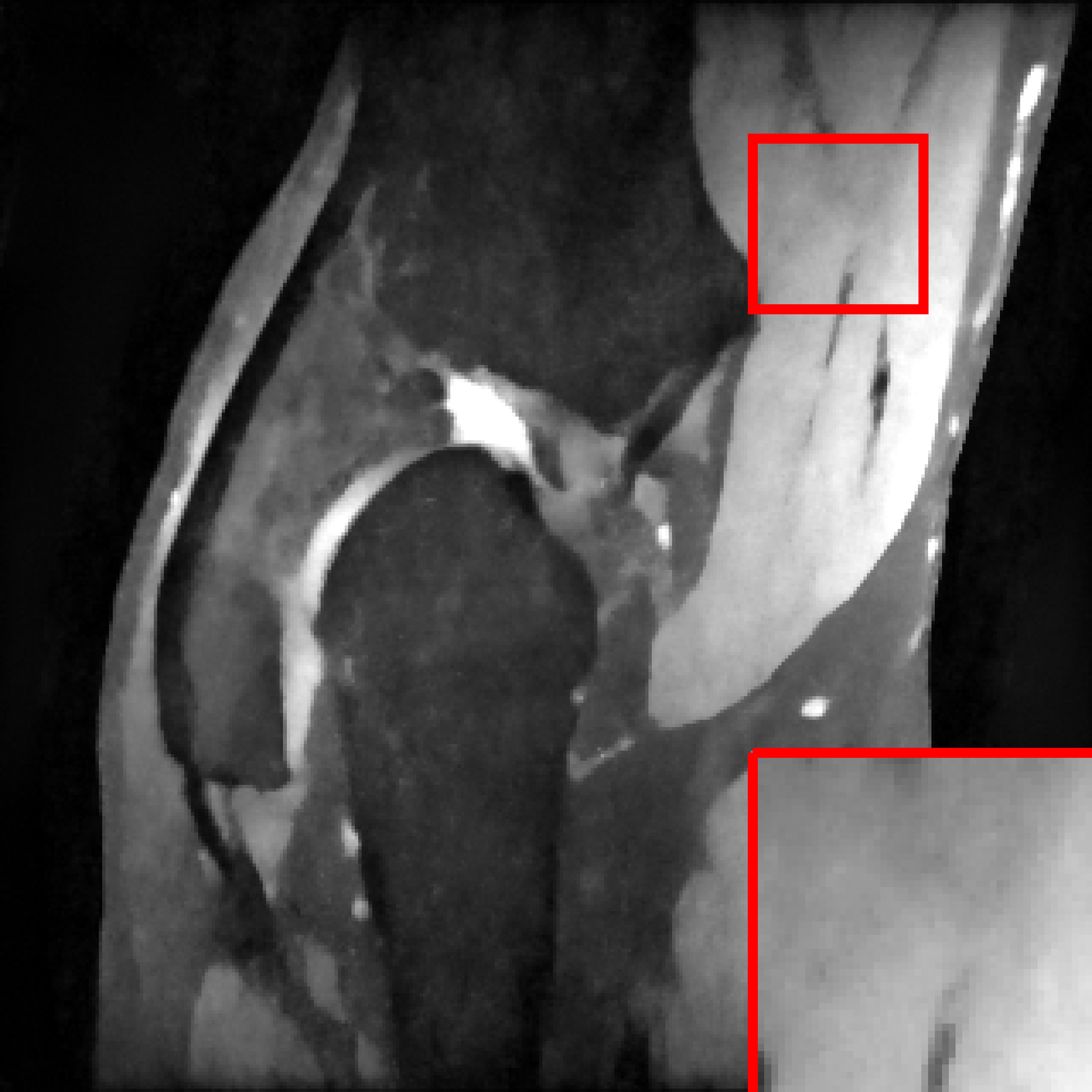}}}};

    \node[above=1pt, at=(img_12x_same.north),font=\color{blue}, align=center] {No shift};
    \node[above=1pt, at=(img_12x_2x.north),font=\color{red}, align=center] {$2 \times$-shift};
    \node[above=1pt, at=(img_12x_4x.north),font=\color{orange}, align=center] {$4 \times$ -shift};

    \node[left=1pt, at=(img_12x_same.west), node distance=0.1cm, rotate=90, anchor=center,yshift=0.1cm,xshift=0.0cm,font=\color{black}] {12x - acc.};
    
    \node[left=1pt, at=(img_48x_same.west), node distance=0.1cm, rotate=90, anchor=center,yshift=0.1cm,xshift=0.0cm,font=\color{black}] {48x - acc.};

    \node[at=(img_48x_4x.east),xshift=0.0cm, yshift=0.00cm]{
        
        \begin{groupplot}[group style={group size=1 by 1},
            width=0.28\columnwidth,
            height=0.28\columnwidth,
            tick align=outside,
            tick pos=left,
            x grid style={darkgray176},
            xtick style={color=black},
            y grid style={darkgray176},
            scaled ticks=false,
            ytick style={color=black},
            legend image code/.code={
                \draw[#1] (0cm,0.0cm) -- (0.4cm,0.0cm); 
            }, 
            legend style={
              fill opacity=0.8,
              font=\small,
              draw opacity=1,
              text opacity=1,
              at={(1.4, 1.0)},
              anchor=north,
              draw=gray}
        ]

        \nextgroupplot[ylabel={PSNR (dB)}, xlabel={acceleration $R$}, xmin=1, xmax=51, ymin=38, ymax=43, title={Quantitative results}, align=center,
            xtick={4, 12, 24, 36, 48},
            xticklabels={4, 12, 24, 36, 48},
            ytick={38, 39, 40, 41, 42, 43},
            yticklabels={38, 39, 40, 41, 42, 43}
        ]
        
            \addplot [draw=blue, fill=blue, forget plot, mark=*, only marks, mark size=1.5]
            table {%
                4 42.316
                8 41.289
                16 40.477
                24 39.946
                36 39.57
                48 39.3
            };
            \addplot [very thick, blue, dashed]
            table {%
                4 42.316
                8 41.289
                16 40.477
                24 39.946
                36 39.57
                48 39.3
            };
            \addlegendentry{No shift}
            
            \addplot [draw=red, fill=red, forget plot, mark=*, only marks, mark size=1.5]
            table {%
                4 42.25
                8 41.222
                16 40.38
                24 39.8222
                36 39.444
                48 39.0
            };
            \addplot [very thick, red, dashed]
            table {%
                4 42.25
                8 41.222
                16 40.38
                24 39.8222
                36 39.444
                48 39.0
            };
            \addlegendentry{$2$x-shift}
            
            \addplot [draw=orange, fill=orange, forget plot, mark=*, only marks, mark size=1.5]
            table {%
                4 41.926
                8 40.829
                16 40.089
                24 39.414
                36 38.972
                48 38.5
            };
            \addplot [very thick, orange, dashed]
            table {%
                4 41.926
                8 40.829
                16 40.089
                24 39.414
                36 38.972
                48 38.5
            };
            \addlegendentry{$4$x-shift}
            
            \addplot [draw=purple, fill=purple, forget plot, mark=*, only marks, mark size=1.5]
            table {%
                4 42.202
                8 41.181
                16 40.502
                24 39.921
                36 39.54
                48 39.21
            };
            \addplot [very thick, purple]
            table {%
                4 42.202
                8 41.181
                16 40.502
                24 39.921
                36 39.54
                48 39.21
            };
            \addlegendentry{Diverse}
    
        \end{groupplot}
    };

\end{tikzpicture}

%% file: figures_onecolumn_appendix/st3_shift_pngs__sensitivity_ps.tikz
\begin{tikzpicture}

    \definecolor{green}{RGB}{0,128,0}
    \definecolor{orange}{RGB}{255,165,0}
    \definecolor{purple}{RGB}{128,0,128}

    \newcommand{\pngwidth}{0.13\textwidth}

    \node (img_12x_same) {\reflectbox{\rotatebox[origin=c]{180}{\includegraphics[width=\pngwidth]{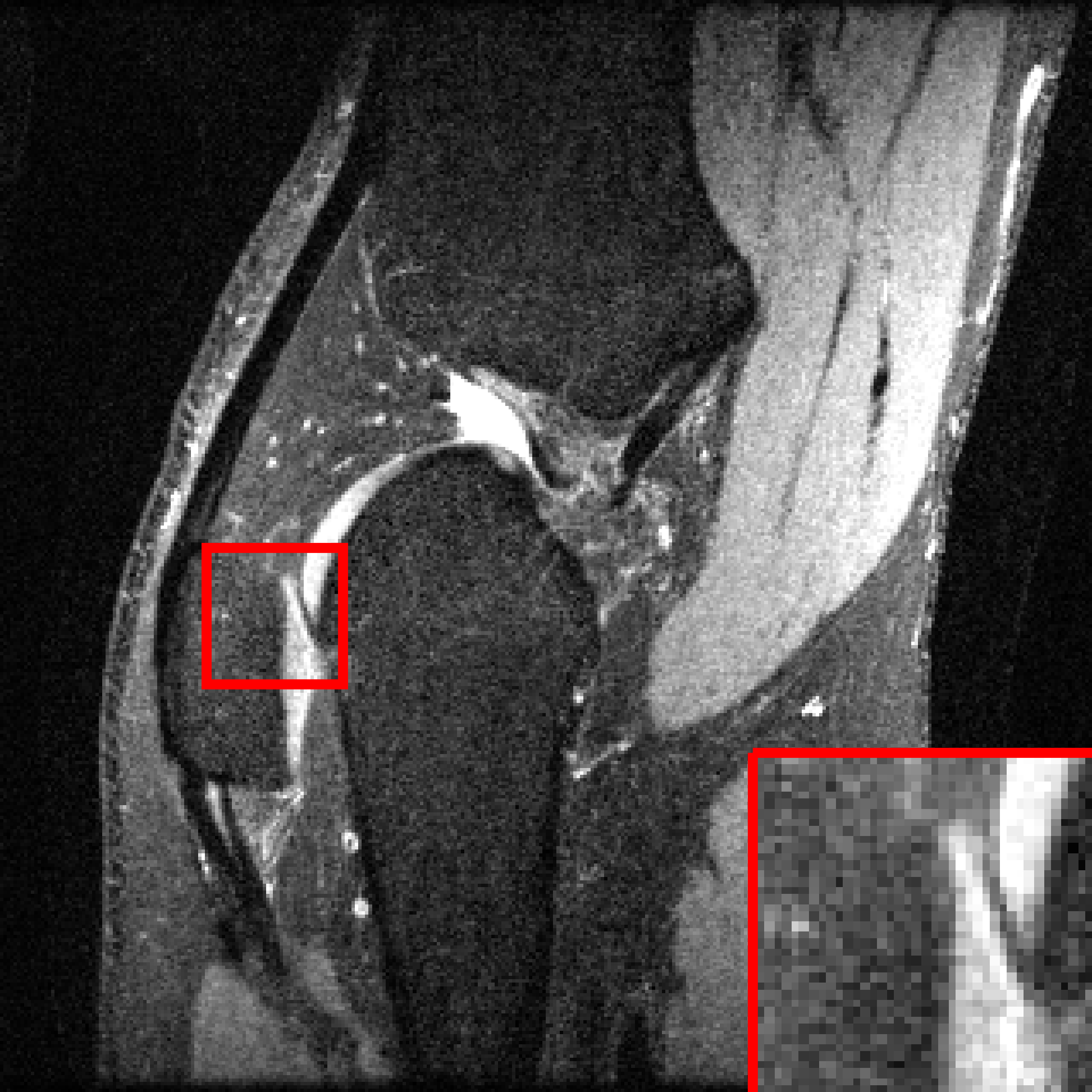}}}};
    \node [right=-0.2cm, at=(img_12x_same.east)](img_12x_2x) {\reflectbox{\rotatebox[origin=c]{180}{\includegraphics[width=\pngwidth]{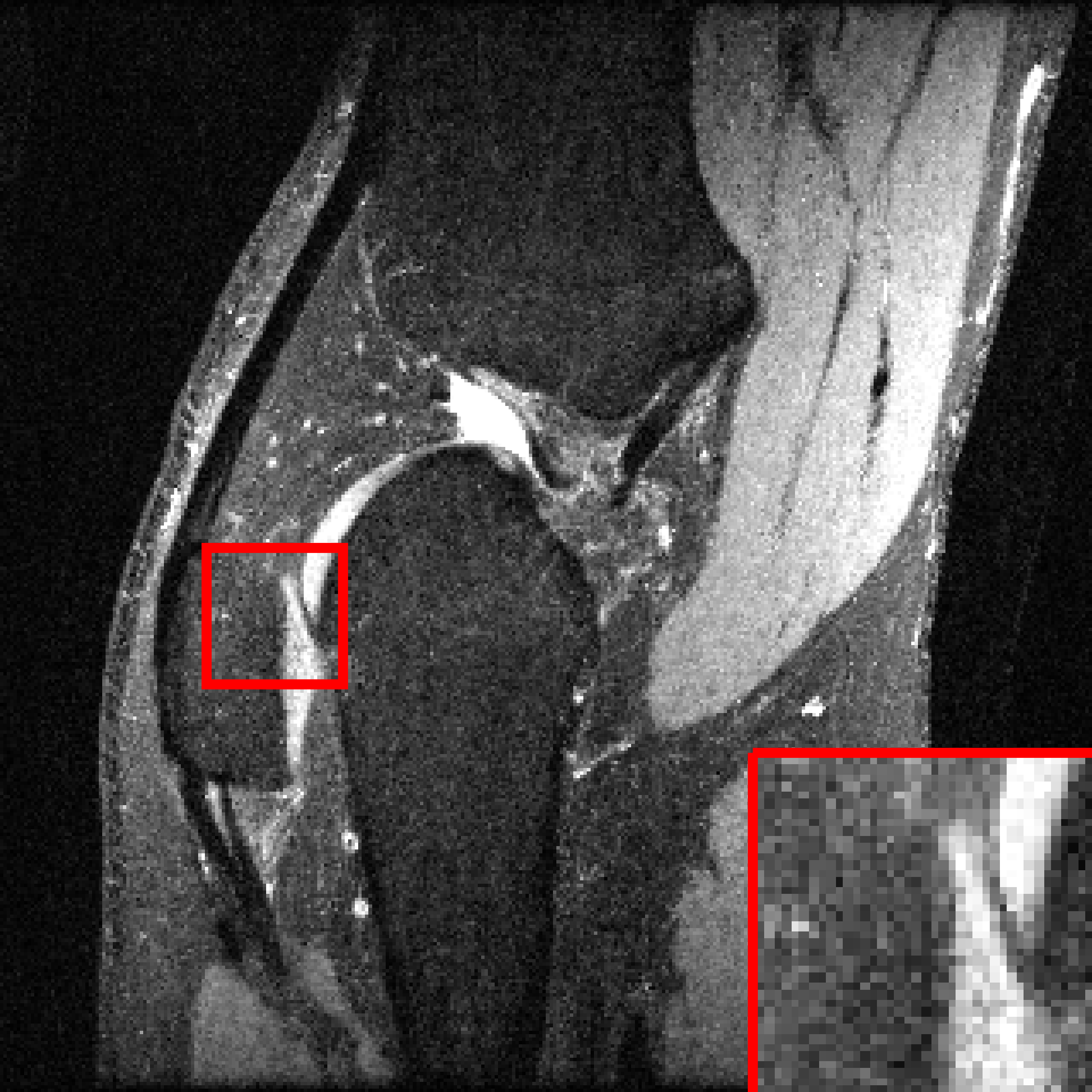}}}};
    \node [right=-0.2cm, at=(img_12x_2x.east)](img_12x_4x) {\reflectbox{\rotatebox[origin=c]{180}{\includegraphics[width=\pngwidth]{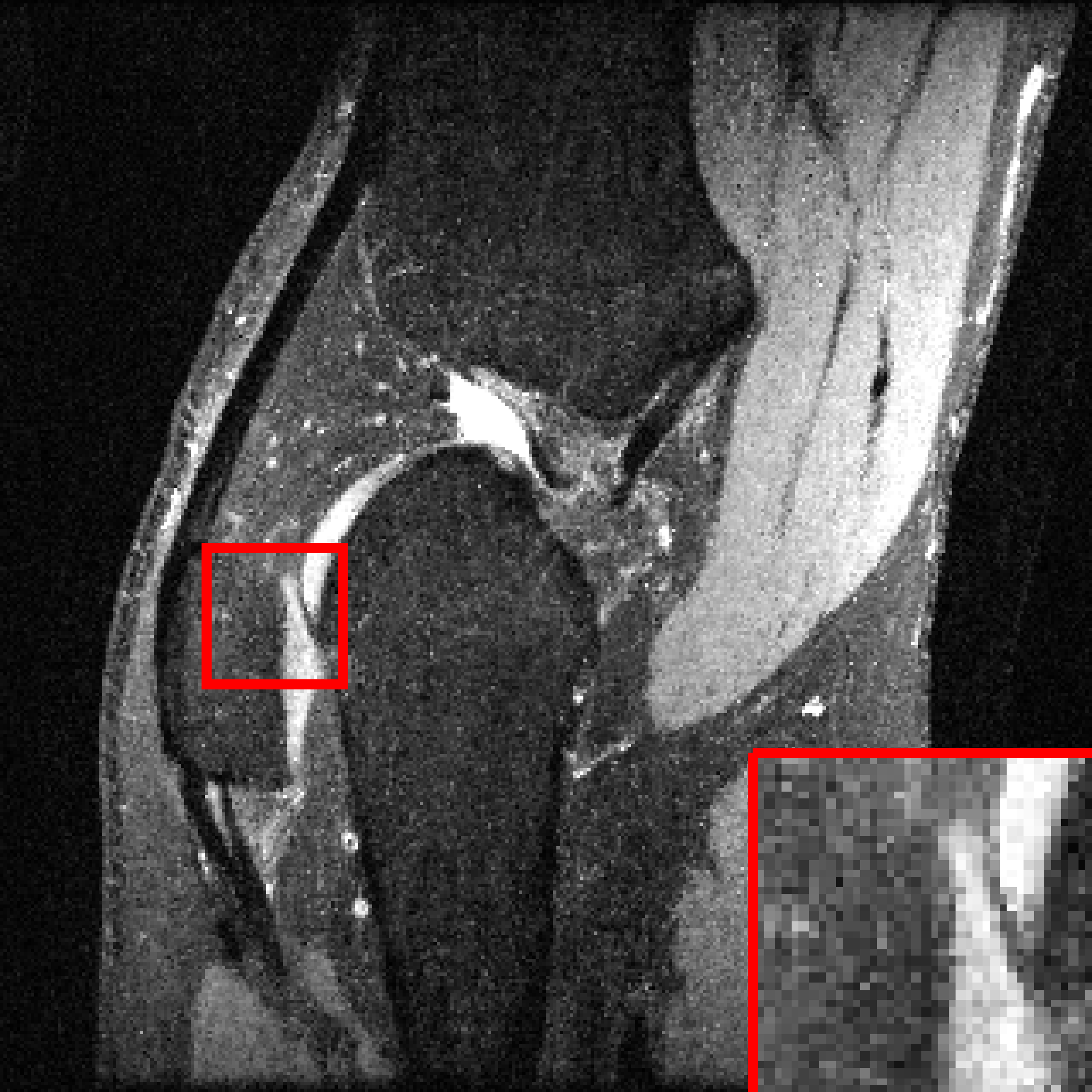}}}};
    
    \node [below=-0.2cm, at=(img_12x_same.south)](img_48x_same) {\reflectbox{\rotatebox[origin=c]{180}{\includegraphics[width=\pngwidth]{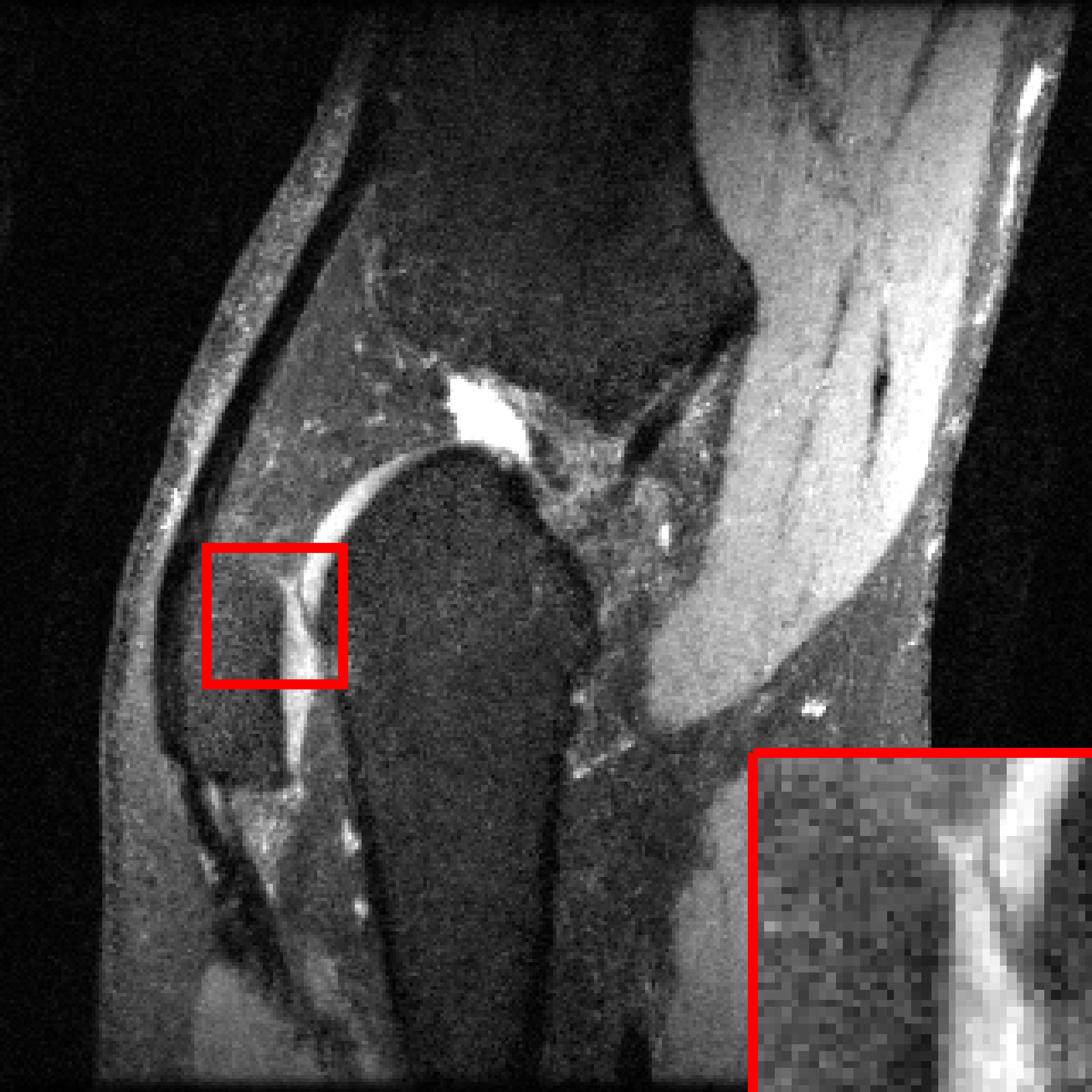}}}};
    \node [below=-0.2cm, at=(img_12x_2x.south)](img_48x_2x) {\reflectbox{\rotatebox[origin=c]{180}{\includegraphics[width=\pngwidth]{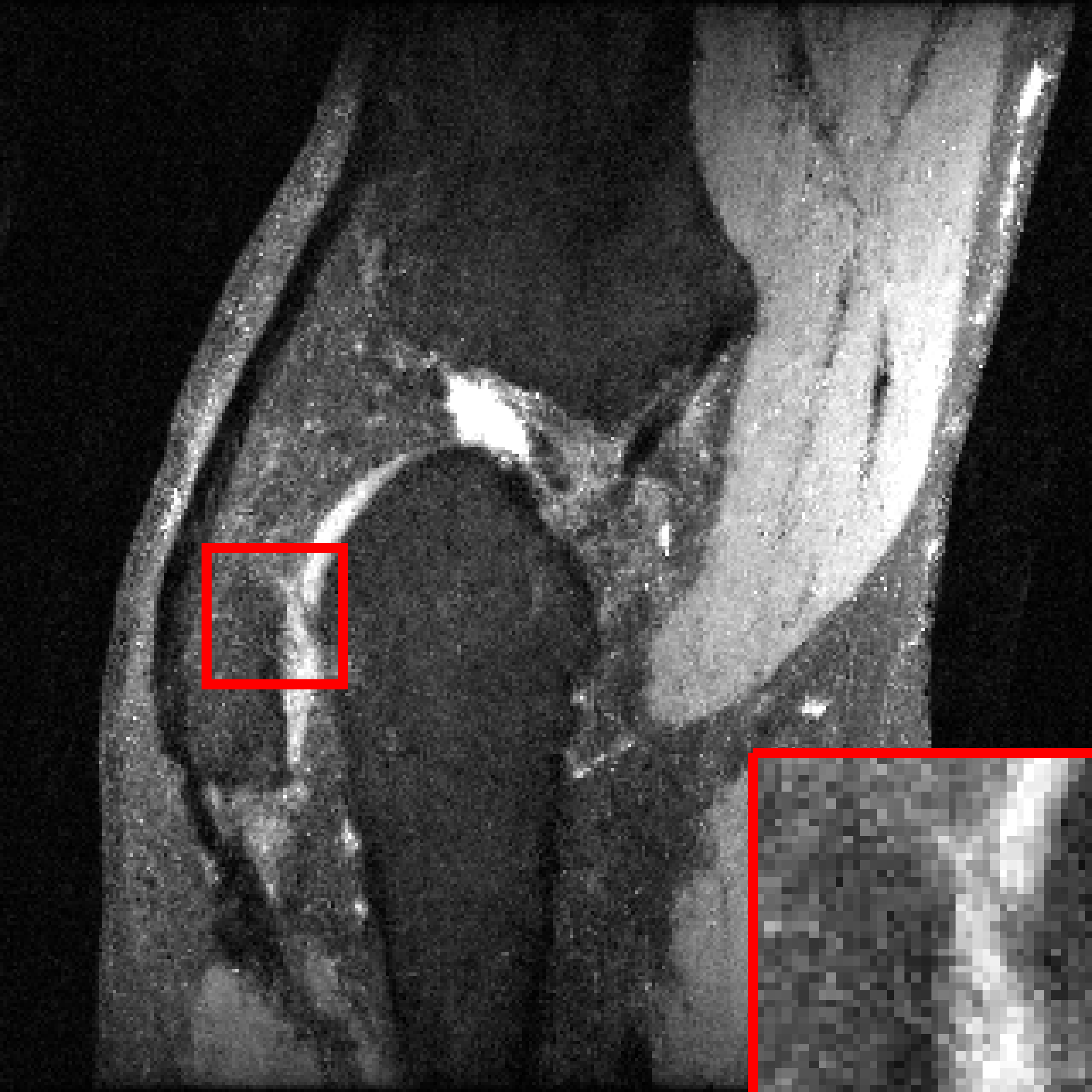}}}};
    \node [below=-0.2cm, at=(img_12x_4x.south)](img_48x_4x) {\reflectbox{\rotatebox[origin=c]{180}{\includegraphics[width=\pngwidth]{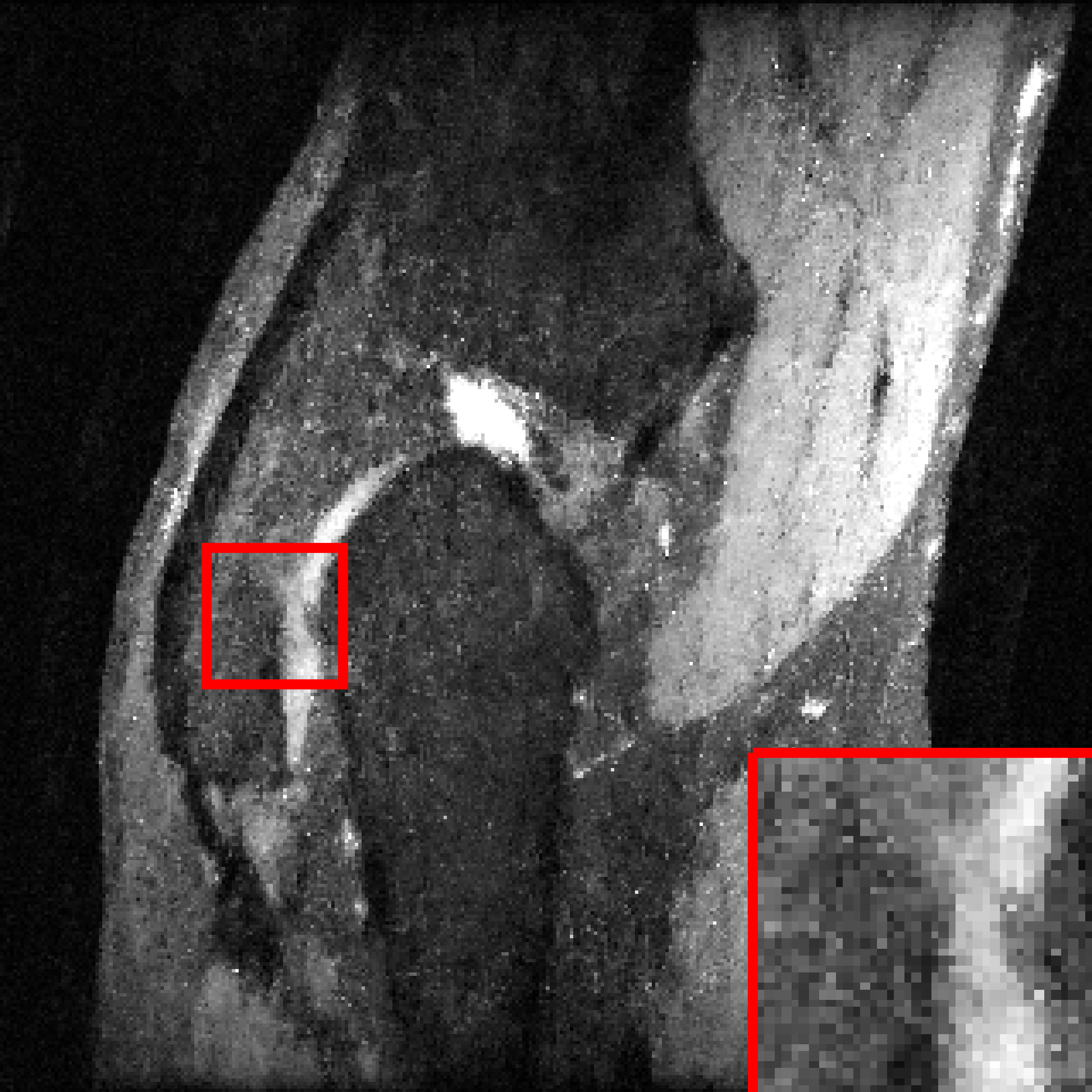}}}};

    \node[above=1pt, at=(img_12x_same.north),font=\color{blue}, align=center] {No shift};
    \node[above=1pt, at=(img_12x_2x.north),font=\color{red}, align=center] {$2 \times$-shift};
    \node[above=1pt, at=(img_12x_4x.north),font=\color{orange}, align=center] {$4 \times$ -shift};

    \node[left=1pt, at=(img_12x_same.west), node distance=0.1cm, rotate=90, anchor=center,yshift=0.1cm,xshift=0.0cm,font=\color{black}] {12x - acc.};
    
    \node[left=1pt, at=(img_48x_same.west), node distance=0.1cm, rotate=90, anchor=center,yshift=0.1cm,xshift=0.0cm,font=\color{black}] {48x - acc.};
 
    \node[at=(img_48x_4x.east), xshift=0.5cm, yshift=0.00cm]{
        
        \begin{groupplot}[group style={group size=1 by 1},
            width=0.28\columnwidth,
            height=0.28\columnwidth,
            tick align=outside,
            tick pos=left,
            x grid style={darkgray176},
            xtick style={color=black},
            y grid style={darkgray176},
            scaled ticks=false,
            ytick style={color=black},
            legend image code/.code={
                \draw[#1] (0cm,0.0cm) -- (0.4cm,0.0cm); 
            }, 
            legend style={
              fill opacity=0.8,
              font=\small,
              draw opacity=1,
              text opacity=1,
              at={(1.4, 1.0)},
              anchor=north,
              draw=gray}
        ]

        \nextgroupplot[ylabel={PSNR (dB)}, xlabel={acceleration $R$}, xmin=1, xmax=51, ymin=35.5, ymax=39.5, title={Quantitative results}, align=center,
            xtick={4, 12, 24, 36, 48},
            xticklabels={4, 12, 24, 36, 48},
            ytick={36, 37, 38, 39},
            yticklabels={36, 37, 38, 39}
        ]
        
            \addplot [draw=blue, fill=blue, forget plot, mark=*, only marks, mark size=1.5]
            table {%
                4 39.097
                8 38.288
                12 38.086
                16 37.894
                24 37.696
                36 37.524
                48 37.458
            };
            \addplot [very thick, blue, dashed]
            table {%
                4 39.097
                8 38.288
                12 38.086
                16 37.894
                24 37.696
                36 37.524
                48 37.458
            };
            \addlegendentry{No shift}
            
            \addplot [draw=red, fill=red, forget plot, mark=*, only marks, mark size=1.5]
            table {%
                4 39.105
                8 38.023
                12 37.664
                16 37.351
                24 37.022
                36 36.759
                48 36.629
            };
            \addplot [very thick, red, dashed]
            table {%
                4 39.105
                8 38.023
                12 37.664
                16 37.351
                24 37.022
                36 36.759
                48 36.629
            };
            \addlegendentry{$2$x-shift}
            
            \addplot [draw=orange, fill=orange, forget plot, mark=*, only marks, mark size=1.5]
            table {%
                4 39.115
                8 37.983
                12 37.568
                16 37.261
                24 36.842
                36 36.465
                48 36.257
            };
            \addplot [very thick, orange, dashed]
            table {%
                4 39.115
                8 37.983
                12 37.568
                16 37.261
                24 36.842
                36 36.465
                48 36.257
            };
            \addlegendentry{$4$x-shift}
    
        \end{groupplot}
    };

\end{tikzpicture}

%% file: figures_onecolumn_appendix/ssim__comp__sensitivities.tikz
\begin{tikzpicture}

    \definecolor{green}{RGB}{0,128,0}
    \definecolor{orange}{RGB}{255,165,0}
    \definecolor{purple}{RGB}{128,0,128}

    \begin{groupplot}[group style={group size=2 by 1,
            horizontal sep=20pt},
            width  = 0.28\columnwidth,
            height = 0.28\columnwidth,
            tick align=outside,
            tick pos=left,
            x grid style={darkgray176},
            xtick style={color=black},
            y grid style={darkgray176},
            scaled ticks=false,
            ylabel={SSIM (sagittal)},
            xlabel={acceleration $R$},
            ytick style={color=black},
            xmin=1, xmax=51,
            ymin=0.82, ymax=0.96,
            ytick={0.83, 0.86, 0.89, 0.92, 0.95},
            yticklabels={0.83, 0.86, 0.89, 0.92, 0.95},
            xtick={4, 12, 24, 36, 48},
            xticklabels={4, 12, 24, 36, 48},
            legend image code/.code={
                \draw[#1] (0cm,0.0cm) -- (0.3cm,0.0cm); 
            }, 
            legend style={
              legend columns=1,
              fill opacity=0.8,
              font=\small,
              draw opacity=1,
              text opacity=1,
              at={(1.5, 1.0)},
              anchor=north,
              draw=gray}
        ]

        \nextgroupplot[title={Variational method}
        ]
            \addplot [draw=blue, fill=blue, forget plot, mark=*, only marks, mark size=1.5]
            table {%
                4 0.943
                8 0.9231
                12 0.911
                16 0.9019
                24 0.892
                36 0.8808
                48 0.8713
            };
            \addplot [very thick, blue]
            table {%
                4 0.943
                8 0.9231
                12 0.911
                16 0.9019
                24 0.892
                36 0.8808
                48 0.8713
            };

            \addplot [draw=red, fill=red, forget plot, mark=*, only marks, mark size=1.5]
            table {%
                4 0.9407
                8 0.9201
                12 0.9075
                16 0.8977
                24 0.8877
                36 0.8763
                48 0.8668
            };
            \addplot [very thick, red, dashed]
            table {%
                4 0.9407
                8 0.9201
                12 0.9075
                16 0.8977
                24 0.8877
                36 0.8763
                48 0.8668
            };

            \addplot [draw=orange, fill=orange, forget plot, mark=*, only marks, mark size=1.5]
            table {%
                4 0.9408
                8 0.9198
                12 0.9067
                16 0.897
                24 0.8859
                36 0.8743
                48 0.864
            };
            \addplot [very thick, green, dotted]
            table {%
                4 0.9408
                8 0.9198
                12 0.9067
                16 0.897
                24 0.8859
                36 0.8743
                48 0.864
            };
        
        \nextgroupplot[ylabel={}, yticklabels={}, title={Posterior sampling}, title style={yshift=-0.08cm}]
        
            \addplot [draw=blue, fill=blue, forget plot, mark=*, only marks, mark size=1.5]
            table {%
                4 0.9058
                8 0.8907
                12 0.8825
                16 0.8745
                24 0.8684
                36 0.8611
                48 0.8545
            };
            \addplot [very thick, blue]
            table {%
                4 0.9058
                8 0.8907
                12 0.8825
                16 0.8745
                24 0.8684
                36 0.8611
                48 0.8545
            };
            \addlegendentry{No shift}

            \addplot [draw=red, fill=red, forget plot, mark=*, only marks, mark size=1.5]
            table {%
                4 0.9009
                8 0.8837
                12 0.8739
                16 0.864
                24 0.8562
                36 0.8469
                48 0.8388
            };
            \addplot [very thick, red, dashed]
            table {%
                4 0.9009
                8 0.8837
                12 0.8739
                16 0.864
                24 0.8562
                36 0.8469
                48 0.8388
            };
            \addlegendentry{$2 \times$-shift}

            \addplot [draw=orange, fill=orange, forget plot, mark=*, only marks, mark size=1.5]
            table {%
                4 0.8969
                8 0.8778
                12 0.8669
                16 0.856
                24 0.8462
                36 0.837
                48 0.8297
            };
            \addplot [very thick, orange, dotted]
            table {%
                4 0.8969
                8 0.8778
                12 0.8669
                16 0.856
                24 0.8462
                36 0.837
                48 0.8297
            };
            \addlegendentry{$4 \times$-shift}
    
    \end{groupplot}

\end{tikzpicture}

%% file: figures_onecolumn_appendix/inn_repr_width.tikz
\begin{tikzpicture}
    \begin{axis}[
        title={},
        width=0.32\columnwidth,
        height=0.32\columnwidth,
        ymin=28, ymax=48,
        xmin=80, xmax=520,
        title={Approximation error},
        xlabel={Layer width},
        ylabel={PSNR (dB)},
        grid=both, 
        grid style={dashed, gray!30},
        scatter/classes={
            c5={mark=square*,blue}, 
            c6={mark=triangle*,red}, 
            c7={mark=o,green!50!black} 
        },
        legend style={
            font=\small,
            at={(1.6,1)},
            anchor=north,
            legend columns=1
        }, 
        scaled x ticks=false
    ]
        \addplot[scatter,only marks,scatter src=explicit symbolic]
            coordinates {
                (150, 31.273313522338867) [c5]
                (100, 31.704458236694336) [c5]
                (200, 34.1119384765625) [c5]
                (250, 31.92991065979004) [c5]
                (300, 37.84635925292969) [c5]
                (350, 38.28199005126953) [c5]
                (400, 37.195579528808594) [c5]
                (500, 44.97571563720703) [c5]
                (450, 42.647666931152344) [c5]
                (100, 32.50981140136719) [c6]
                (100, 29.996049880981445) [c7]
                (150, 29.3846378326416) [c6]
                (400, 40.89384460449219) [c6]
                (250, 34.17762756347656) [c6]
                (400, 39.03569793701172) [c7]
                (250, 35.347755432128906) [c7]
                (200, 33.93609619140625) [c6]
                (500, 31.076276779174805) [c7]
                (350, 38.29784393310547) [c7]
                (200, 33.3294677734375) [c7]
                (450, 37.842742919921875) [c7]
                (300, 35.683998107910156) [c7]
                (150, 31.323606491088867) [c7]
                (500, 44.071632385253906) [c6]
                (350, 39.59910583496094) [c6]
                (450, 43.73741149902344) [c6]
                (300, 35.29485321044922) [c6]
                (400, 38.7154541015625) [c5]
                (500, 46.198036193847656) [c5]
                (450, 44.03202819824219) [c5]
            };
        \legend{$5$ hidden layers, $6$ hidden layers, $7$ hidden layers}

        \addplot[dotted, thick, black] coordinates {(0, 42.3) (520, 42.3)};

        \addlegendentry{Voxel rep.}
    \end{axis}
    
\end{tikzpicture}

%% file: figures_onecolumn_appendix/gaussian_vs_voxelrep.tikz
\begin{tikzpicture}
    \newcommand{\pngwidth}{0.2\textwidth}
    \node [right=0.5cm](img_target) {\includegraphics[width=\pngwidth]{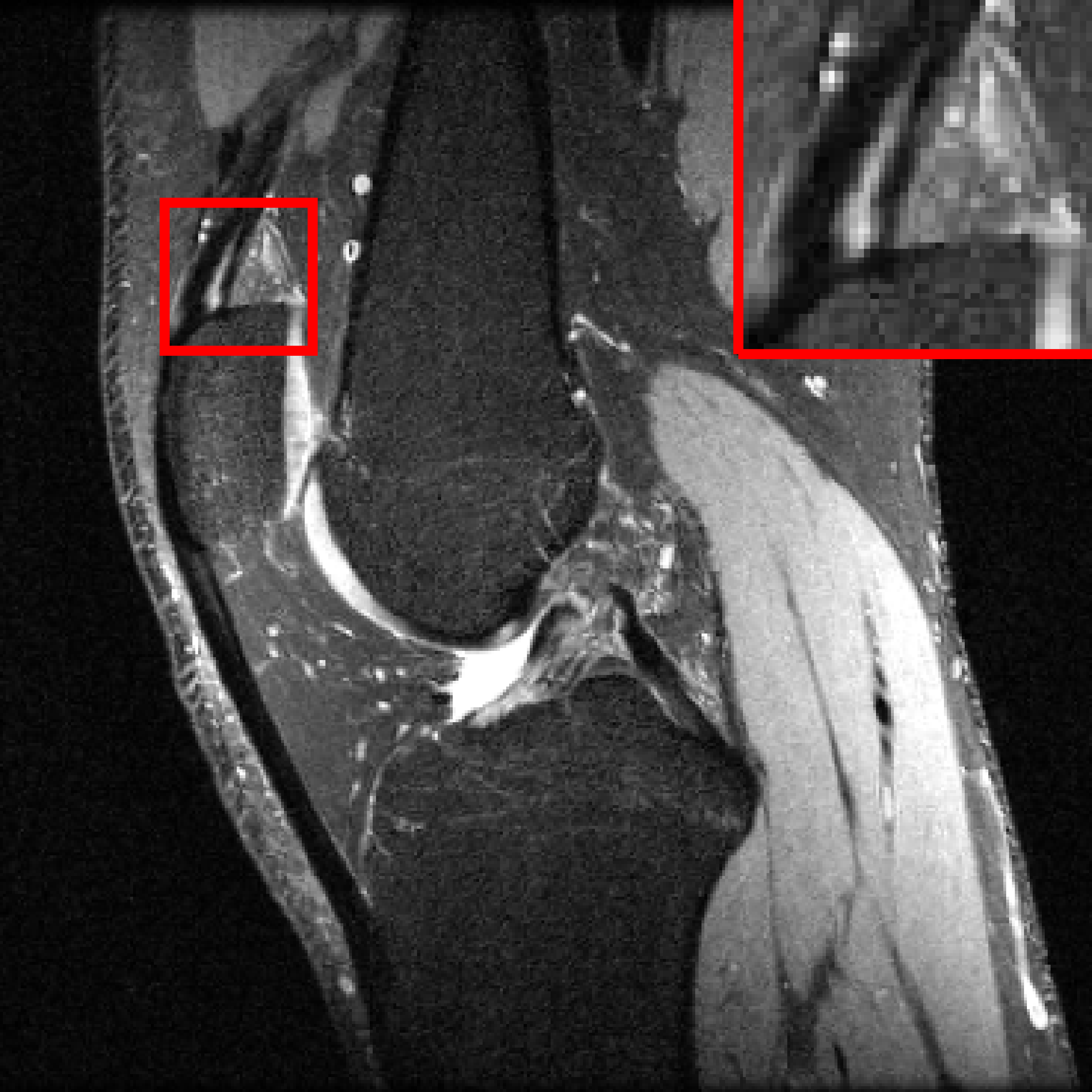}};
    \node [right=0.5cm, at=(img_target.east)](img_diffusion) {\includegraphics[width=\pngwidth]{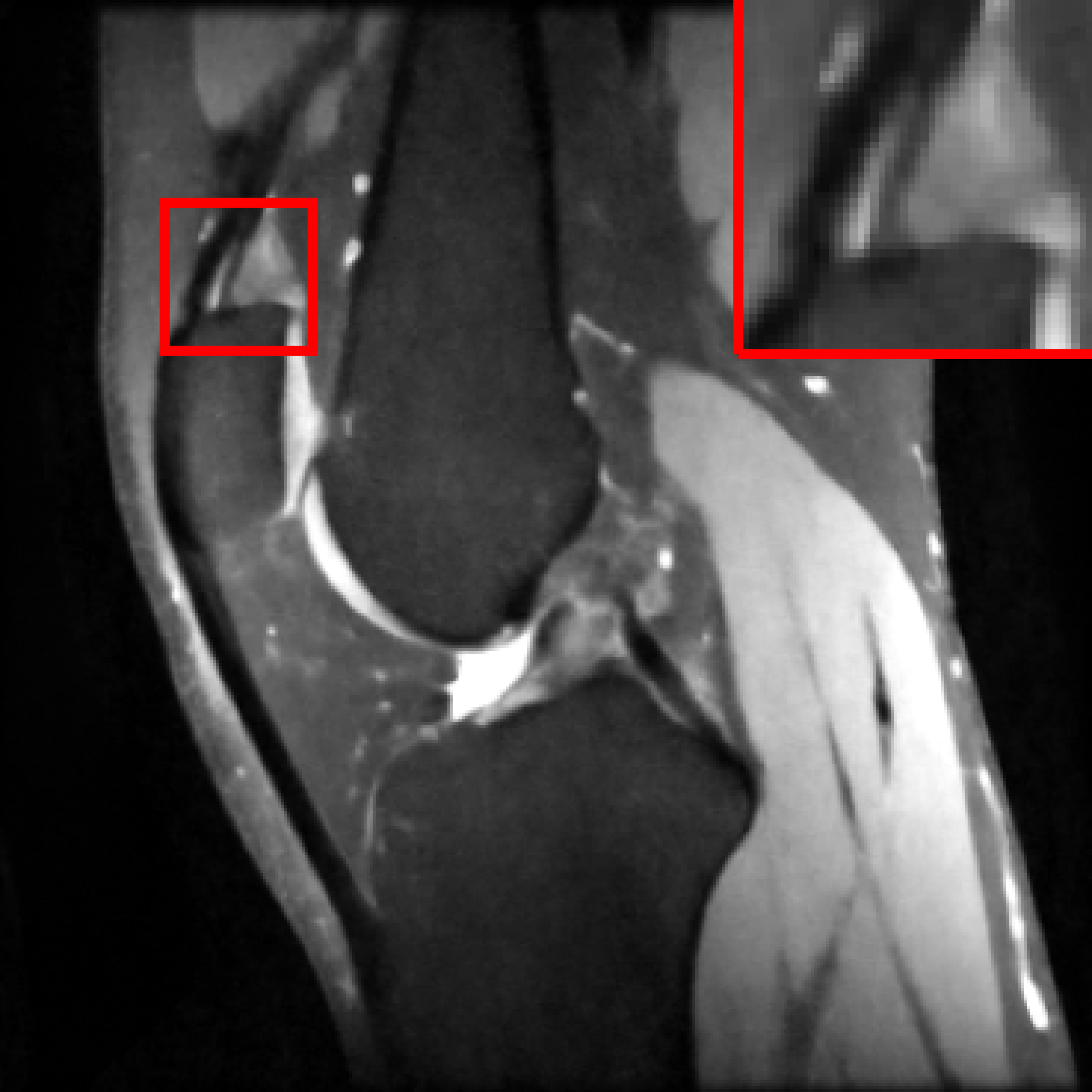}};
    \node [right=0.5cm, at=(img_diffusion.east)](img_diffusion_gauss) {\includegraphics[width=\pngwidth]{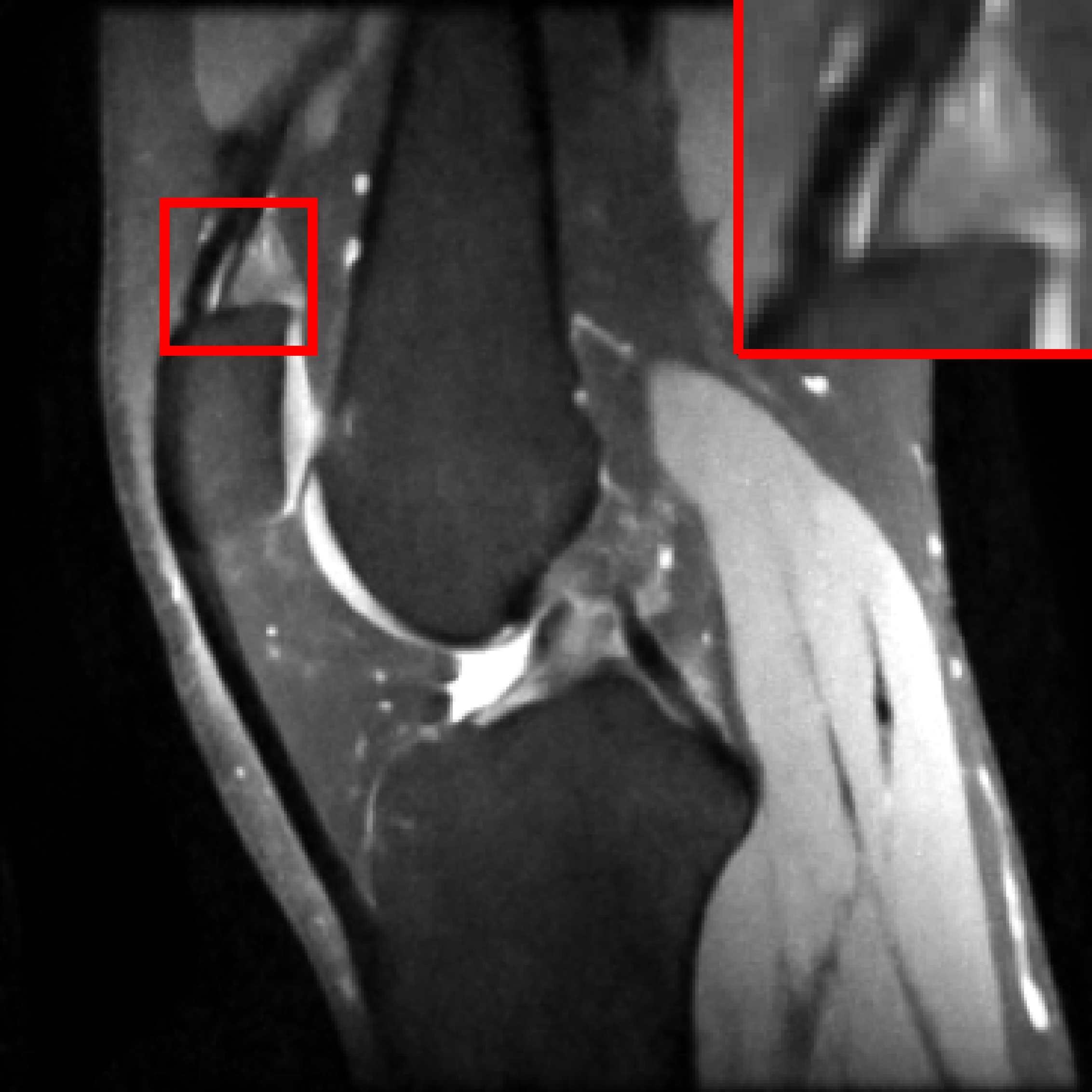}};
    \node[above=1pt, at=(img_target.north), font=\color{black}, align=center] {Target};
    \node[above=1pt, at=(img_diffusion.north),font=\color{black}, align=center] {Voxel-based \\ PSNR: $39.21$dB};
    \node[above=1pt, at=(img_diffusion_gauss.north),font=\color{black}, align=center] {Gaussian splatting \\PSNR: $39.31$dB};
\end{tikzpicture}

%% file: figures_onecolumn_appendix/kernel_interpolations_zero_pad.tikz
\begin{tikzpicture}

    \definecolor{green}{RGB}{0,128,0}
    \definecolor{orange}{RGB}{255,165,0}
    \definecolor{purple}{RGB}{128,0,128}

    \begin{groupplot}[group style={group size=2 by 1,
            horizontal sep=15pt},
            width  = 0.28\columnwidth,
            height = 0.28\columnwidth,
            tick align=outside,
            tick pos=left,
            x grid style={darkgray176},
            xtick style={color=black},
            y grid style={darkgray176},
            scaled ticks=false,
            ylabel={SSIM (sagittal)},
            xlabel={acceleration $R$},
            ytick style={color=black},
            xmin=1, xmax=51,
            ymin=34, ymax=43,
            ytick={35, 37, 39, 41, 43},
            yticklabels={35,37,39,41,43},
            legend image code/.code={
                \draw[#1] (0cm,0.0cm) -- (0.3cm,0.0cm); 
            }, 
            legend style={
              legend columns=1,
              fill opacity=0.8,
              font=\small,
              draw opacity=1,
              text opacity=1,
              at={(1.5, 1.0)},
              anchor=north,
              draw=gray},
        ]

        \nextgroupplot[ylabel={PSNR (dB)}, xlabel={acceleration $R$}, xmin=1, xmax=51, title={$\Vrecon = 0.5$mm},
            xtick={4, 12, 24, 36, 48},
            xticklabels={4, 12, 24, 36, 48}
        ]

            \addplot [draw=purple, fill=purple, forget plot, mark=*, only marks, mark size=1.5]
            table {%
            4 41.892
            8 41.026
            12 40.593
            16 40.286
            24 39.891
            36 39.508
            48 39.314
            };
            \addplot [very thick, purple, dotted]
            table {%
            4 41.892
            8 41.026
            16 40.286
            24 39.891
            36 39.508
            48 39.314
            };

            \addplot [draw=orange, fill=orange, forget plot, mark=*, only marks, mark size=1.5]
            table {%
            4 41.926
            8 40.829
            16 40.089
            24 39.414
            36 38.972
            48 38.5
            };
            \addplot [very thick, orange]
            table {%
            4 41.926
            8 40.829
            16 40.089
            24 39.414
            36 38.972
            48 38.5
            };

            \addplot [draw=blue, fill=blue, forget plot, mark=*, only marks, mark size=1.5]
            table {%
                4 41.7879
                8 40.41799
                12 39.7262
                16 39.2855
                24 38.5595
                36 37.967
                48 37.644
            };
            \addplot [very thick, blue]
            table {%
                4 41.7879
                8 40.41799
                12 39.7262
                16 39.2855
                24 38.5595
                36 37.967
                48 37.644
            };
        
            \addplot [draw=brown, fill=brown, forget plot, mark=*, only marks, mark size=1.5]
            table {%
                4 41.8891
                8 40.5543
                12 39.87
                16 39.45
                24 38.75
                36 38.1560
                48 37.8307
            };
            \addplot [very thick, brown]
            table {%
                4 41.8891
                8 40.5543
                12 39.87
                16 39.45
                24 38.75
                36 38.1560
                48 37.8307
            };

            \addplot [draw=lime, fill=lime, forget plot, mark=*, only marks, mark size=1.5]
            table {%
                4 40.21
                8 38.88
                12 38.15
                16 37.67
                24 37.036
                36 36.28
                48 35.88
            };
            
            \addplot [very thick, lime]
            table {%
                4 40.21
                8 38.88
                12 38.15
                16 37.67
                24 37.036
                36 36.28
                48 35.88
            };
    
        \nextgroupplot[xlabel={acceleration $R$}, xmin=3, xmax=17, title={$\Vrecon = 2$mm}, align=center,
            ylabel={},
            xtick={4, 8, 12, 16},
            xticklabels={4, 8, 12, 16},
            yticklabels={}
        ]

            \addplot [draw=purple, fill=purple, forget plot, mark=*, only marks, mark size=1.5]
            table {%
                4.0000 42.1862
                8.0000 39.9634
                12.0000 38.7077
                16.0000 37.9871
            };
            \addplot [very thick, purple, dotted]
            table {%
                4.0000 42.1862
                8.0000 39.9634
                12.0000 38.7077
                16.0000 37.9871
            };
            \addlegendentry{Baseline}

            \addplot [draw=orange, fill=orange, forget plot, mark=*, only marks, mark size=1.5]
            table {%
                4.0000 41.8807
                8.0000 39.5402
                12.0000 38.2246
                16.0000 37.4802
            };
            \addplot [very thick, orange]
            table {%
                4.0000 41.8807
                8.0000 39.5402
                12.0000 38.2246
                16.0000 37.4802
            };
            \addlegendentry{Fixed}

            \addplot [draw=brown, fill=brown, forget plot, mark=*, only marks, mark size=1.5]
            table {%
                4.0000 41.805
                8.0000 39.08981
                12.0000 37.48902
                16.0000 36.62812
            };
            \addplot [very thick, brown]
            table {%
                4.0000 41.805
                8.0000 39.08981
                12.0000 37.48902
                16.0000 36.62812
            };
            \addlegendentry{Bilinear}

            \addplot [draw=blue, fill=blue, forget plot, mark=*, only marks, mark size=1.5]
            table {%
                4.0000 42.30799
                8.0000 39.68783
                12.0000 38.07757
                16.0000 37.18916
            };
            \addplot [very thick, blue]
            table {%
                4.0000 42.30799
                8.0000 39.68783
                12.0000 38.07757
                16.0000 37.18916
            };
            \addlegendentry{Fourier}

            \addplot [draw=lime, fill=lime, forget plot, mark=*, only marks, mark size=1.5]
            table {%
                4 39.00
                8 35.8644
                12 34.798
                16 34.188
            };
            \addplot [very thick, lime]
            table {%
                4 39.00
                8 35.8644
                12 34.798
                16 34.188
            };
            \addlegendentry{ZeroPad}
    
    \end{groupplot}

\end{tikzpicture}

%% file: figures_onecolumn_appendix/ahead_shift_pngs__sensitivity.tikz
\begin{tikzpicture}

    \definecolor{green}{RGB}{0,128,0}
    \definecolor{orange}{RGB}{255,165,0}
    \definecolor{purple}{RGB}{128,0,128}

    \newcommand{\pngwidth}{0.13\textwidth}

    \node (img_12x_same) {\rotatebox[origin=c]{0}{\includegraphics[width=\pngwidth]{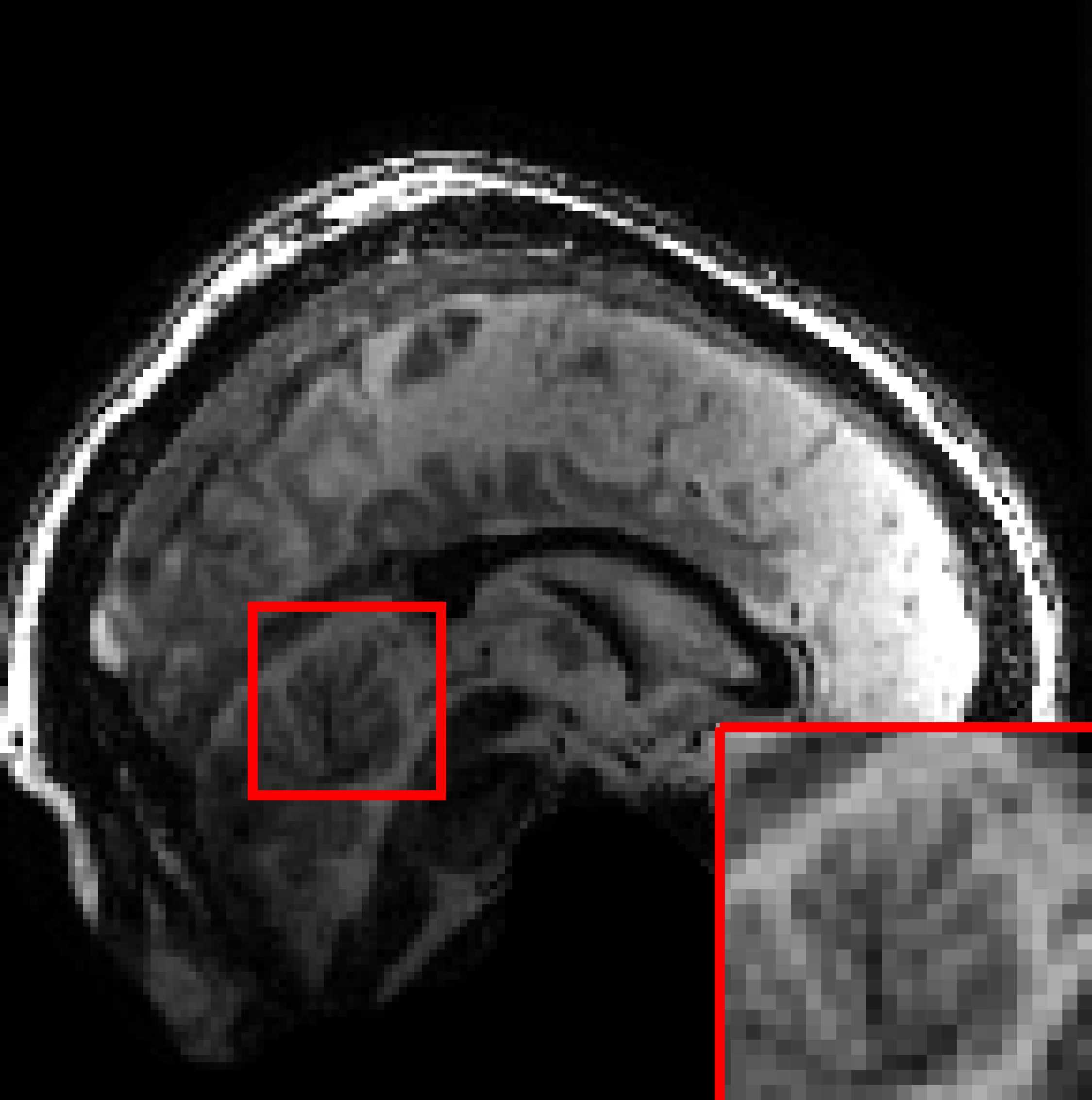}}};
    \node [right=-0.2cm, at=(img_12x_same.east)](img_12x_2x) {\rotatebox[origin=c]{0}{\includegraphics[width=\pngwidth]{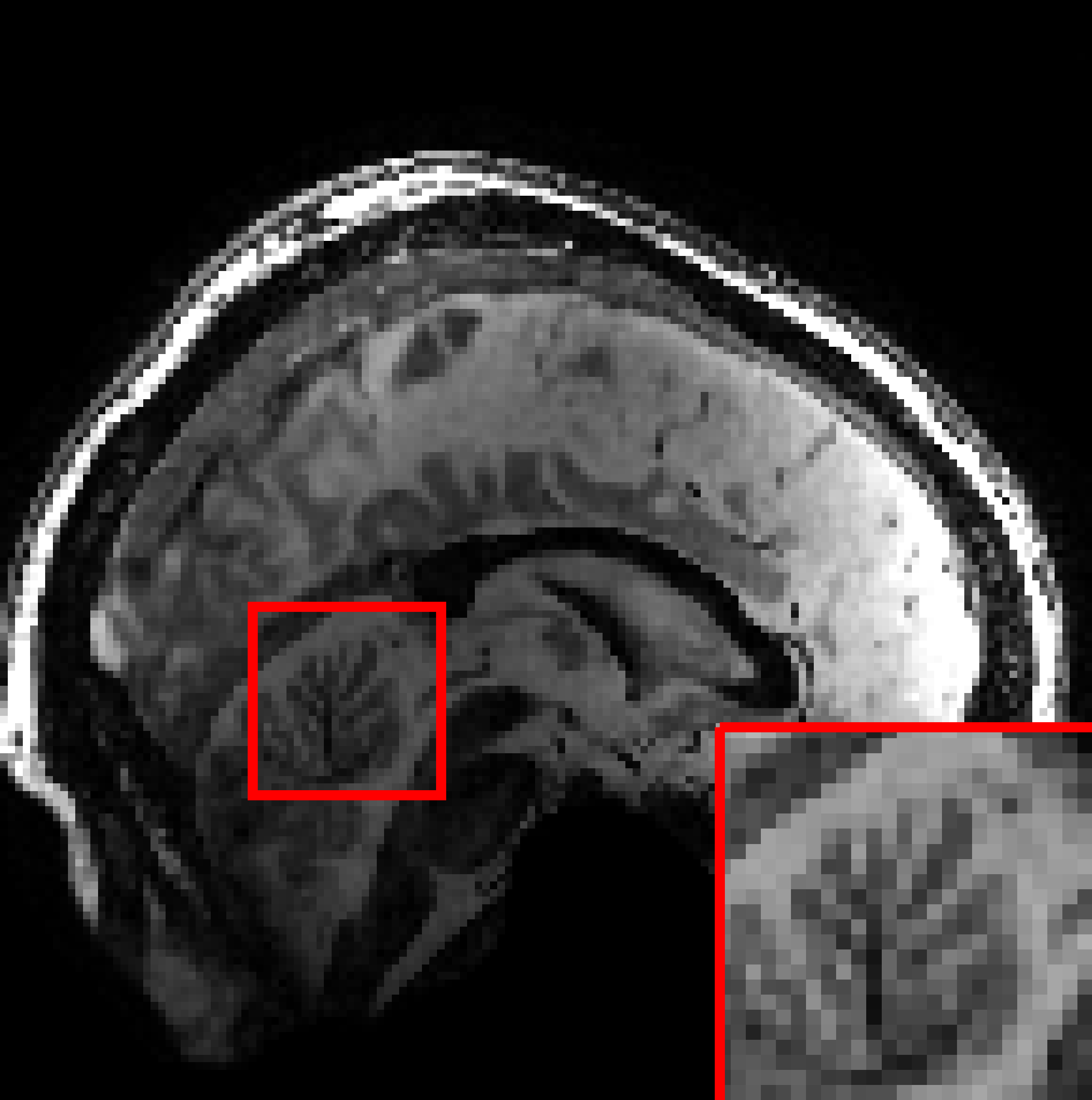}}};
    \node [right=-0.2cm, at=(img_12x_2x.east)](img_12x_4x) {\rotatebox[origin=c]{0}{\includegraphics[width=\pngwidth]{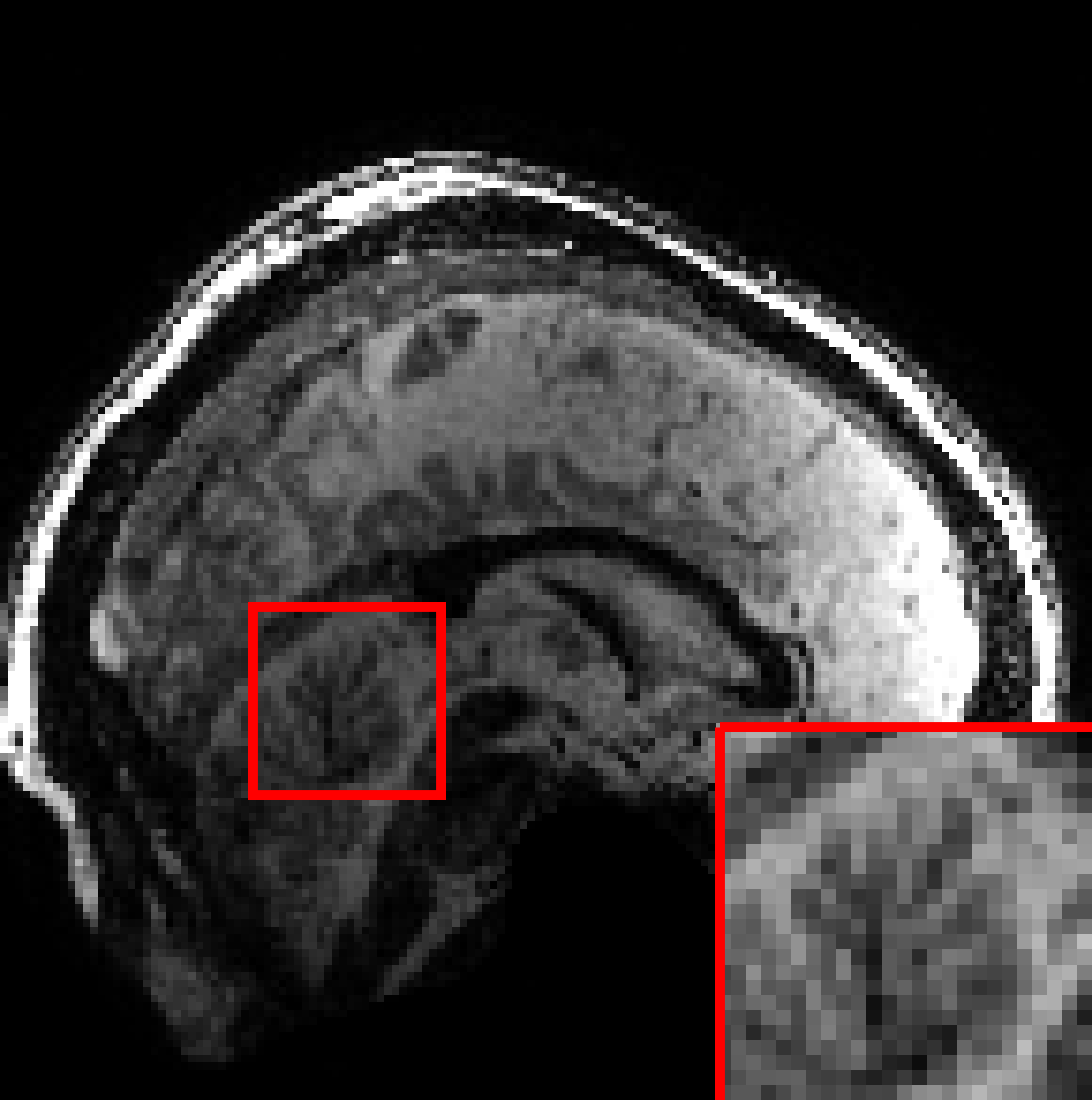}}};
    
    \node[above=1pt, at=(img_12x_same.north),font=\color{blue}, align=center] {\footnotesize $\Vtrain$=$0.7$mm};
    \node[above=1pt, at=(img_12x_2x.north),font=\color{red}, align=center] {\footnotesize $\Vtrain$=$1.4$mm};
    \node[above=1pt, at=(img_12x_4x.north),font=\color{orange}, align=center] {\footnotesize $\Vtrain$=$2.8$mm};
    
    \node[below=1pt, at=(img_12x_same.south),font=\color{purple}, align=center] (labelDiverse) {\footnotesize Diverse};
    \node[below=1pt, at=(img_12x_2x.south),font=\color{black}, align=center] (labelTarget) {\footnotesize Target};
    \node[below=1pt, at=(img_12x_4x.south),font=\color{black}, align=center] (labelPseudorec) {\footnotesize Unreg.};

    \node [below=-0.15cm, at=(labelDiverse.south)](img_48x_same) {\rotatebox[origin=c]{0}{\includegraphics[width=\pngwidth]{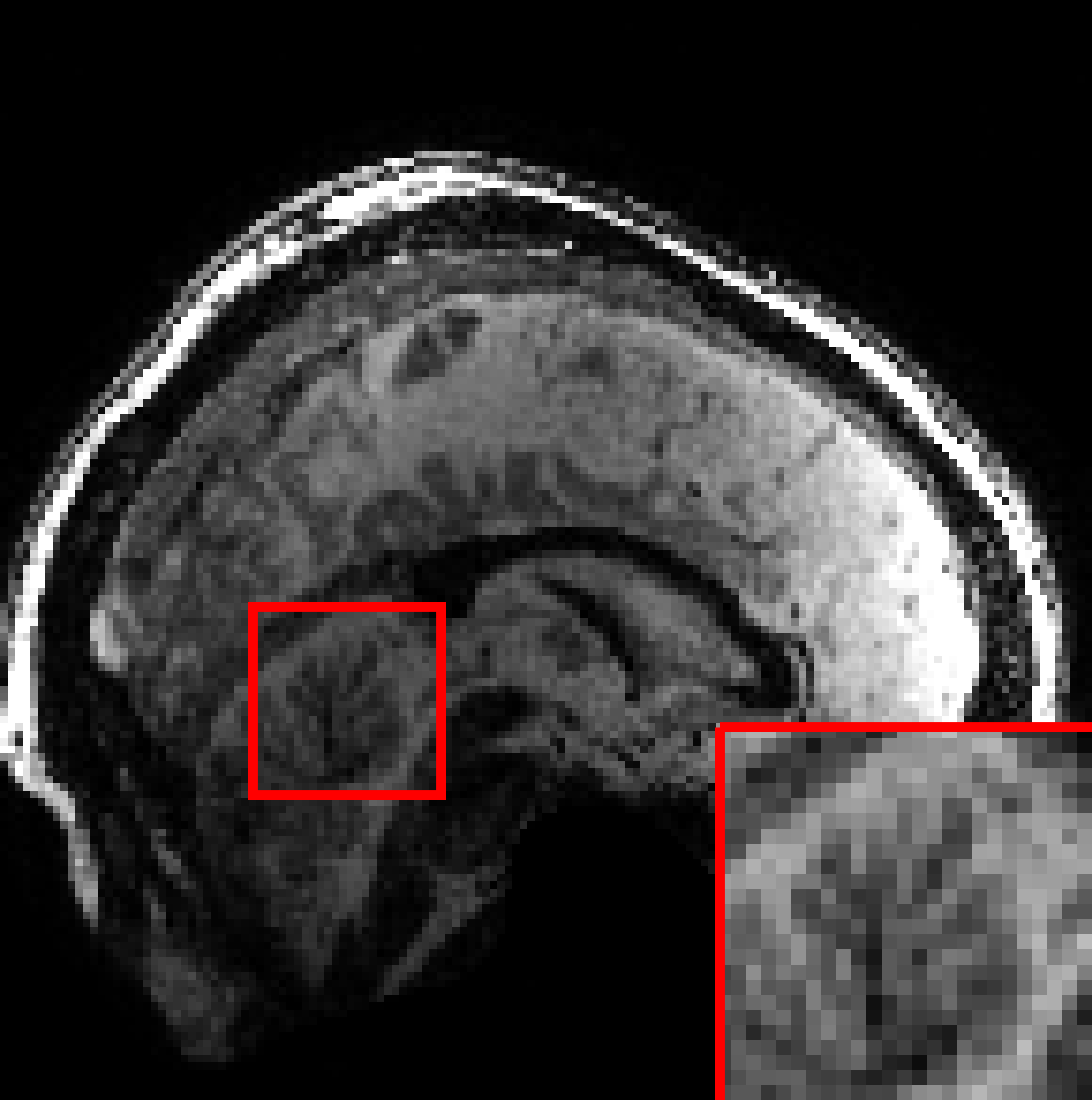}}};
    \node [below=-0.2cm, at=(labelTarget.south)](img_48x_2x) {\rotatebox[origin=c]{0}{\includegraphics[width=\pngwidth]{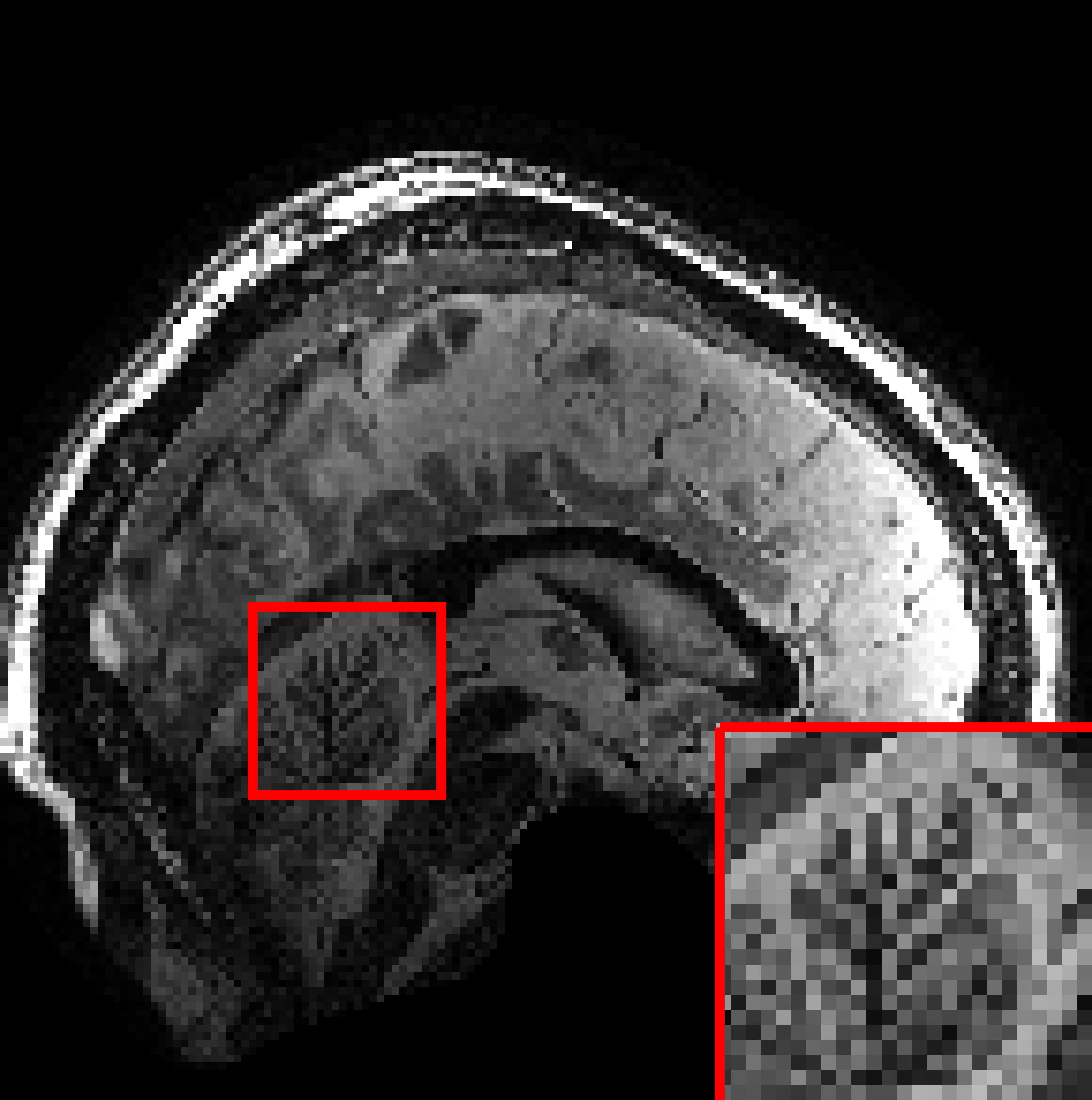}}};
    \node [below=-0.2cm, at=(labelPseudorec.south)](img_48x_4x) {\rotatebox[origin=c]{0}{\includegraphics[width=\pngwidth]{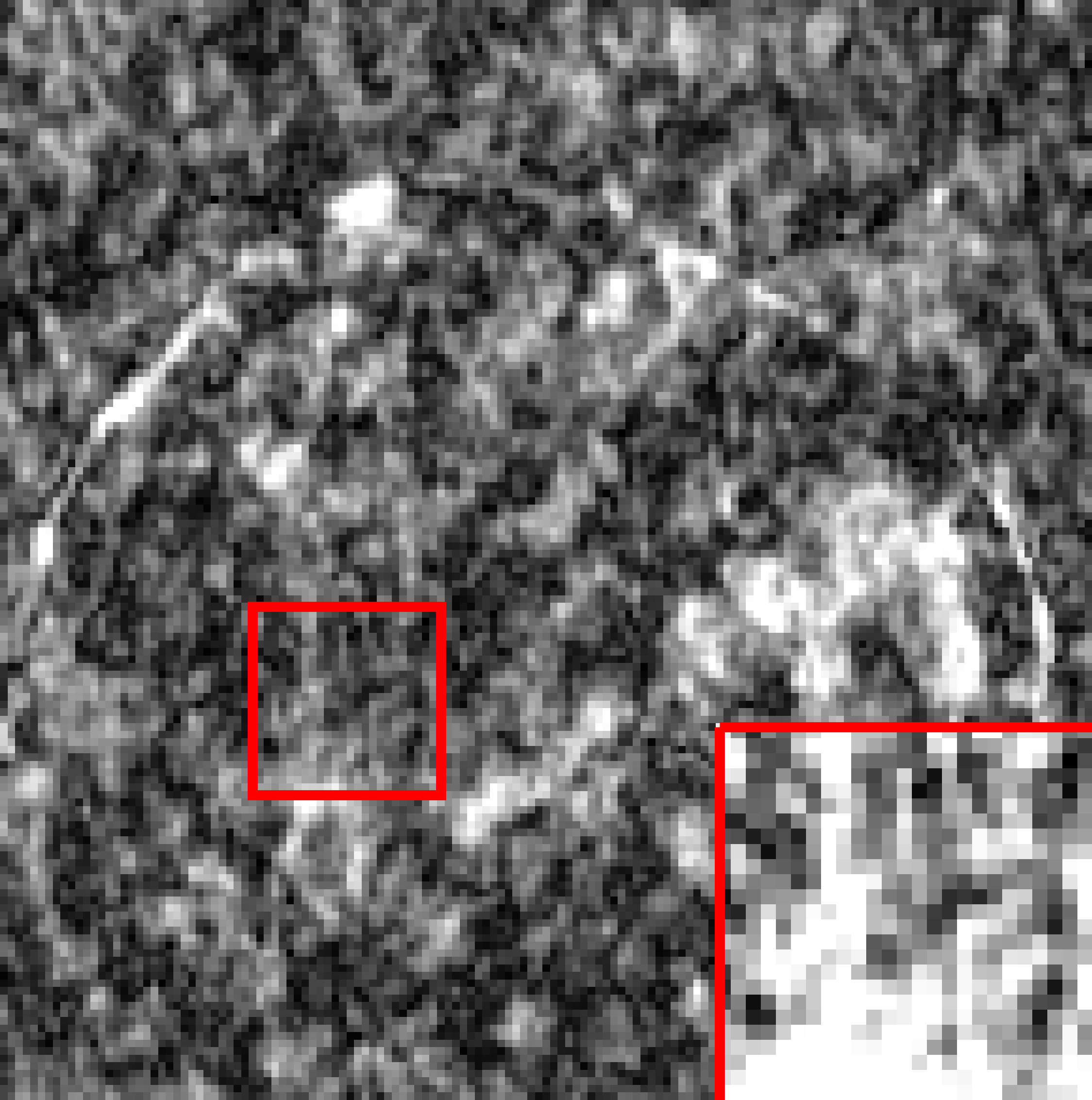}}};

    \node[at=(img_48x_4x.east)]{
        
        \begin{groupplot}[group style={group size=1 by 1},
            width=0.32\columnwidth,
            height=0.32\columnwidth,
            tick align=outside,
            tick pos=left,
            x grid style={darkgray176},
            xtick style={color=black},
            y grid style={darkgray176},
            scaled ticks=false,
            ytick style={color=black},
            legend image code/.code={
                \draw[#1] (0cm,0.0cm) -- (0.4cm,0.0cm); 
            }, 
            legend style={
              fill opacity=0.8,
              font=\small,
              draw opacity=1,
              text opacity=1,
              at={(1.4, 1.0)},
              anchor=north,
              draw=gray}
        ]

        \nextgroupplot[ylabel={PSNR (dB)}, xlabel={acceleration $R$}, xmin=3, xmax=17, ymin=35, ymax=43, title={Quantitative results}, align=center,
            xtick={4, 8, 12, 16},
            xticklabels={4, 8, 12, 16},
            ytick={36, 38, 40, 42},
            yticklabels={36, 38, 40, 42}
        ]
        
            \addplot [draw=blue, fill=blue, forget plot, mark=*, only marks, mark size=1.5]
            table {%
                4   39.723
                8   37.414
                12  36.444
                16  35.828
            };
            \addplot [very thick, blue, dashed]
            table {%
                4   39.723
                8   37.414
                12  36.444
                16  35.828
            };
            \addlegendentry{No shift}
            
            \addplot [draw=red, fill=red, forget plot, mark=*, only marks, mark size=1.5]
            table {%
                4   41.800
                8   39.052
                12  37.889
                16  37.112
            };
            \addplot [very thick, red, dashed]
            table {%
                4   41.800
                8   39.052
                12  37.889
                16  37.112
            };
            \addlegendentry{$2$x-shift}
            
            \addplot [draw=orange, fill=orange, forget plot, mark=*, only marks, mark size=1.5]
            table {%
                4 40.963
                8 37.838
                12 36.813
                16 36.044
            };
            \addplot [very thick, orange, dashed]
            table {%
                4 40.963
                8 37.838
                12 36.813
                16 36.044
            };
            \addlegendentry{$4$x-shift}
            
            \addplot [draw=purple, fill=purple, forget plot, mark=*, only marks, mark size=1.5]
            table {%
                4 41.702
                8 38.939
                12 37.851
                16 37.121
            };
            \addplot [very thick, purple]
            table {%
                4 41.702
                8 38.939
                12 37.851
                16 37.121
            };
            \addlegendentry{Diverse}
    
        \end{groupplot}
    };

\end{tikzpicture}